\documentclass[journal]{IEEEtran}  
\IEEEoverridecommandlockouts                              

\usepackage{pifont}
\let\oldding\ding 
\renewcommand{\ding}[2][1]{\scalebox{#1}{\oldding{#2}}} 
\usepackage{adjustbox}
\usepackage{booktabs}
\usepackage{amsmath,amsfonts}
\usepackage{algorithm}
\usepackage{algpseudocode}
\usepackage{makecell}
\usepackage{amssymb}
\usepackage{xcolor}
\usepackage{soul}
\usepackage{array}
\usepackage[table,xcdraw]{xcolor}
\usepackage{textcomp}
\usepackage{stfloats}
\usepackage{url}
\usepackage{array,multirow,graphicx}
\usepackage{subcaption}
\usepackage{float}
\usepackage{gensymb}
\usepackage{caption}
\usepackage{placeins}
\usepackage{multirow}
\usepackage{flushend}
\usepackage{verbatim}
\usepackage{graphicx}
\usepackage{cite}
\usepackage{hyperref}
\graphicspath{{./Fig/}}	
\usepackage[english]{babel}

\DeclareUnicodeCharacter{2061}{}
\newcolumntype{M}[1]{>{\centering\arraybackslash}m{#1}}
\newcolumntype{N}{@{}m{0pt}@{}}   
\usepackage[flushleft]{threeparttable}
\definecolor{HLsoftyellow}{RGB}{8,232,222}

\begin{document}
\title{\LARGE \bf ARTiS: An Adaptive Robotic Gripper for \\ Enhanced Tool Manipulation in Disassembly Applications}
\author{Roman Mykhailyshyn,~\IEEEmembership{Senior Member,~IEEE,} Yukiyasu Domae,~\IEEEmembership{Senior Member,~IEEE,} \\ and Kensuke Harada,~\IEEEmembership{Fellow,~IEEE}


\thanks{Roman Mykhailyshyn is with the Embodied AI Research Team, National Institute of Advanced Industrial Science and Technology (AIST), Tokyo 135-0064, Japan, and on leave from the Graduate School of Engineering Science, The University of Osaka, Osaka 560-8531, Japan. (Corresponding author e-mail: {roman.mykhailyshyn@aist.go.jp})} %
\thanks{Yukiyasu Domae is with the Embodied AI Research Team, National Institute of Advanced Industrial Science and Technology (AIST), Tokyo 135-0064, Japan. (e-mail: domae.yukiyasu@aist.go.jp)}%
\thanks{Kensuke Harada is with the Graduate School of Engineering Science, The University of Osaka, Osaka 560-8531, Japan, and is also with the Embodied AI Research Team, National Institute of Advanced Industrial Science and Technology (AIST), Tokyo 135-0064, Japan. (e-mail: {harada@sys.es.osaka-u.ac.jp})}}%

\markboth{IEEE Transactions on Automation Science and Engineering,~Vol.~00, No.~0, ~2026}%
{Shell \MakeLowercase{\textit{et al.}}: ARTiS: An Adaptive Robotic Gripper for Enhanced Tool Manipulation in Disassembly Applications}


\maketitle
\thispagestyle{empty}
\pagestyle{empty}

\begin{abstract}
Grasping and holding tools while using them presents a considerable challenge not only for robots but also for humans. Such a challenge is particularly noticeable in processes involving assembly and disassembly, where efficiency and consistency depend on performing rapidly adaptive tasks. Nonetheless, contemporary robotic grasping technologies that can securely manipulate tools during operation frequently have significant constraints. In this paper, introduce ARTiS (\underline{A}daptive \underline{R}obotic \underline{T}ool Gripper \underline{i}n Disassembly \underline{S}ystems), a novel gripper that combines the adaptability of soft grippers, the dexterity of anthropomorphic hands, and the robustness of rigid mechanisms with a soft palm and fingertips. This unique combination makes it possible to hold tools securely in a variety of situations through using active jamming in the palm and fin-ray adaptation in fingertips. Furthermore, high finger dexterity is achieved through the seven degrees of freedom design, which enables the fingertips to orient to any surface, both for automated solutions and collaborative tasks. A comprehensive evaluation was conducted using a range of conventional disassembly tools to assess the gripper's compliance, durability, and functional versatility. More information, hardware instructions, and videos at
\url{https://romanmykhailyshyn.github.io/artis/} 
\end{abstract}

\def\abstractname{Note to Practitioners}
\begin{abstract}
This paper addresses the challenge of grasping and fixing tools in the hand to ensure their use during disassembly. The case study considers the use of three of the most commonly used tools (screwdriver, hammer, and drill) when disassembling individual machine components into smaller components. Many existing gripping devices can grasp hand tools and partially use them; however, they do not provide robust holding and adaptation to surfaces of different geometries. This paper proposes a design of a three-fingered gripping device that uses an effective combination of a jamming palm, rigid fingers, and a fin-ray fingertip for grasping/using hand tools. Using the proposed model, one can calculate the maximum permissible force applied to tools of different lengths. Knowing the frictional parameters between the gripper elements and the tools, one can determine which parameters to choose for the disassembly application based on experimental results. Positive results are highlighted in the grasping of tools without the loss of contact after use. Negative aspects are shown in the formation of vibrations in electric tools while using them with the gripper. Future research plans to improve the design of the gripper by modeling it, real-to-sim transfer, and developing a metal industrial prototype with robot learning implementation.
\end{abstract}


\section{Introduction}
\noindent Automation of assembly/disassembly applications~\cite{vongbunyong2015disassembly} has a significant impact in robotics, as it reveals the main challenges~\cite{foo2022challenges} and benefits of using fully automated and human-collaborative lines~\cite{lee2024review}. The standard method of using tools in automation is to rigidly fix them to the end effector of the robot arm or use automated tool change mechanisms. Instead, there are many available gripping mechanisms~\cite{hernandez2023current, mykhailyshyn2022systematic} that can be used to grasp tools and parts in robotic cells. However, the success of tool grasping is not the primary criterion for the successful assembly or disassembly application of a particular type of gripper. Rather, adaptability to different tools, robustness~\cite{hoffmann2014adaptive}, and tool fixation during the execution of the corresponding applications.

To address these challenges, bionic~\cite{MultiChoiceGripper, yamanaka2020prosthetic, cai2023hand, lee2024index}, soft~\cite{chu2023passively, zhou2017soft}, and reconfigurable~\cite{townsend2000barretthand, kim2025lbh} grippers are used in the design of grasping devices with the necessary adaptability parameters. Also, to increase robustness, anthropomorphic~\cite{grebenstein2011dlr, gao2021dexterous}, three-fingered~\cite{backus2016adaptive, bunis2018caging, golan2020variable}, and combined~\cite{borras2018kit, klas2021kit, han2022screwdriving} designs were proposed. However, existing designs are based on one (adaptive fingers~\cite{chu2023passively}, etc.) or a combination of two innovations (adaptive fingers with dexterity~\cite{MultiChoiceGripper}, adaptive palm~\cite{kremer2023trigger} with dexterous fingers~\cite{lee2021soft}, etc.), which does not allow for completely solving the problem of holding tools during their use. Since the robotic assembly and disassembly must facilitate rapid adaptation to both objects (products) and tools used for this purpose. For example, suppose that it is \textit{disassembling individual elements of a car into smaller components}. In that case, even the same car model of different years of manufacture will have different fasteners, bolts, and other tools for disassembly. This requires significant adaptability, tool fixation, and robustness of the robotic system when using different tools. Two methods of holding instruments are usually used in disassembly~\cite{feix2015grasp, li2020unfastening}, etc.): power wrap and power sphere with fingers, which are not available for existing grippers. We identify the biggest problems for grippers to reproduce, a power sphere fingers holding, which allows the palm to provide robust fixation of the tools, and the fingers to adjust the position of the tool tip.

\begin{figure}[t]
    \centering
    \includegraphics[width=1\linewidth,clip ,trim=3pt 5pt 0pt 3pt]{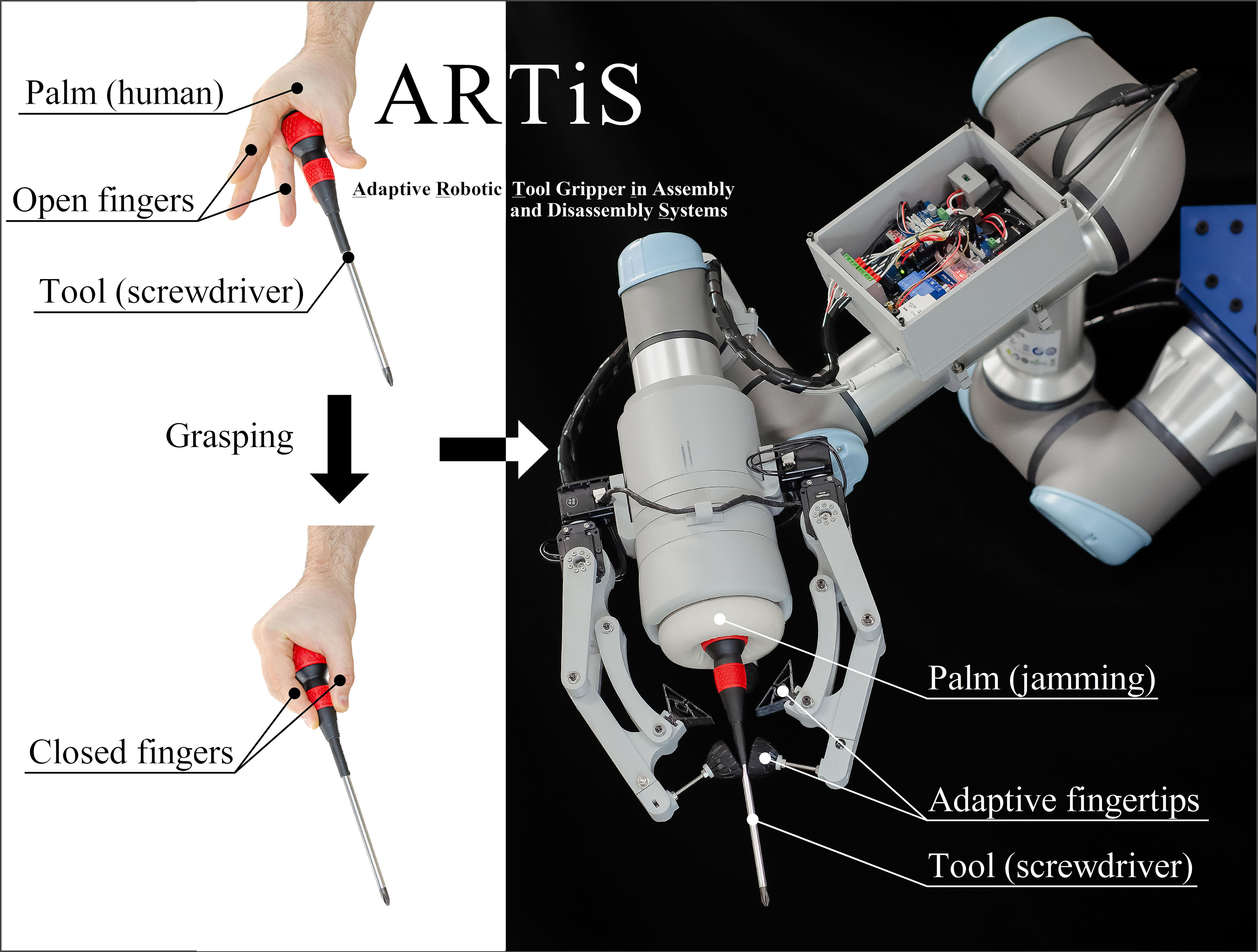}
    \caption{\underline{A}daptive \underline{R}obotic \underline{T}ool gripper \underline{i}n disassembly \underline{S}ystems that holds a screwdriver like a human hand.}
    \label{fig1}
\end{figure}

\begin{figure}[!t]
\centering
 	\begin{subfigure}[b]{0.2\linewidth}
         \centering
        \includegraphics[width=1\linewidth, clip, trim=10pt 0pt 0pt 0pt]{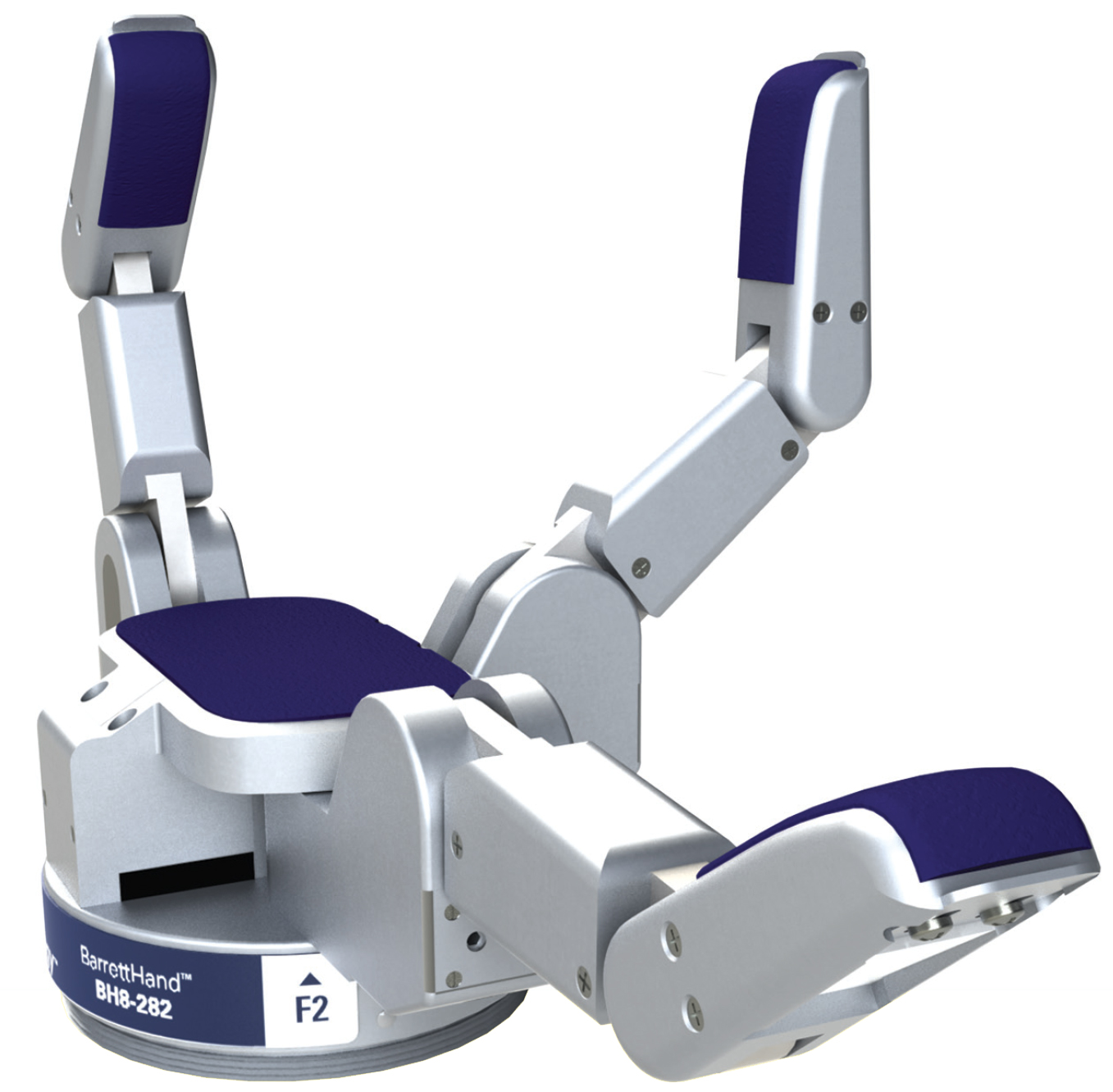}
        \vspace{-6mm}
        \caption{}
        \label{subfig2_1}
    \end{subfigure}
  	\begin{subfigure}[b]{0.165\linewidth}
         \centering
         \includegraphics[width=1\linewidth, clip, trim=220pt 0pt 0pt 320pt]{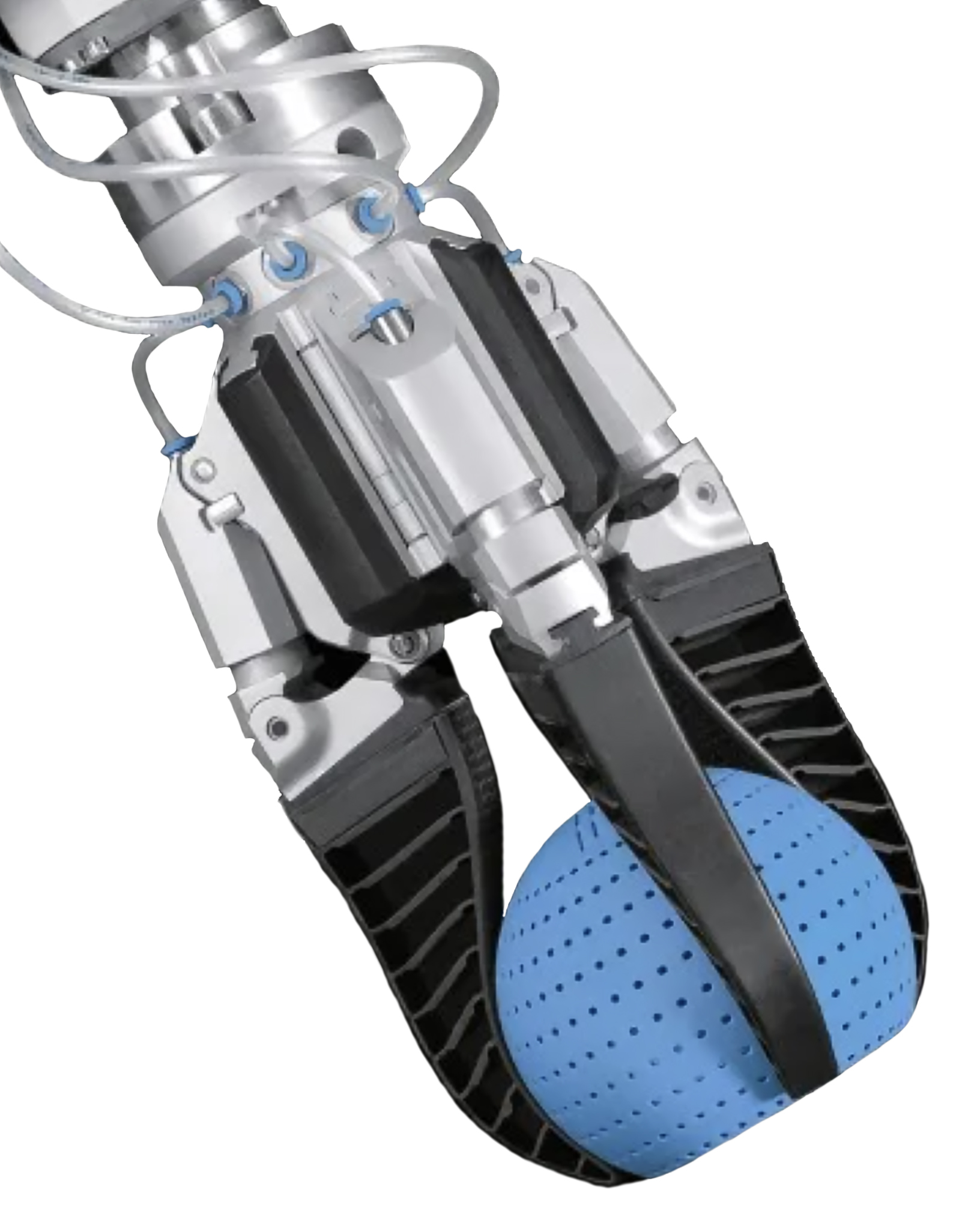}
         \vspace{-6mm}
        \caption{}
        \label{subfig2_2}
    \end{subfigure}
  	\begin{subfigure}[b]{0.24\linewidth}
         \centering
         \includegraphics[width=1\linewidth, clip, trim=140pt 0pt 140pt 0pt]{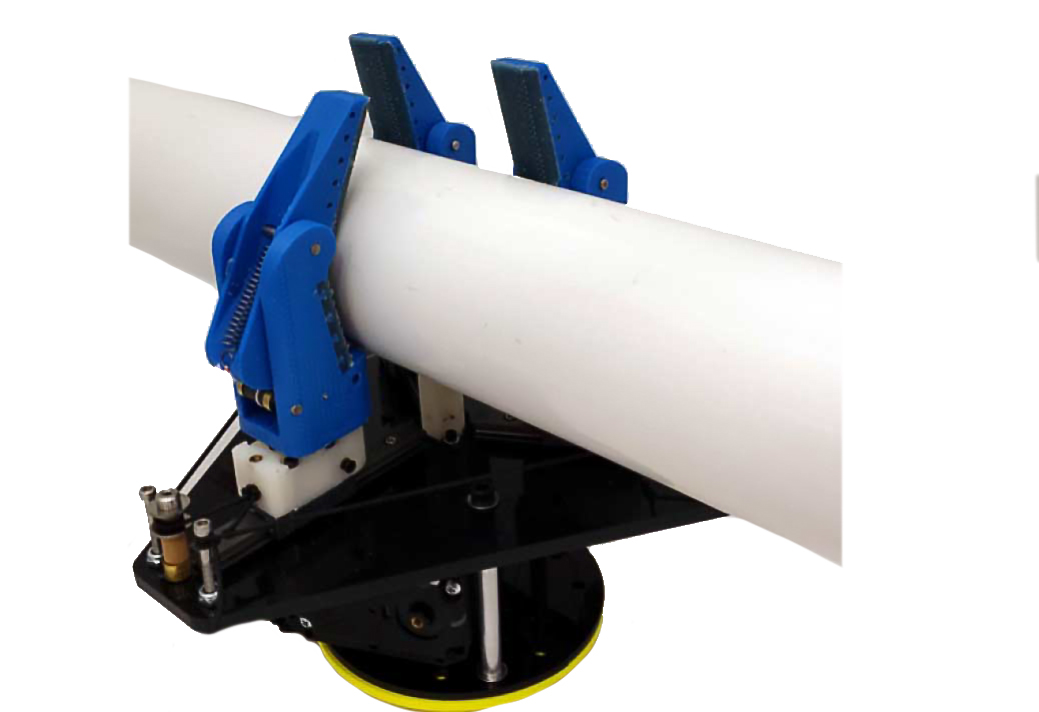}
         \vspace{-6mm}
        \caption{}
        \label{subfig2_3}
	\end{subfigure}
  	\begin{subfigure}[b]{0.15\linewidth}
         \centering
         \includegraphics[width=1\linewidth, clip, trim=0pt 0pt 0pt 0pt]{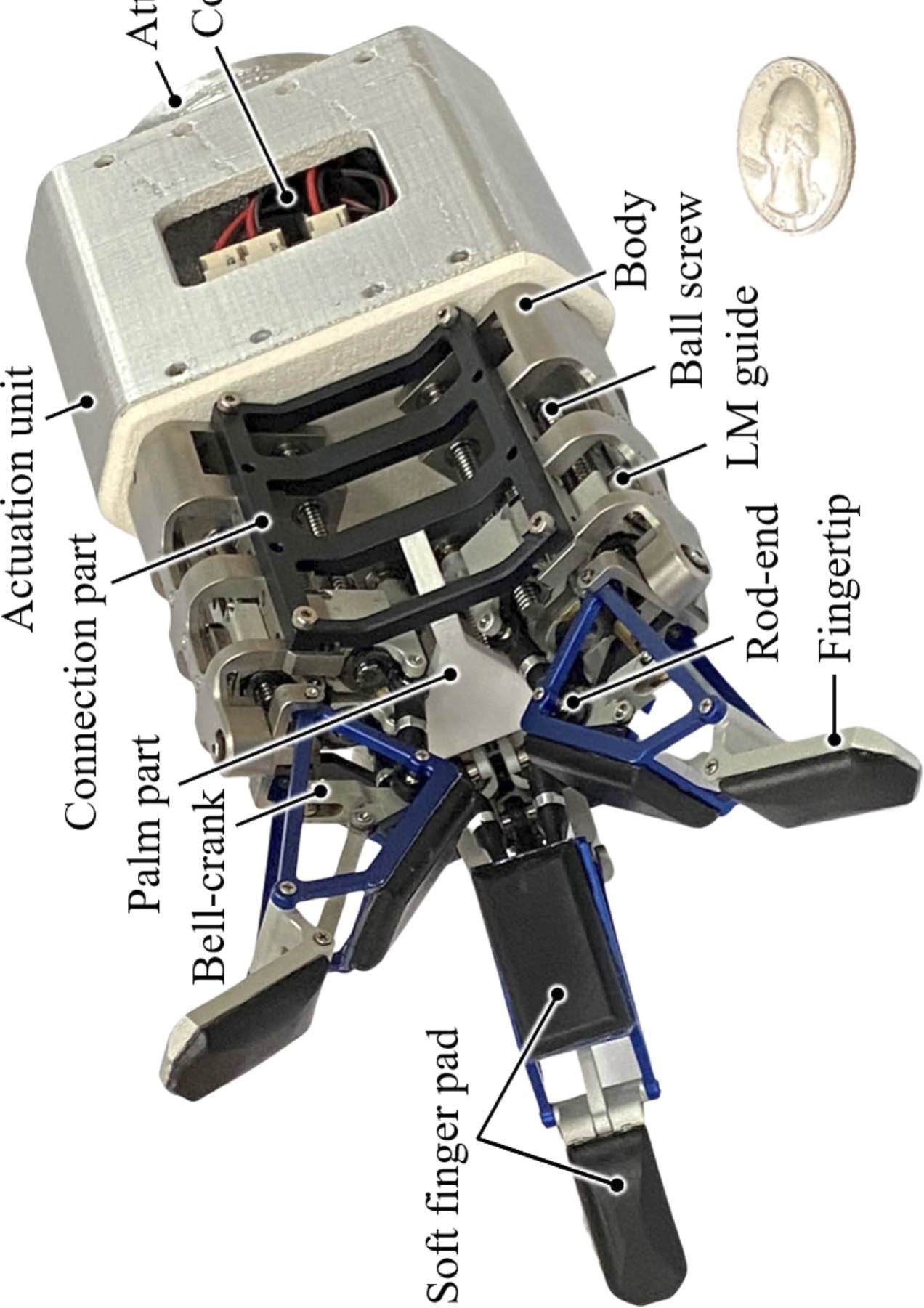}
         \vspace{-6mm}
        \caption{}
        \label{subfig2_4}
	\end{subfigure}
\begin{subfigure}[b]{0.19\linewidth}
         \centering
        \includegraphics[width=1\linewidth, clip, trim=0pt 20pt 0pt 0pt]{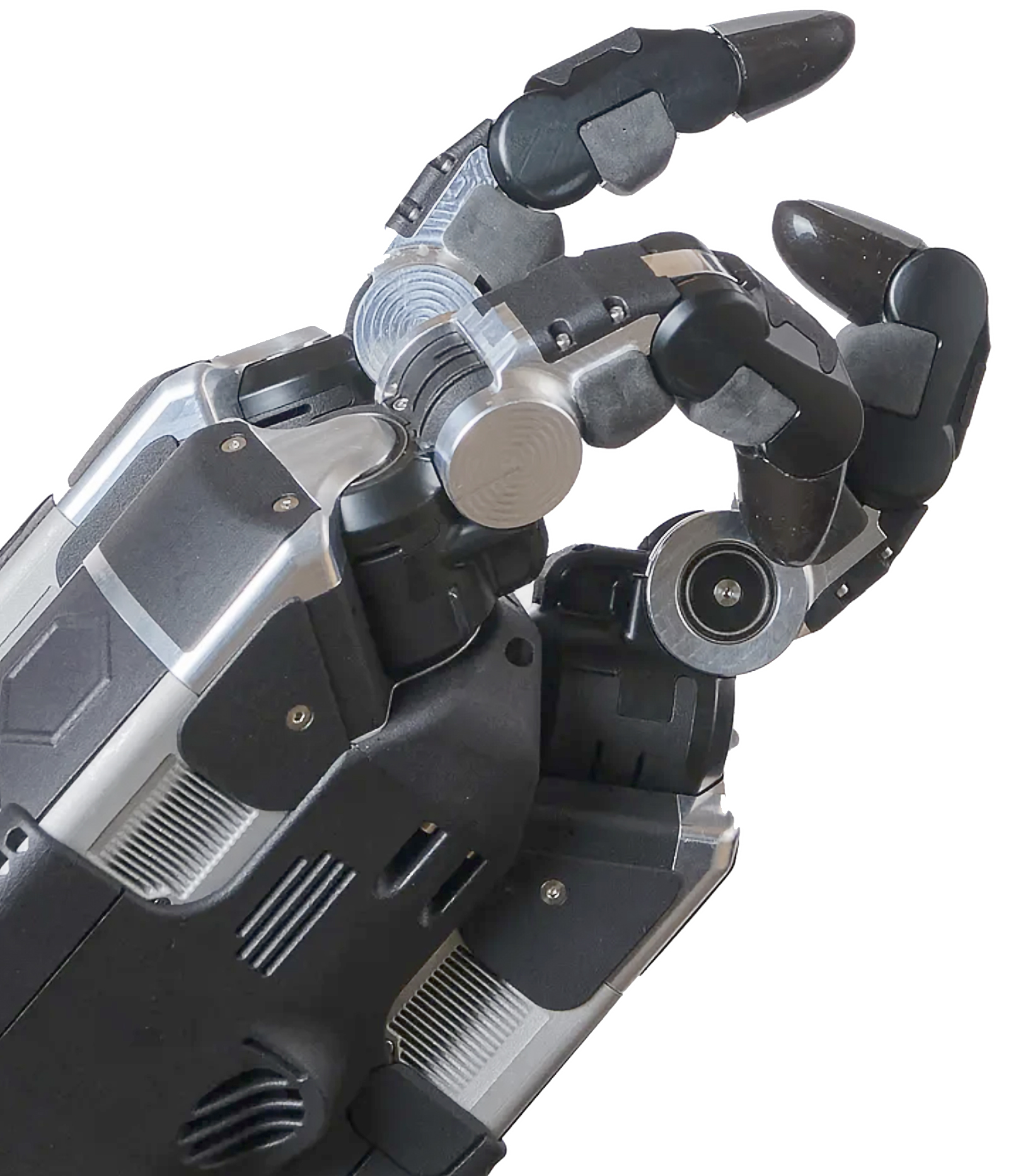}
        \vspace{-6mm}
        \caption{}
        \label{subfig2_5}
    \end{subfigure} \\
    \begin{subfigure}[b]{0.165\linewidth}
         \centering
         \includegraphics[width=1\linewidth, clip, trim=0pt 20pt 0pt 0pt]{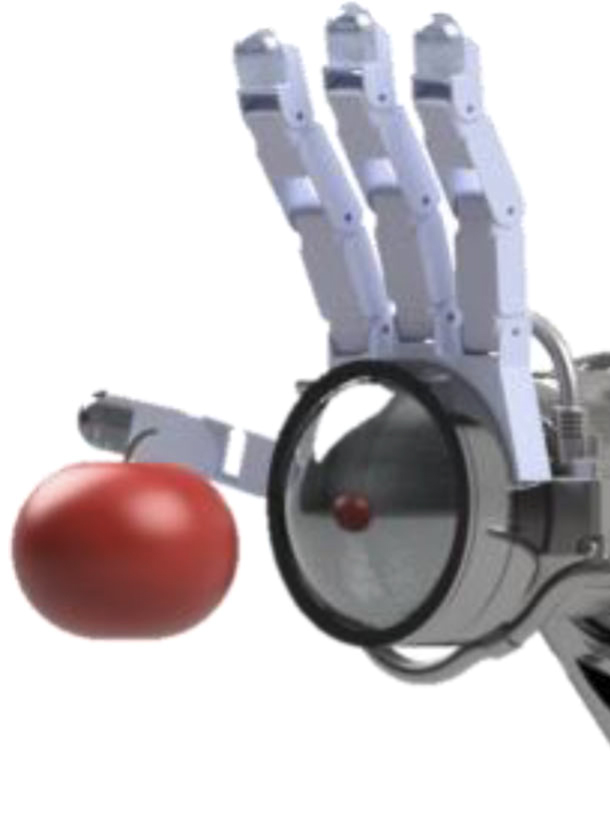}
         \vspace{-6mm}
        \caption{}
        \label{subfig2_6}
	\end{subfigure}
  	\begin{subfigure}[b]{0.23\linewidth}
         \centering
         \includegraphics[width=1\linewidth, clip, trim=0pt 0pt 0pt 0pt]{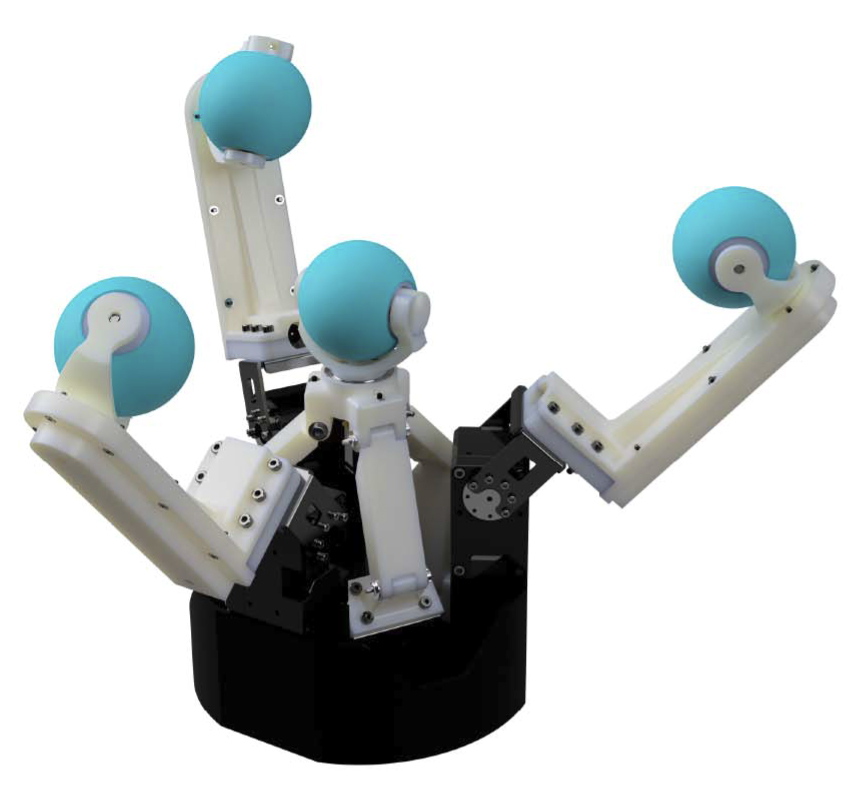}
         \vspace{-6mm}
        \caption{}
        \label{subfig2_7}
    \end{subfigure}
  	\begin{subfigure}[b]{0.18\linewidth}
         \centering
         \includegraphics[width=1\linewidth, clip, trim=0pt 0pt 0pt 0pt]{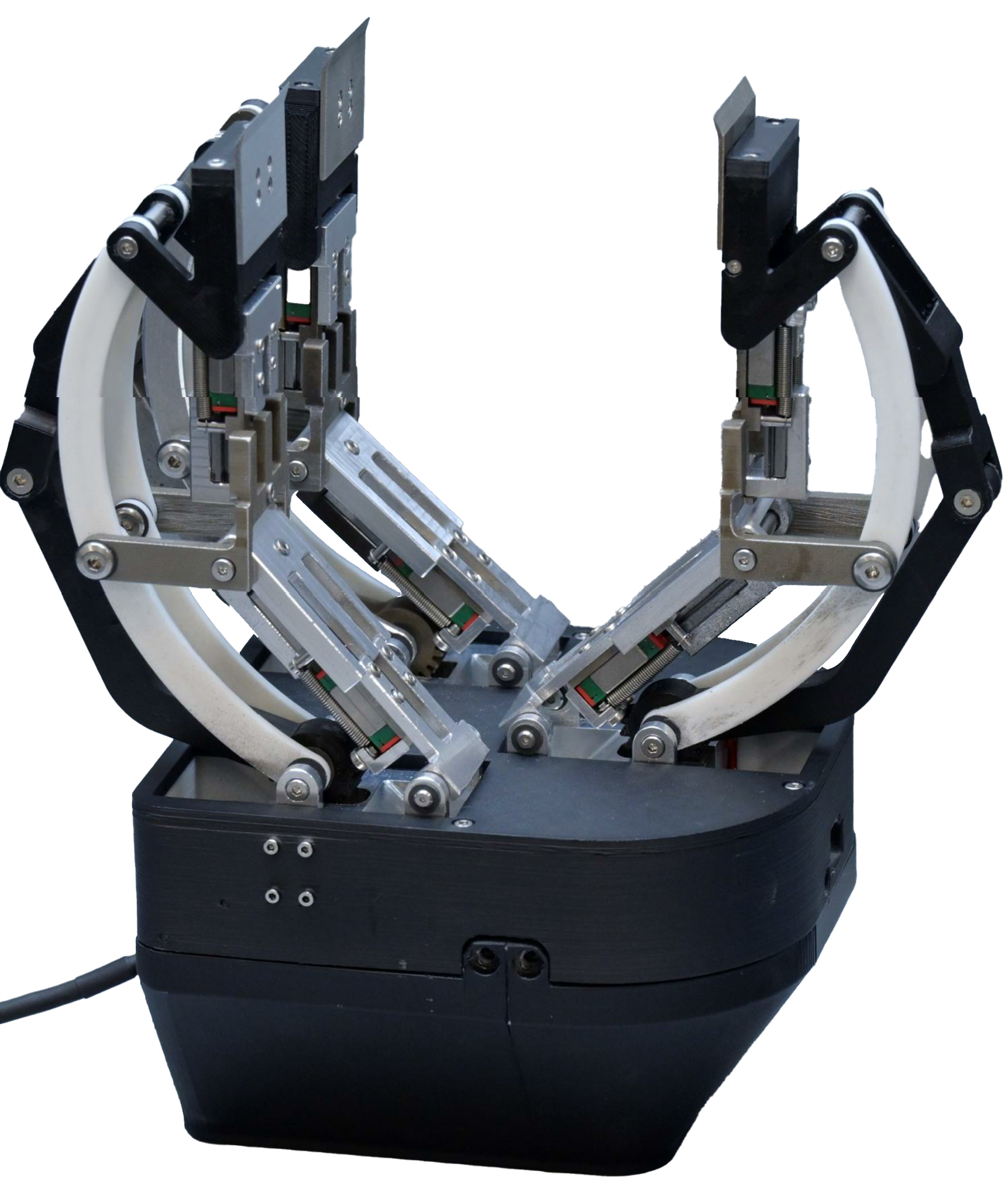}
         \vspace{-6mm}
        \caption{}
        \label{subfig2_8}
	\end{subfigure}
  	\begin{subfigure}[b]{0.17\linewidth}
         \centering
         \includegraphics[width=1\linewidth, clip, trim=0pt 0pt 0pt 0pt]{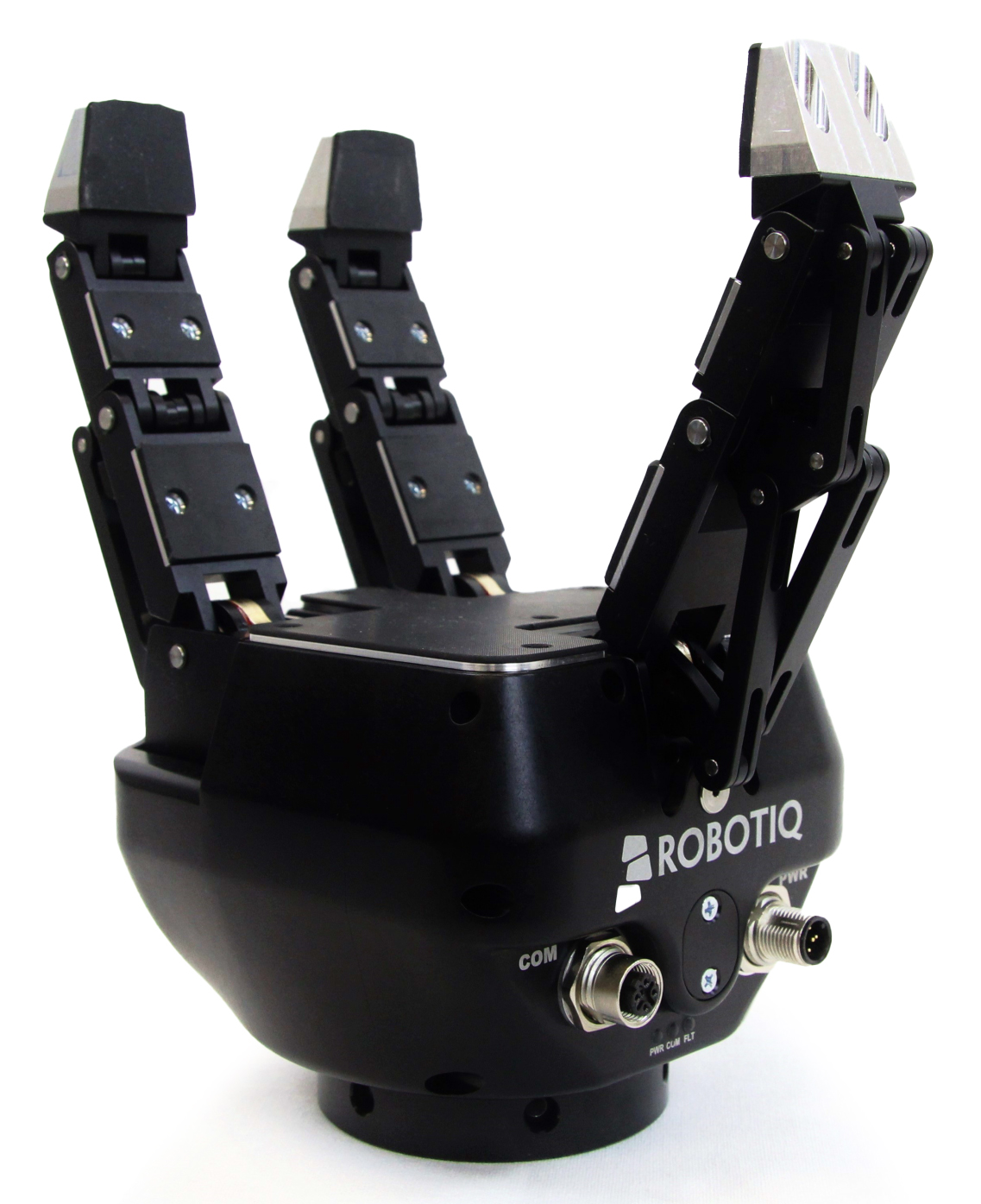}
         \vspace{-6mm}
        \caption{}
        \label{subfig2_9}
	\end{subfigure}
    \begin{subfigure}[b]{0.205\linewidth}
         \centering
         \includegraphics[width=1\linewidth, clip, trim=30pt 0pt 0pt 0pt]{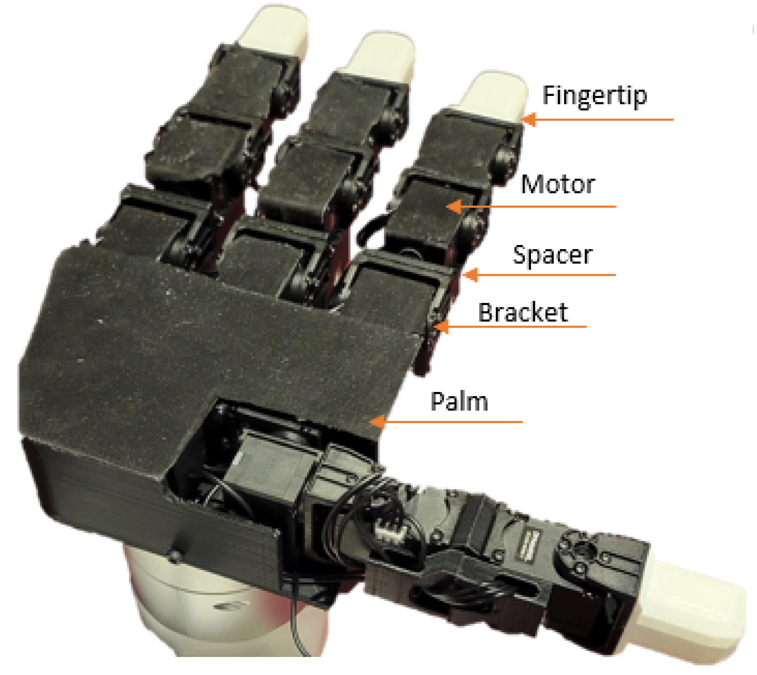}
         \vspace{-6mm}
        \caption{}
        \label{subfig2_10}
	\end{subfigure}
    \\
        \vspace{-1mm}
 \caption{Diverse three-finger grippers with active and passive adaptability: (a) BarrettHand grasper programmably part handling and assembly~\cite{townsend2000barretthand}; (b) Festo LLC multi-choice-gripper~\cite{MultiChoiceGripper}; (c) adaptive three-fingered prismatic gripper with passive rotational joints~\cite{backus2016adaptive}; (d) INDEX dexterous robotic gripper~\cite{lee2024index}; (e) Shadow Robot’s gripper DEX-EE~\cite{ackerman2024shadow}; (f) soft jamming palm With with finger gripper~\cite{lee2021soft}; (g) roller grasper V3 for in-hand manipulation~\cite{yuan2024design}; (h) under-actuated robotic gripper~\cite{li2024under}; (i) 3-finger adaptive robot gripper~\cite{sadun2016grasping}; (j) LEAP Hand~\cite{Shaw2023LEAPHL}.}
 \label{fig2}
\end{figure}

Therefore, there is a significant gap in grippers that would combine: sufficient dexterity to ensure finger reconfiguration for tool tip control, adaptive fingertips to ensure reliable contact and robustness, and an active palm that would allow it to adapt to different shapes of the tools with the function of their fixation. By considering an adaptive three-finger gripper (Fig.~\ref{subfig2_3}) that can easily grip screwdrivers of various shapes~\cite{backus2016adaptive}, it is possible to achieve a high success rate for pick-and-place operations. However, if you try to use a screwdriver while holding it in this gripper, it will start to spin without transmitting sufficient torque to the screw. This is due to the lack of a locking palm or a large contact area with increased friction to prevent the tool from spinning. Therefore, the absence of one of these elements leads to narrow possibilities for grasping and holding tools during use.

In this paper, we present ARTiS, an adaptive robotic gripper designed not only for grasping tools, but also for fixing, reorienting, and using them during disassembly tasks (Fig.~\ref{fig1}). The design was motivated by representative tools used in automotive component disassembly provided by Denso Robotics, while its adaptability and generalization capability were further evaluated using different objects.

The main novelty of ARTiS is its palm-finger cooperative tool-fixation architecture, in which the active palm and novel fingertips perform complementary roles during tool interaction. Unlike conventional grippers that mainly rely on finger contact, ARTiS uses an adaptive jamming palm (with unique granules) to adapt to and fix the tool body, while actively reorienting fingers, stabilizing the grasp, and adjusting the tool pose. The contributions of this paper are summarized as follows:
\begin{itemize}
\item A palm-finger cooperative gripper architecture that combines an adaptive jamming palm with actively reconfigurable fingers for robust tool fixation.
\item A uniquely shaped palm jamming cup has been designed with a cutout for perpendicular holding of tools and a filter for unique rigid glass granules.
\item Novel 3D-adaptive pose-specific fingertips, designed to improve contact adaptation with different tool handles and support stable tool interaction.
\item Experimental validation of tool use, rather than only tool grasping, including grasping, fixing, reorientation, and operation of screwdrivers, hammers, and drills.
\item Extended evaluation of grasping generalization and a teaching mode for human-robot collaboration.
\end{itemize}

Beyond the target disassembly tools, ARTiS is further evaluated on selected YCB benchmark objects from food, kitchen, and tool categories to assess its generalization grasping capability. In addition, a human study on unscrewing three screws is conducted to evaluate the effectiveness of the teaching mode. This mode allows a human operator to configure and record gripper states for repeated collaborative tasks, such as fixing a workpiece during manual operation, and can also provide kinesthetic demonstrations for imitation learning.


\vspace{-2mm}
\section{Related Works}

\noindent A literature review revealed a large number of gripping device designs that have the ability to grasp tools, have a palm with appropriate robustness. The most common is two-finger grippers, but they are rarely used for grasping tools, as they do not provide sufficient holding stability~\cite{hoffmann2014adaptive}. Therefore, most grippers for assembly and disassembly use three fingers with their positions at 120 degrees relative to the gripper axis~\cite{townsend2000barretthand, MultiChoiceGripper,  ackerman2024shadow, lee2024index, lee2021soft,  li2024under, li2015reconfigurable, backus2016adaptive, golan2020variable, wang2024robot, yuan2020design, yuan2024design, cai2023hand, yamanaka2020prosthetic, jiang2020design, zhou2017soft, liu2020optimal, bunis2018caging, sadun2016grasping, mathew2021resoft, wang2025dexgrip, subramaniam2020design, beddow2021caging, pagoli2021soft, kim2020blt, kim2025lbh, kang2026design, khalid2025design} (Fig.~\ref{fig2}). This is completely justified from the point of view of grasping taxonomy using the human hand~\cite{feix2015grasp}. However, the presence of three fingers in the design does not make the gripper suitable for the use of hand tools in disassembly applications. Therefore, to analyze existing gripper analogs, a comparison of three/four-fingered grippers was compiled in Table~\ref{tab:comp}. The first part presents a description of the structural components of the fingers, fingertips, and palm. The second part presents the necessary properties (robustness, grasping versatility, task versatility, tool usage) for robotic disassembly applications based on mechanical capability. The importance of each of these necessary grasping properties for robotic disassembly applications is described in Section~\ref{GR}.

\begin{table*}[!t]
\centering
\scriptsize
\caption{Comparison of robotics three/four-fingered grippers, based on structural components (number of fingers, degrees of freedom per finger, finger actuation type, finger sensing, number of fingertips per finger, fingertip type, fingertip sensing, palm type, palm actuation type, palm sensing) and necessary properties for robotics disassembly applications based on mechanical capabilities.}
\label{tab:comp}
\scriptsize
\resizebox{\textwidth}{!}{%
\setlength\tabcolsep{2pt}
\renewcommand{\arraystretch}{1.25}
\begin{tabular}{ccccccccccccccc}
\hline
\multirow{2}{*}{Gripping Device} &
  \multicolumn{4}{c}{Fingers} &
  \multicolumn{3}{c}{Fingertip} &
  \multicolumn{3}{c}{Palm} &
  \multirow{2}{*}{Robustness} &
  \multirow{2}{*}{Grasping Versatility} &
  \multirow{2}{*}{Task Versatility} &
  \multirow{2}{*}{Tools Using} \\ \cline{2-11}
 & Number & DoF & Actuation  & \multicolumn{1}{c|}{Sensing} & Number & Type & \multicolumn{1}{c|}{Sensing} & Type & Actuation & Sensing &  &  &  &  \\ \hline
BarrettHand~\cite{townsend2000barretthand} (Fig.~\ref{subfig2_1}) &  3 &  3 &  motor (differential) &  torque &  1 & friction surface &
  capacitive &  friction surface &  N/A &  capacitive &  - &  \checkmark &  \checkmark &  \checkmark \\ 
 Festo Multi-Choice-Gripper~\cite{MultiChoiceGripper} (Fig.~\ref{subfig2_2})  &  3 &  2 &  pneumatic (piston) &  - &  1 &  fin-ray (passive) &  N/A &  - &  - &  - &  \checkmark &  - &  - &  - \\ 
Adaptive Gripper with Passive Joints~\cite{backus2016adaptive} (Fig.~\ref{subfig2_3})  &  3 & 2 &   motor (tendon)  &    - &    1 &  friction surface &  - &  - &   - &  - &  \checkmark &  - &  - &   -\\
Passive Adaptive Gripper with Active Surfaces~\cite{cai2023hand} &  3 & 1 & servo (direct)  & torque   &    1 &  fin-ray (active)  & - &  - &  - &  - & \checkmark & - & - & - \\
Shadow Robot’s Gripper DEX-EE~\cite{ackerman2024shadow} (Fig.~\ref{subfig2_5}) &  3 &  4 & servo (tendon) &   force  & 3 &  silicon &  vision, proximity &  - & - & - & \checkmark & - & \checkmark & \checkmark \\
Variable-Structure Robot Hand~\cite{golan2020variable} &  3   & 2   &  motor (differential)  &  -  & 1  &  cylindrical &  -     &   -  &    -   &   -  & \checkmark & - & - & - \\
 INDEX Gripper~\cite{lee2024index} (Fig.~\ref{subfig2_4}) &  3 &  3 & motor (differential) &  - &   2 & silicon &  - & -  & - & - & - & \checkmark & \checkmark & - \\
 Soft Jamming Palm with Fingers~\cite{lee2021soft} (Fig.~\ref{subfig2_6}) &  3-4 & 3 & servo (direct)  &  -  &  3 &  silicon &  -  & jamming (active) &  pneumatic & pressure &   -  &  \checkmark & \checkmark & \checkmark \\
 Roller Grasper V3~\cite{yuan2024design} (Fig.~\ref{subfig2_7}) &  3 & 3 & servo (direct)  &  -  & 1  & roller (silicon) & - & roller (active) &  servo (direct) &  - & - & \checkmark & \checkmark & - \\
 Under-Actuated Robotic Gripper~\cite{li2024under} (Fig.~\ref{subfig2_8}) &  3 & 1 &   motor (differential) &  -  &  1  &  friction surface &  - &   - &   -  &  -   & \checkmark & - & - & - \\
 3-Finger Adaptive Robot Gripper~\cite{sadun2016grasping} (Fig.~\ref{subfig2_9}) & 3  & 3 & motor (differential) & - &  3 & friction surface &  - &  friction surface &  N/A &  - &  - &  \checkmark &  \checkmark &  \checkmark \\
 LEAP Hand~\cite{Shaw2023LEAPHL} (Fig.~\ref{subfig2_10}) &  4 & 3 & servo (direct)   &   torque   &   4  & silicon &  - &  friction surface &  -    &  - & - & \checkmark & \checkmark & - \\
 Hand with Tendon-Driven Telescopic Fingers~\cite{wang2024robot} & 3 & 1 & servo (tendon)  & torque  &  4  & friction surface & - & - & - & - & \checkmark & \checkmark & - & - \\
 Three-Finger Soft Robotic Gripper~\cite{liu2020optimal} & 3 &  1 &  motor (differential) &  -  & 1  &  fin-ray (passive) &   - & - & -  & - & \checkmark & - & - & - \\
 DexGrip: Multi-modal Soft Gripper~\cite{wang2025dexgrip} &  3 & 1 & servo (direct)  & -   &   1 &  fin-ray (active)  & - & suction cup  &  servo &  - & \checkmark & - & \checkmark & - \\
 Hybrid Soft Gripper with Active Palm~\cite{subramaniam2020design} & 3 & 1 & pneumatic (soft) & - & 4 & silicon & - & inflatable & pneumatic (soft) & - & \checkmark & - & - & - \\
 Caging Inspired Gripper~\cite{beddow2021caging} & 3 & 1 & motor (differential)   &  -  & 1  & metal (flexible) & -   &  soft (pasive)  &   motor (differential)  & -  & -  & - & \checkmark & - \\
 Soft Gripper with an Active Palm and Fingers~\cite{pagoli2021soft} &  3 & 3 & pneumatic (soft) & - & 1 & silicon & - & suction cup  &  servo &  - & \checkmark & \checkmark & - & - \\
 BLT Gripper: Gripper with Active Transition~\cite{kim2020blt} & 3  & 2 &  motor (differential)   & - &  1 & belt (active) & - & - & - & - & \checkmark & \checkmark & - & - \\
 LBH Gripper: Linkage-Belt Adaptive Gripper~\cite{kim2025lbh}& 3 & 2 &  motor (differential)  & - &  1  & belt (active) & - & - & - & - & \checkmark & \checkmark & - & - \\
 Reconfigurable three-finger gripper~\cite{kang2026design} & 3 & 3 &  servo (linkage) &  torque &  2 & friction surface  &    - & friction surface &  N/A   &  -  & - & \checkmark & \checkmark & - \\
 Universal Gripper With Self-Adaptable Fingers~\cite{khalid2025design} & 3 & 1 &  motor (linkage)  &   - &  1 &  fin-ray (passive)  &  -  & - & - & -  & \checkmark & - & - & - \\ 
 \hline
 ARTiS (Fig.~\ref{fig1}) &
  3 &
  3 &
  servo (linkage) &
  torque &
  2 &
  fin-ray (passive) &
  N/A &
  jamming (active) &
  pneumatic &
  - &
  \checkmark &
  \checkmark &
  \checkmark &
  \checkmark \\ \hline
\end{tabular}
}
\end{table*}

\subsubsection{Insufficient Finger Reorientation and Fingertip Adaptation}

Gripping devices that have less than or equal to 2 degrees of freedom (DoF) per finger do not allow for control of their position and/or orientation to ensure adaptation to objects (tools) of different shapes/sizes during grasping. Therefore, the grippers are unable to provide Grasping Versatility (Table~\ref{tab:comp}, all grippers with such parameters have a minus (-) for this property). Only some gripper designs are made to actively change the position~\cite{golan2020variable, townsend2000barretthand} (Fig.~\ref{subfig2_1}) or orientation~\cite{MultiChoiceGripper, li2015reconfigurable, lee2024index, ackerman2024shadow, lee2021soft} (Fig.~\ref{subfig2_2},~\ref{subfig2_4},~\ref{subfig2_5}, ~\ref{subfig2_6}) of the fingers. However, a large number of weak motors rigidly connected to the fingers (differential, linkage, direct)~\cite{townsend2000barretthand, lee2024index, lee2021soft, sadun2016grasping, Shaw2023LEAPHL} lead to a decrease in the robustness of the gripper and can easily lead to breakage due to the lack of damping on the fingers (Fig.~\ref{subfig2_1},~\ref{subfig2_4},~\ref{subfig2_5},~\ref{subfig2_9},~\ref{subfig2_10}). Of course, there are design options where passive soft elements~\cite{cai2023hand, kim2025lbh, liu2020optimal, zhou2017soft, backus2016adaptive, mathew2021resoft} are used that allow partial change in the position or orientation of the fingers during grasping (Fig.~\ref{subfig2_2},~\ref{subfig2_3}), or even scroll to enable in-hand manipulation (Fig.~\ref{subfig2_7}). However, these adaptive grippers have insufficient degrees of freedom and/or bulky adaptive elements. The adaptive elements are so large that it is not possible to predict how they will adapt during application, and they subsequently cause problems with the stability of holding the tools during use. With ARTiS, our goal is to give the fingers the ability to continuously change position and orientation, which contributes to the adaptability of the rigid elements. This is important from the point of view of choosing a rational position of the fingers with the fingertips to apply forces in a specific order and, if necessary, adjust the position or orientation of the tool tip. On the other hand, in the adaptability of human fingers in ARTiS, we enforce passively adaptive fingertips to provide maximum contact area while damping the interaction of the tools with the fingers.


\subsubsection{Insufficient Tool Fixation}

Among all three finger grippers analyzed (Table~\ref{tab:comp}), most grippers do not have a palm, which significantly minimizes the possibility of fixing tools or providing an additional surface to which the tool can be pressed~(Fig.~\ref{fig2_5}). An exception is Shadow Robot’s gripper DEX-EE~\cite{ackerman2024shadow}, due to its large number of fingertips per finger, which allows robust tool fixation. However, due to the inability to fully open the fingers and the lack of a palm, it does not provide sufficient grasping versatility. Therefore, an active palm plays an extremely important role in the grasping and holding of tools, which was well studied in the paper~\cite{pozzi2023actuated}. There are several different types of palm used in gripper designs, such as jamming~\cite{lee2021soft, li2019novel, harada2016proposal, zhang2022finger, kremer2023trigger}, semi-rigid~\cite{beddow2021caging, teeple2021active, mathew2021resoft}, and soft, which are separate~\cite{subramaniam2020design, pagoli2021soft, liu2023dexterous} and are part~\cite{zhao2023palm, kim2020blt} of the fingers. In most of these active palm designs, the focus is on increasing the adaptability of the hand by taking the shape of the object. This includes increasing the contact area and the friction force to prevent slippage and increase grasping stability. Such adaptability is quite reasonable from the point of view of pick-and-place operations, but when talking about using tools in assembly or disassembly, it is necessary for the palm of our gripper to fix the tool~(Fig.~\ref{fig2_5}) and carry the main load.

As a result, demonstrate that no gripper satisfies all the necessary properties for robotic disassembly applications (Table~\ref{tab:comp}). However, several grippers can use some tools in assembly/disassembly: BarrettHand~(Fig.~\ref{subfig2_1}), Shadow Robot’s Gripper~(Fig.~\ref{subfig2_5}), soft jamming palm with finger gripper~(Fig.~\ref{subfig2_6}), and 3-finger gripper~(Fig.~\ref{subfig2_9}). The most efficient and complex are grippers with three fingers and an integrative active palm, and the closest to satisfying all properties is the soft jamming palm with a 4-finger gripper~(Fig.~\ref{subfig2_6}). Since this gripper was developed for household applications, there are many problems for disassembly applications: 1 - the small diameter and depth of the palm (which do not allow the tool to be deeply inserted for reliable fixing); 2 - the round shape of the rigid palm flange (which does not allow the grasp of a hammer or other horizontally positioned tool to be deeply inserted into the palm); 3 - the use of a deformable material in the design of the jamming palm (which causes additional adaptability when activating the palm and reduces the ability to withstand the torque of the tool); 4 - the fragility of the fingers and the stiffness of the fingertips (which does not allow sufficient force to be applied to the tools). At ARTiS, our approach to achieving the above-described parameters is a combination of continuously oriented fingers with adaptive fingertips (robustness and adaptability) and an active jamming palm (fixation). Together, the combination of these features gives the versatility of power grasping tools and their subsequent use. The additional use of smart servos in the design allows for force and position feedback, which enables human-robot interaction.

\section{Requirements in a Gripper Tool Operation}\label{GR}
\begin{figure}[!t]
\centering
 	\begin{subfigure}[b]{1\linewidth}
         \centering
    \includegraphics[width=1\linewidth, clip, trim=0pt 0pt 0pt 0pt]{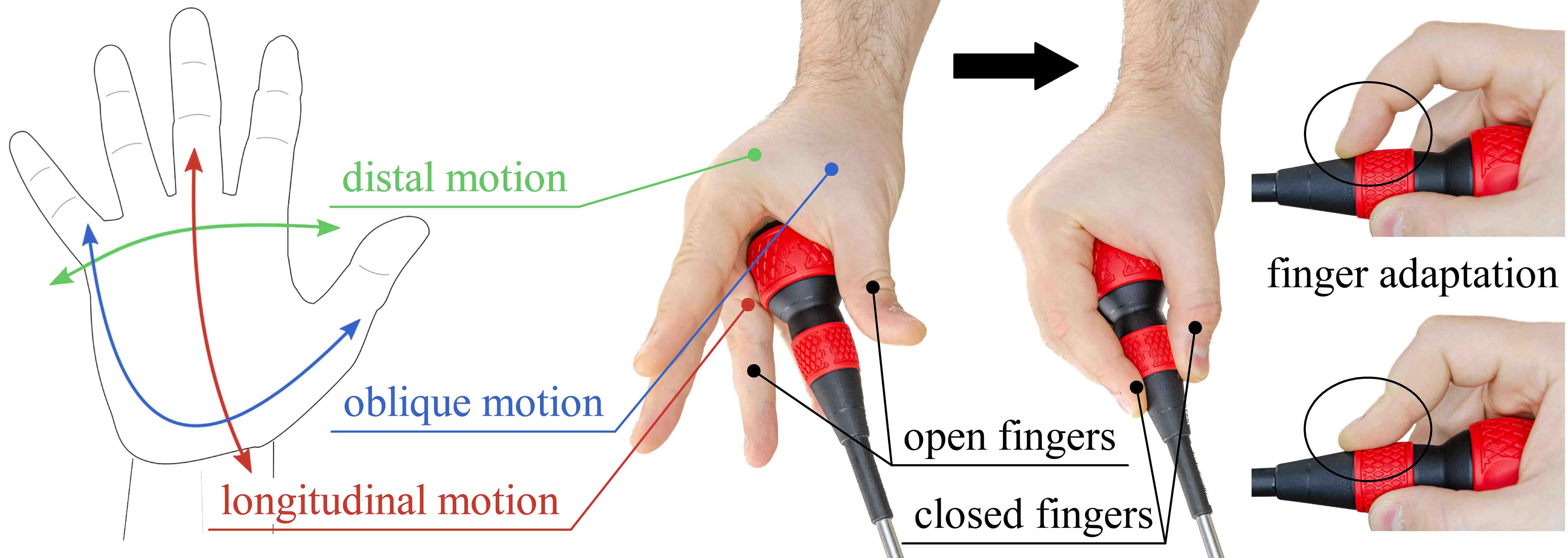}
        \vspace{-5.5mm}
        \caption{}
        \label{fig2_5}
    \end{subfigure} 
    \\
  	\begin{subfigure}[b]{1\linewidth}
         \centering
         \vspace{2mm}
    \includegraphics[width=1\linewidth, clip, trim=0pt 0pt 0pt 5pt]{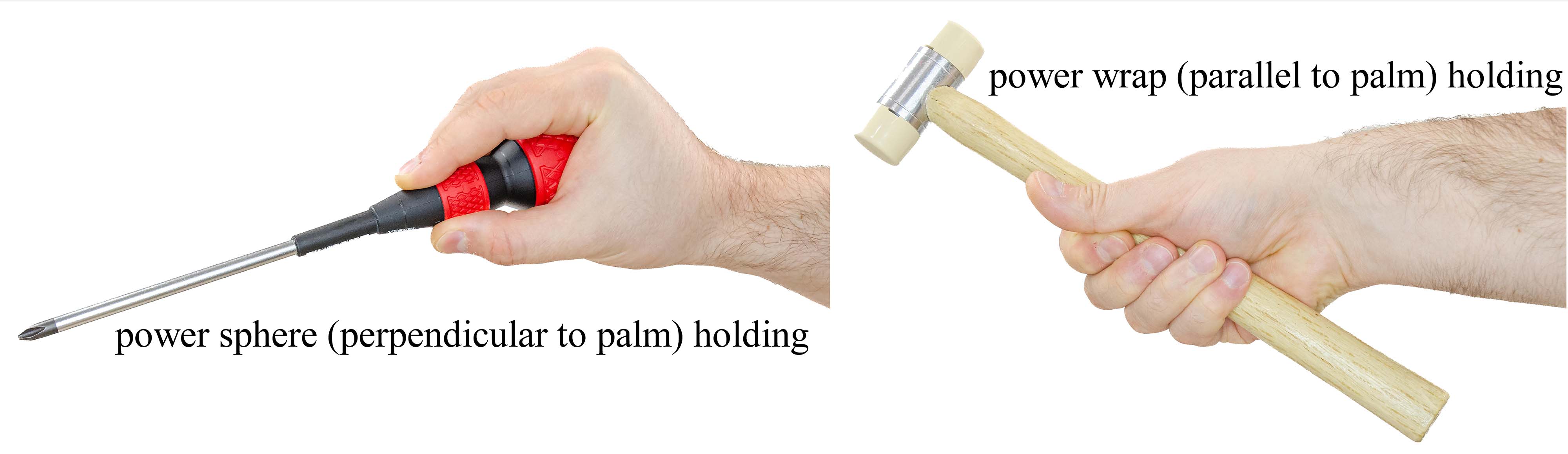}
         \vspace{-5.5mm}
        \caption{}
        \label{fig2_6}
	\end{subfigure}\\
        \vspace{-1mm}
\caption{Human hand with tools: (a) adaptability/motion of human palm and finger while holding/fixing a tool in the hand; (b) two main methods of grasping/holding tools (power sphere and wrap) during use (named based on the human hand grasping taxonomy~\cite{feix2015grasp}).}
\label{fig2_7}
\end{figure}

\noindent
The human hand is a remarkable adaptive end-effector of the arm for holding and using various tools~\cite{gustus2012human}. Many existing grippers for assembly and disassembly attempt to be inspired by the human hand. If looking at the performance of a human hand while holding and fixing a tool for use~(Fig.~\ref{fig2_5}), one can see that the palm wrap/adapts to the surface of the tool and, due to muscle tension, fixes the tool, preventing it from slipping~\cite{feix2015grasp, pozzi2023actuated}. 
Equally important is the use of fingertip adaptability, which increases contact area, frictional properties, and tool tip control~\cite{ringwald2023towards, spiers2018variable, wu2024bionic}. For example, when considering the process of disassembling an automotive air-conditioning unit (most parts and fasteners are plastic except for the main fan, tubes, and actuators). There are three main tools used: screwdrivers of various sizes and shapes (unscrewing and breaking connections), hammers (breaking plastic components), and drills (destroying connections that cannot be broken with other elements). The use of tools for various tasks of unscrewing bolts/screws and breaking connections involves several options for holding the tool~Fig.~\ref{fig2_6}: power sphere (perpendicular to the plane of the palm) and power wrap (parallel to the plane of the palm). Other scientific groups also hold the view on the importance of combined solutions~\cite{jeong2024bariflex, klas2021kit, borras2018kit} and different grasping methods~\cite{pagoli2021soft, harada2016proposal, fang2020learning} in the development of gripping devices for tool use~\cite{hoffmann2014adaptive, sun2016robotic}.

\subsection{Justification of Gripper Properties}
\noindent The design of an effective gripper for using tools (such as screwdrivers, hammers, and drills) must include several specific properties: \textit{1) Robustness} - adaptability of the gripper contact surfaces to various tool shapes and the ability to dampen excess forces and vibrations transmitted from the tool; \textit{2) Grasping versatility} - using active elements for tool fixation to ensure power holding, with the ability to adjust the position for precise tool handling; \textit{3) Task Versatility} - simple control methods for different strategies to enable rapid implementation. 


\subsubsection{Robustness} The gripper gains robustness through its \textit{ability to adapt and resist damage from interaction with disassembly environments}. There are many methods to achieve the robustness of the gripper, but it is always necessary to balance between a rigid mechanism and adaptability. Since, for disassembly applications, it is important to fix the tool in the gripper, there is a limit to the number of possible soft elements. In this case, it is feasible to achieve sufficient robustness by damping one of the interactive components and using force feedback to stop the robot in case of a collision. In our case, the gripper must provide lateral tool-tip force (20\% greater than the one that will result slipping of the fingertips during use of the tool, which will vary depending on its length) while maintaining the ability to control the tool-tip orientation (power sphere holding) and withstand the loads encountered when using a hammer (power wrap holding).


\subsubsection{Grasping versatility} The high gripper versatility aims to~\textit{grasp tools and parts in a wide range of sizes and shapes}.
Due to high dexterity, gripping devices that use an anthropomorphic design~\cite{grebenstein2011dlr, liu2023dexterous, gao2021dexterous} achieved greater versatility by taking the optimal position and orientation of the gripper contact surfaces (fingertips). However, such designs can cause problems with gripper control, even when there is an active palm~\cite{li2019novel, lee2021soft}. Therefore, the classic solution is to use a three-finger gripper~\cite{bunis2018caging, golan2020variable, geies2020grasping}, which allows both power sphere and wrap grasping. However, such grippers have a limited number of degrees of freedom (DoF) and suffer from the need to adapt to non-symmetrical surfaces~\cite{liu2020optimal}. This can be solved by providing passive adaptability of the fingertips and additional DoF for orienting the fingers in space. In our case, the gripper must provide a 100\% grasping success rate for more than half of each tool type and transmit torque (which will depend on the application - breaking, unscrewing, etc.) to them during use.

\subsubsection{Task Versatility} In the disassembly application, the robot must perform many different tasks: grasping, using a tool, breaking, holding, and fixing parts while working collaboratively with a human. The versatility of functions lies in the \textit{ability to provide not only the tools used but also other operations during human interaction}. These properties are achieved due to various parameters of the grasping system, such as the continuous and intelligent movement of the component, with force and position feedback. The versatility of the tasks is determined by the application that needs to be performed: grasping tools, their use, and breaking connections (fully automated disassembly); simple adaptive human control of robot-grasping elements (human-collaborative disassembly). In our case, the use of a gripper should reduce the time required to unscrew 3 screws and ensure the performance of a certain task when a human interacts with the proposed system.

\begin{figure*}[tb]
    \centering
    \includegraphics[width=0.69\linewidth,clip ,trim=0pt 0pt 0pt 0pt]{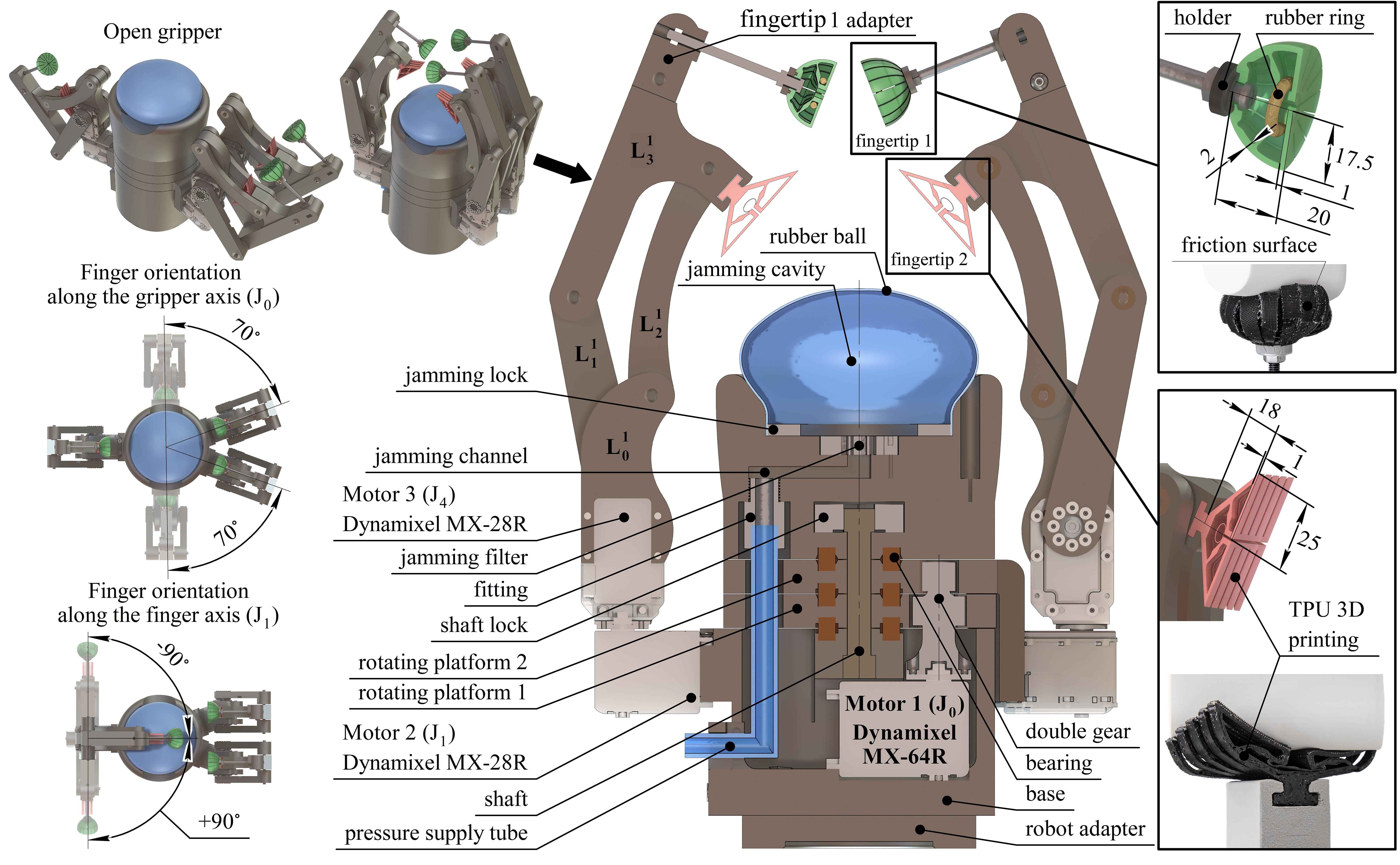}
    \includegraphics[width=0.3\linewidth,clip ,trim=0pt 0pt 0pt 0pt]{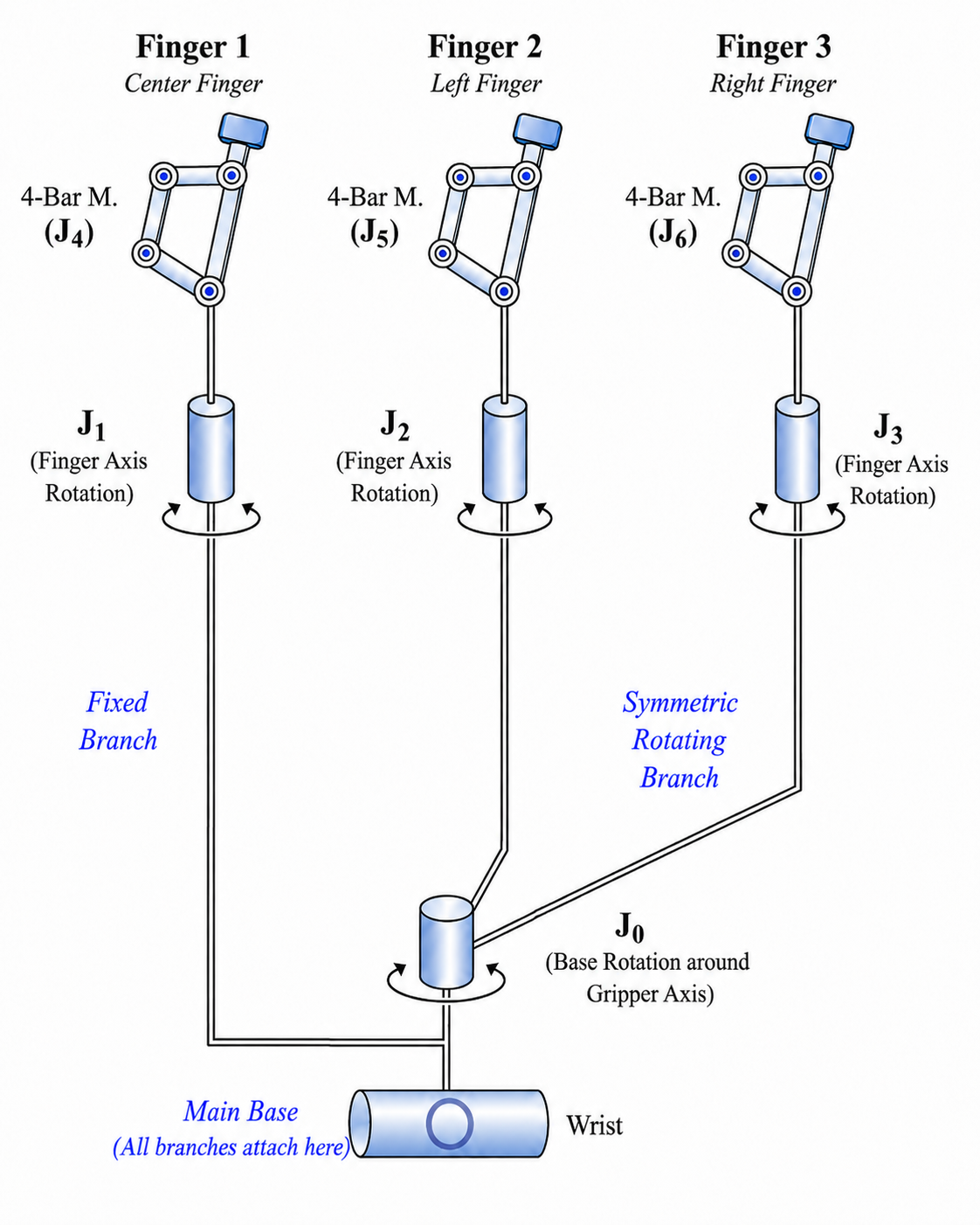}
    \vspace{-1.5mm}
    \caption{3-finger gripper design with jamming palm, where: (left, top to bottom) fully open fingers are shown, movement of fingers around the gripper axis, orientation of fingers around their axis; (center) section of the gripper along the axis of symmetry with designation of all structural elements; (right, top to bottom) structural overview of fingertip 1 and 2 (all dimensions in mm); and on right kinematic diagram of the ARTiS.}
    \label{fig_3}
\end{figure*}

\section{Design Concept}
\noindent
The design decisions made for the ARTiS Fig.~\ref{fig_3} are made in such a way as to ensure robustness and versatility (described above). To ensure these required parameters, the design included 7 motors, and a separate control box for versatile control of both electrical and pneumatic elements of the design. 

\subsection{Main Design}
\noindent To provide ARTiS with a wide working range of palm usage, it is necessary for the gripper's fingers to completely leave the working area of the jamming part (Open gripper Fig.~\ref{fig_3}), and for this purpose, the classic four-bar finger mechanism with two fingertips (1 and 2) is required. Thanks to classical kinematics, it is possible to solve the system of equations to find the position of the fingertip depending on the angle of rotation $0 \leq \theta_1 \leq 170$ of the driven (motor 3 - $J_4$ Dynamixel MX-28R in Fig.~\ref{fig_3}) link $L_1$:

\vspace{-4mm}
\begin{equation}
\label{deqn_ex1}
l_1 \begin{bmatrix}\cos{\theta_1}\\ \sin{\theta_1} \end{bmatrix} + l_3 \begin{bmatrix}\cos{\theta_3}\\ \sin{\theta_3} \end{bmatrix} = l_0 \begin{bmatrix}\cos{\theta_0}\\ \sin{\theta_0} \end{bmatrix} + l_2 \begin{bmatrix}\cos{\theta_2}\\ \sin{\theta_2} \end{bmatrix}, \\%
\end{equation}

\vspace{-4mm}
\begin{equation}
\label{deqn_ex1_1}
\vec{T} = l_1 \begin{bmatrix}\cos{\theta_1}\\ \sin{\theta_1} \end{bmatrix} + l_3 \begin{bmatrix}\cos{\theta_3}\\ \sin{\theta_3} \end{bmatrix} + r_T \begin{bmatrix}\cos{\theta_3 + \theta_T}\\ \sin{\theta_3 + \theta_T} \end{bmatrix},
\end{equation}

\noindent where $l_0$, $l_1$, $l_2$, $l_3$ - length of the corresponding links, $\theta_0$, $\theta_1$, $\theta_2$, $\theta_3$ - rotation angles of the corresponding links, $T(r_T, \theta_T)$ - fingertip coordinates relative to the end joint of the link $L_3$.

Each gripper finger is equipped with an orientation motor (motor 2 - $J_1$ Dynamixel MX-28R in Fig.~\ref{fig_3}) that changes the orientation of the finger around the finger-axis $J_1$ in the range from -90 to 90 degrees. This degree of freedom for each finger allows ARTiS to provide active orientation of the tool tip during power sphere grasping and increases the adaptability of fingertip positioning during power wrap grasping. One of the fingers is fixed to the base of the gripper, and the other two are driven by motor (motor 1 - $J_0$ Dynamixel MX-64R in Fig.~\ref{fig_3}), which, through a double gear transmission (1:4), transmits movement to rotary platforms 1 and 2. The rotary platforms are driven to symmetrically change the orientation of the fingers relative to the gripper axis from 0 to 70 degrees. With this design, one finger has 2 DoF, and the other two have 3 DoF, which is enough to provide versatility in choosing a grasping finger configuration.

A jamming palm (rubber material, diameter 100 mm, height 60 mm, jamming elements - glass beads FGB-20~\cite{9562008}) is installed on top of the gripper in a cup-shaped cavity. A perpendicular cylindrical slot with a diameter of 60~mm is made in the cup-shaped cavity, which allows deepening the tools into the palm that are lying on the table for the power wrap (parallel to the palm) grasping. A filter is installed at the bottom of the cup through which air enters and is sucked out (jamming channel, fitting, and pressure supply tube), changing the density of the palm while fixing the tools. The design uses flat bearings with a diameter of 50 mm for the interaction of the base and the upper part of the jamming palm. It provides high robustness and allows the palm to interact with the tools with high force.

\begin{figure}[!t]
    \centering
    \includegraphics[width=1\linewidth, clip, trim=0pt 50pt 0pt 50pt]{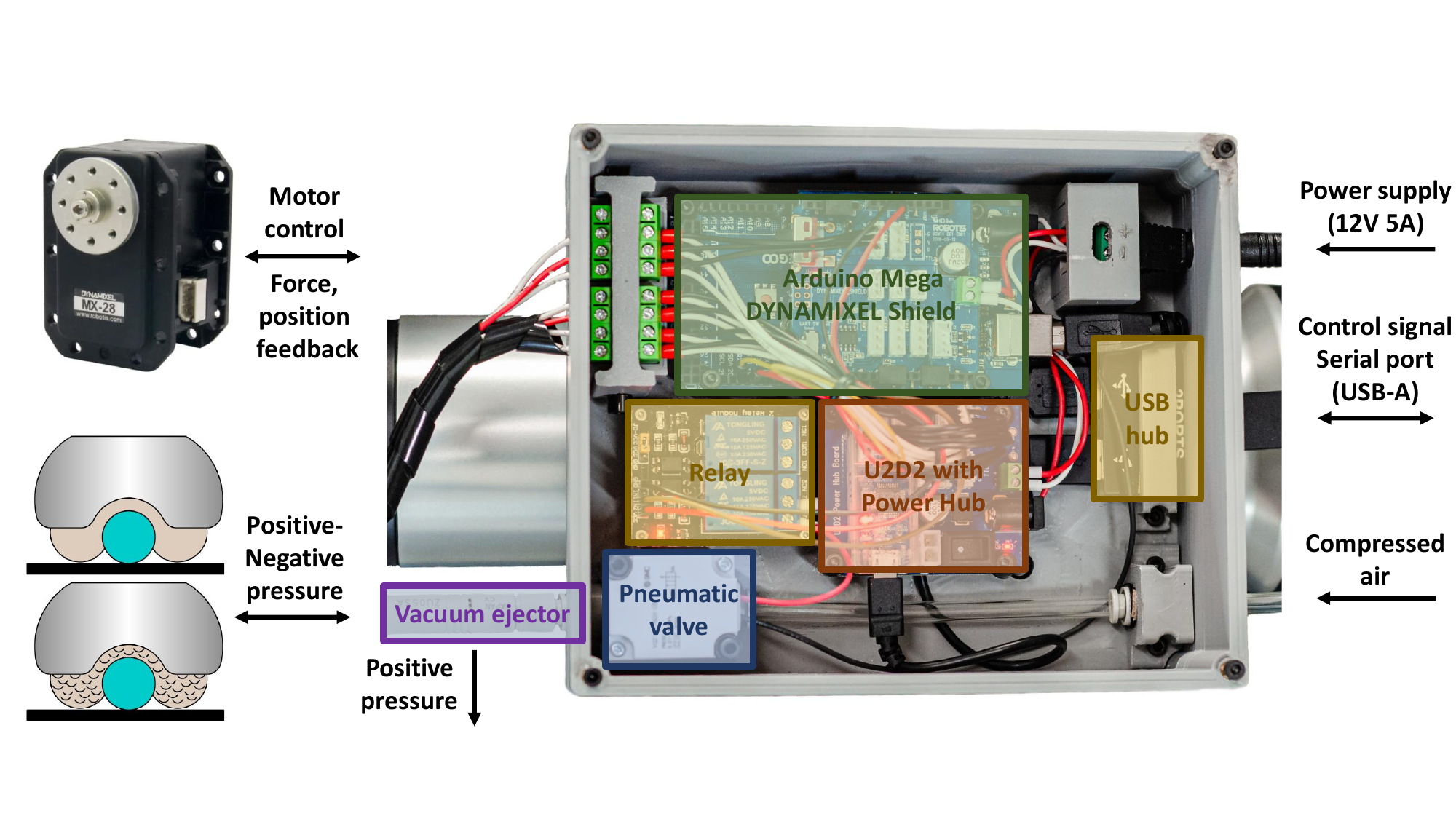}
    \vspace{-1mm}
    \caption{Diagram of the ARTiS control box with input and output parameters.}
    \label{fig_4}
\end{figure}

To control the gripping device, a control box (Fig.~\ref{fig_4}) has been developed to which compressed air (6 mm tube, 0.1-0.9 MPa), electrical power (12V, 5A), and a control signal through a serial port (USB-A) are supplied. For all subsequent experiments, compressed air is supplied at a pressure of 400 kPa, and after opening the pneumatic valve enters the in-line vacuum ejector (ZU05SA SMC), which produces a manufacturing-standard negative pressure of -88 kPa in the palm, and when the pneumatic valve is closed, the vacuum ejector returns the pressure in the palm to atmospheric. The motors are controlled from the control box through TTL, and the stiffness of the palm is controlled by increasing and decreasing the pressure by opening or closing the valve that supplies compressed air to the ejector. Since the design of the control box uses both the U2D2 controller and Arduino with DYNAMIXEL Shield, which are connected in parallel, it is possible to hard program the gripper only via Arduino, as well as perform adaptive control via U2D2 with Arduino (only for relay control). The general specifications of the ARTiS gripping device and API are presented in the project page~\cite{Project_Artis}; the control algorithm is presented in Appendix~\ref{A2}.


\subsection{Highly Adaptive Fingertips}
\begin{figure}[!t]
\centering
 	\begin{subfigure}[b]{0.86\linewidth}
         \centering
         \vspace{2mm}
    \includegraphics[width=1\linewidth, clip, trim=0pt 0pt 0pt 0pt]{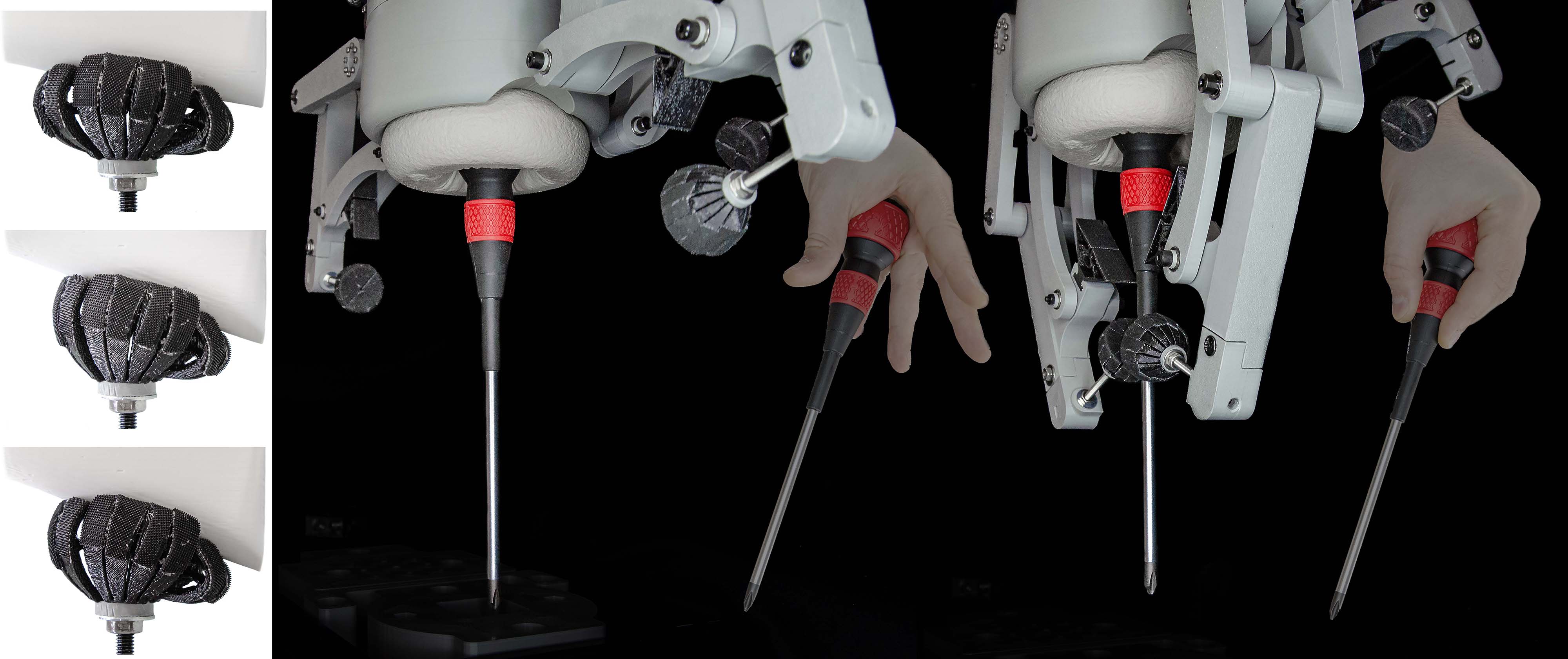}
        \vspace{-5.5mm}
        \caption{}
        \label{fig5_1}
    \end{subfigure} \\
  	\begin{subfigure}[b]{0.86\linewidth}
         \centering
         \vspace{1.5mm}
    \includegraphics[width=1\linewidth, clip, trim=0pt 0pt 0pt 0pt]{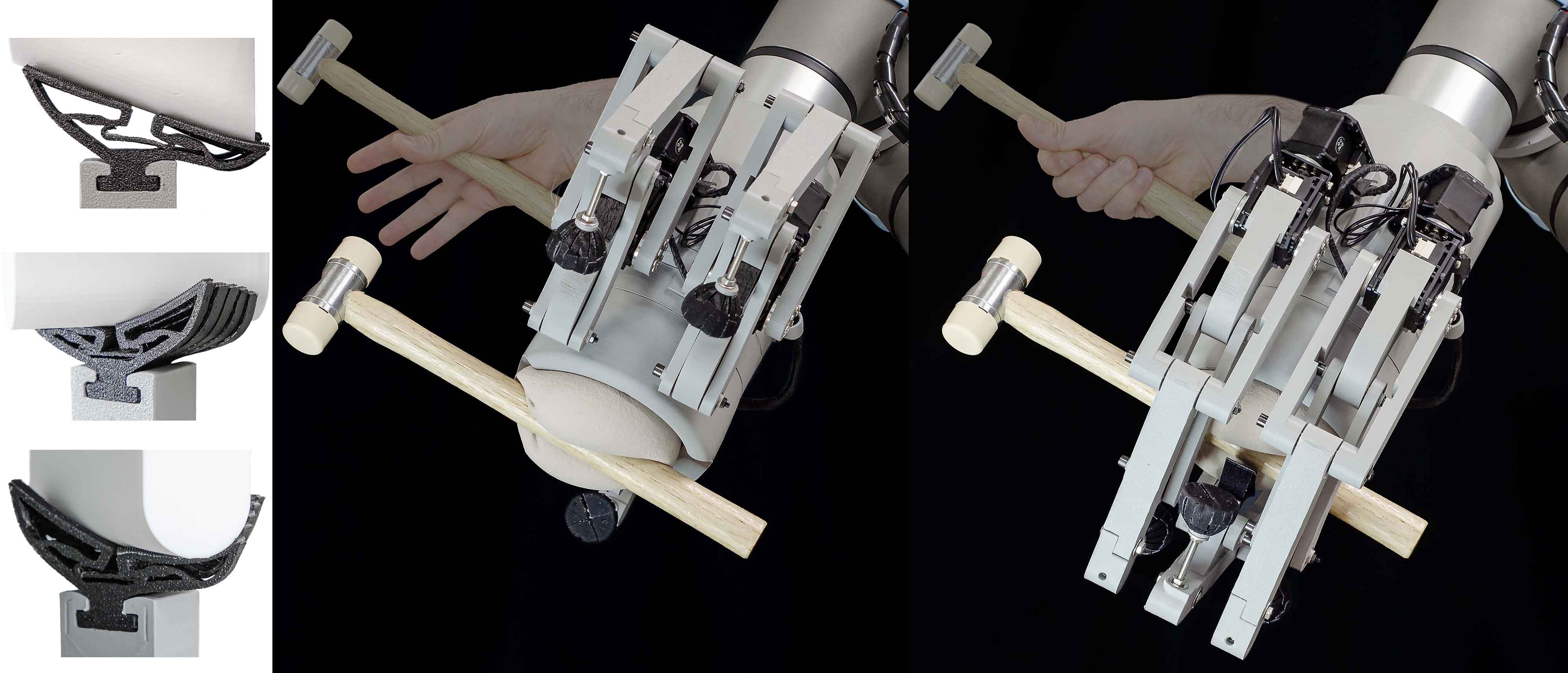}
        \vspace{-5.5mm}
        \caption{}
        \label{fig5_2}
	\end{subfigure}\\
        \vspace{-1.5mm}
\caption{Power grasping and holding tools: (a) sphere using fingertip 1; (b) wrap using fingertip 2.}
\label{fig5}
\end{figure}

\noindent To obtain high adaptability to different tool and part geometries, the novel TPU-95A 3D-printed fingertips 1 and 2 (Fig.~\ref{fig_3}) were developed. The design was inspired by human fingers in their ability to actively adapt to the orientation and geometry of the tools (Fig.~\ref{fig2_5}). Two proposed fingertips were based on the Fin-Ray finger design~\cite{10802217}. Fin-Ray finger design allows for finger adaptability in one plane, which is successful only for symmetrical objects. Instead, when performing a power sphere grasping tool (Fig.~\ref{fig5_1}), the human hand uses a fingertip (end of a finger) that can adapt to the surface of the tool in two planes. 

In addition, it is important that with different orientations of the tool and the fingertip, they should adapt to the surface as much as possible with minimal force. To ensure minimal force for adapting to the tool's geometry and changing orientation, it was decided to make one of the two fixing parts of the Fin-Ray move freely (released) and combine four such structures to obtain a round shape. As a result, fingertip 1 (Fig.~\ref {fig_3}) was obtained, which consists of a rigid holder with a bolt to which four TPU-95A 3D printing elements are attached and are additionally connected to each other with a rubber ring. Thanks to the rubber ring that unites the released fixing part of the fingertip elements, an effect was achieved in which the deformation of one of the fingertip elements by the tool leads to a change in the orientation of all other fingertip elements. 

A rubber friction surface (friction tape TB641\footnote{3M™ Gripping Material TB641 available [Online]: \url{https://www.3m.com/3M/en_US/p/d/b40069682/}}) is glued to the outer surface of the soft elements to ensure minimal slipping during deformation. A round-form fingertip (inspired by the human fingertip) is used, which provides both adaptability to the surface and the orientation of the grasping tools. 
During power wrap grasping, the results indicate that the human hand uses a fingertip (middle finger Fig.~\ref{fig5_2}) that usually deforms and takes the form of a semi-cylindrical tool at different orientations. Also, each finger has the ability to orient relative to the finger axis, which prompts the task of adapting to semi-cylindrical objects with different fingertip orientations. For this purpose, fingertip 2 (Fig.~\ref{fig_3}) was developed, which has oppositely deployed small Fin-Rays with five elements, connected by a thin (0.3 mm) ellipse and, like the previous finger, covered with a friction surface. The presented fingertip 2 design allows for easy adaptation to the orientation of the object with minimal force loads (left top Fig.~\ref{fig5_2}). When changing the orientation of the finger, the fingertip in two folds provides maximum contact area with the tool (left middle and left bottom Fig.~\ref{fig5_2}).

\section{Results}
\noindent In this section, the performance of the developed gripper in disassembly and collaborative applications is evaluated. The experiments are designed to address the following research questions: (1) Is the gripper sufficiently adaptive and robust? (2) How versatile is the gripper in grasping and holding various tools? (3) How effectively can the gripper perform a range of tasks? To address these questions, the gripper's capability to utilize different tools and holding methods (Fig.~\ref{fig5}) is evaluated during the disassembly of an automotive air-conditioning unit.

\subsection{Evaluating Robustness}
\noindent The robustness of the ARTiS gripping device is ensured by the holding method (a combination of palm and finger), adaptability, and the force absorption of fingertip 1 (sphere holding Fig.~\ref{fig5_1}) and 2 (wrap holding Fig.~\ref{fig5_2}).

\subsubsection{Fingertip} 
\begin{figure}[!t]
\centering
    \includegraphics[width=0.775\linewidth, clip, trim=0pt 0pt 5pt 63pt]{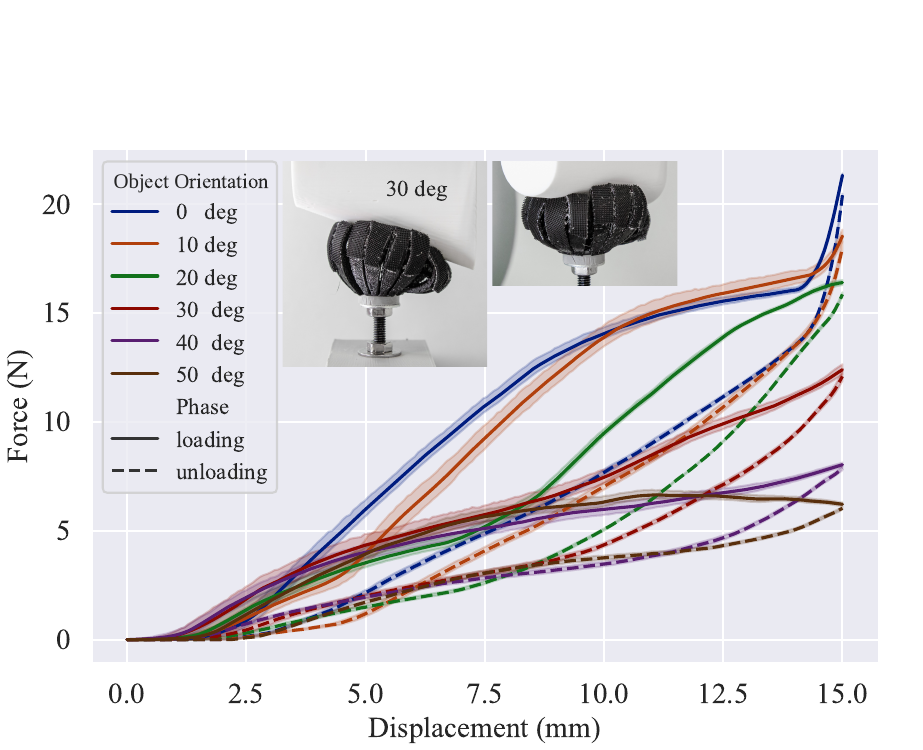}
    \vspace{-3mm}
    \includegraphics[width=0.21\linewidth, clip, trim=0pt 10pt 110pt 0pt]{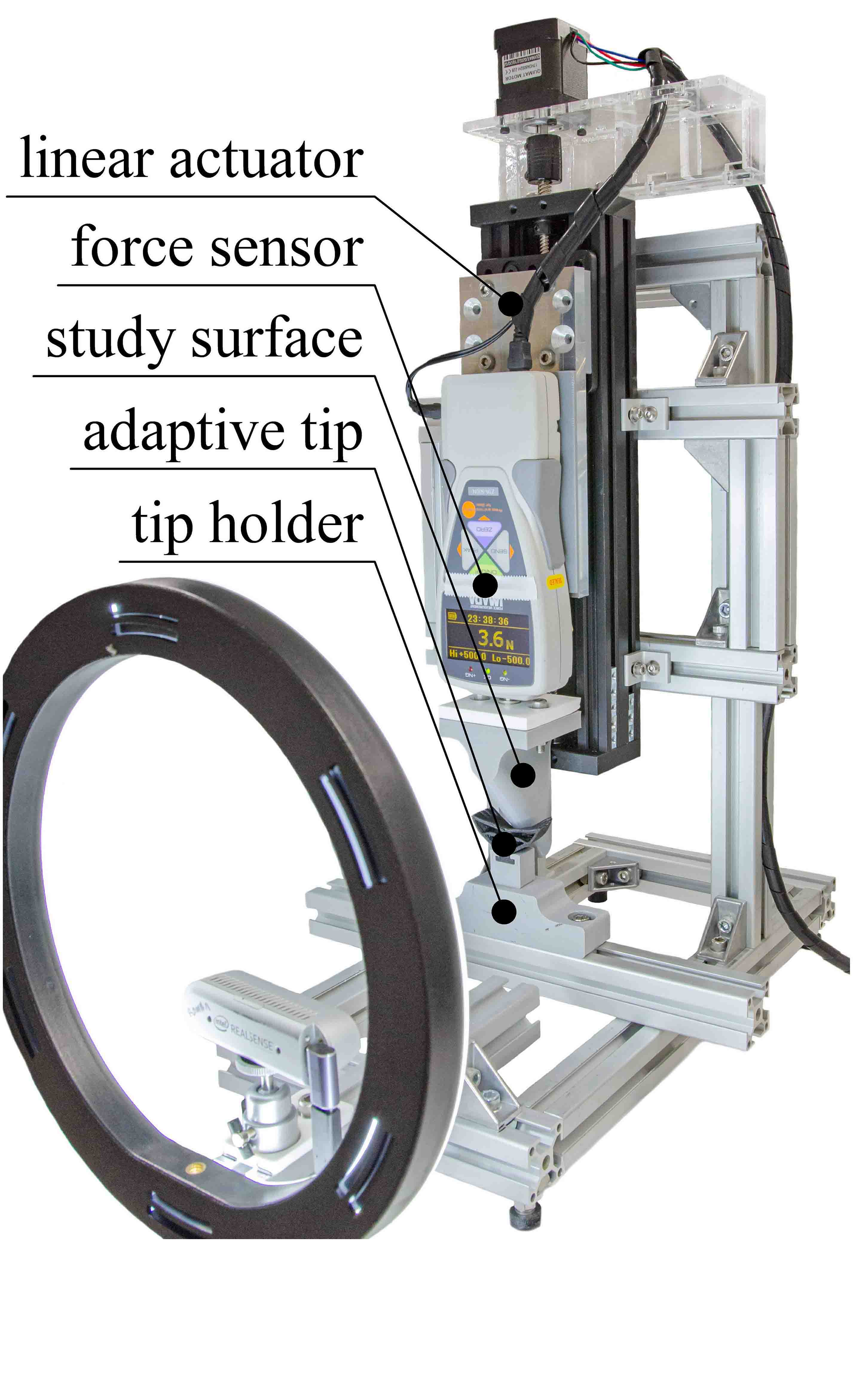}
    \vspace{-2mm}
\caption{Experimental assessment of robustness and adaptability of fingertips. Left - average hysteresis of adaptive fingertip 1 at different object orientations. Right - experimental setup for studying the loading and unloading of the fingertips with different object orientations.}
\label{fig_6}
\end{figure}

For robustness evaluations, it is necessary to analyze the dispersion of the fingertip-object interaction force over multiple iterations. Similarly, for fingertip adaptability evaluations, it is necessary to analyze how the fingertip-object interaction force changes when its orientation changes. One of the available methods for obtaining data on the dispersion and changes in the fingertip-object interaction force is to obtain the average hysteresis of many trials, indicating the dispersion zones. For this purpose, an experimental setup was built (Fig.~\ref{fig_6} - right), consisting of a fixed adaptive fingertip on which objects are pressed at different orientations (from 0 to 50 degrees) while simultaneously obtaining data on displacement and force. To ensure the reliability and robustness of the obtained data, each experiment is performed 30 times, and the average value is displayed with the data fluctuation zone of the same color.

Fingertip 1 is completely symmetrical, so it exhibited hysteresis when changing the object orientation and the pressure at a distance of 15 mm (Fig.~\ref{fig_6} - left). With 30 tests for each object orientation (0, 10, 20, 30, 40, and 50 degrees), a total of 180 experiments were conducted, and the dispersion relative to the average force value did not exceed 0.5 N.  After conducting 180 experiments with the object oriented in one direction, a constant dispersion was obtained throughout the measurement range during loading. During the unloading of the fingertip, dispersion is absent for all experiments. This indicates constant force absorption of fingertip 1 and a high robustness of the design. Analyzing the obtained force distributions (Fig.~\ref{fig_6} - left), seen that the maximum force decreases by an average of 22\% when the angle of object orientation increases compared to the previous one. This tendency is observed until the active surface of the fingertip touches the tip holder, after which the finger behaves as rigid. A uniform decrease in the interaction force allows us to state that the fingertip reorientation process occurs during loading, resulting in maximum adaptability. This, in turn, leads to an increase in the area of contact between the fingertip and the object, and accordingly the friction force between them.

\begin{figure}[!t]
\centering
 	\begin{subfigure}[b]{0.84\linewidth}
         \centering
    \includegraphics[width=1\linewidth, clip, trim=0pt 0pt 5pt 65pt]{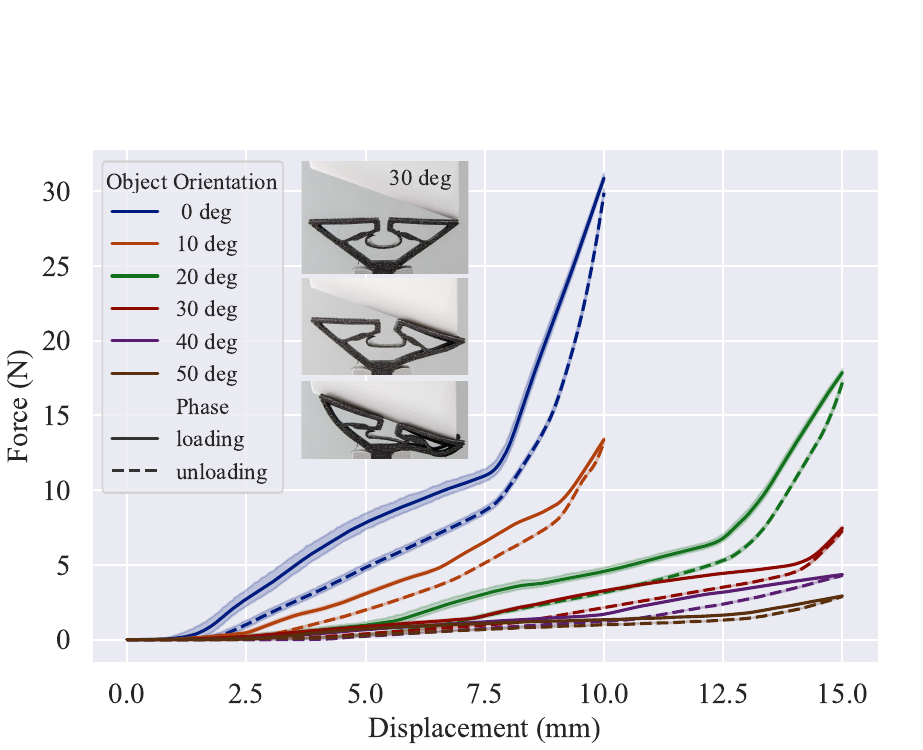}
        \vspace{-5.5mm}
        \caption{}
        \label{subfig7_1} 
    \end{subfigure} 
  	\begin{subfigure}[b]{0.84\linewidth}
         \centering
    \includegraphics[width=1\linewidth, clip, trim=0pt 0pt 5pt 65pt]{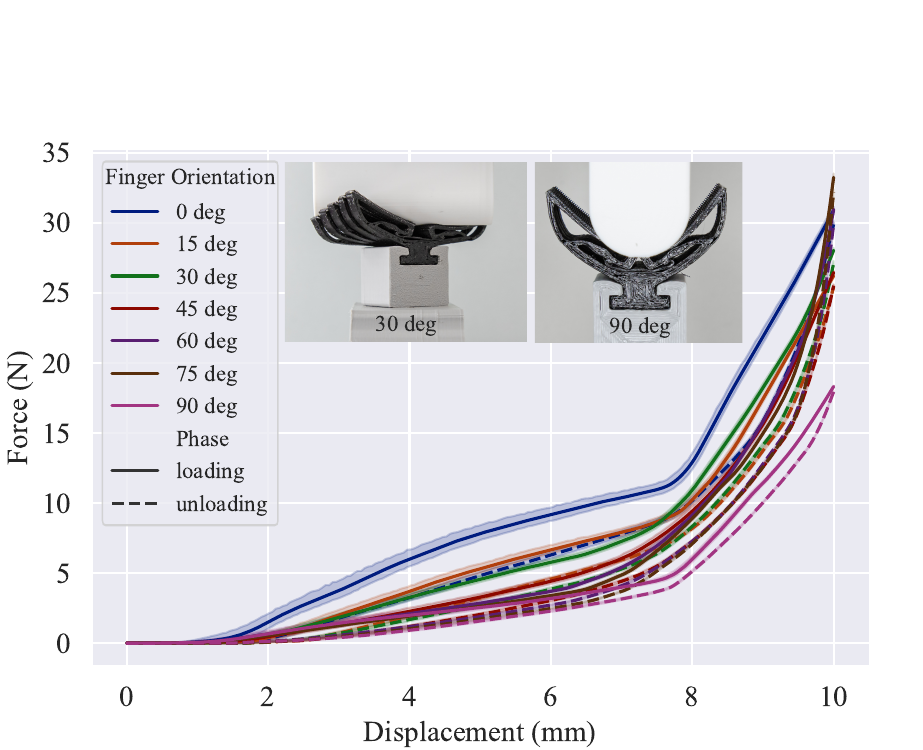}
         \vspace{-5.5mm}
        \caption{}
        \label{subfig7_2}
	\end{subfigure}\\
        \vspace{-1mm}
 \caption{Experimental assessment of the robustness of fingertip 2: (a) hysteresis when changing object orientation (finger orientation 0 degrees); (b) hysteresis when changing fingertip orientation (object orientation 0 degrees).}
 \label{fig_7}
\end{figure}

Since fingertip~2 is not symmetrical due to using seven-cylindrical objects in power grasping, the hysteresis (displacements of 10 and 15~mm) was obtained when changing both the object and the fingertip orientation (Fig.~\ref{fig_7}). By analogy with fingertip~1, fingertip~2 - with changing the object orientation (Fig.~\ref{subfig7_1}), the dispersion relative to the average force value did not exceed 0.5~N. Since the height of fingertip~2 is smaller than that of fingertip~1, the data show a rapid increase in force when the active surface reaches contact with the fingertip holder (after 10~N for 0 and 10~degrees), which is why the experiment was limited for 0 and 10 degrees to 10~mm. For all other object orientation angles, results confirm significant adaptability, whereby connecting the two parts of the fingertip increases the contact area by two times. In the case of changing the fingertip orientation from 0 to 90 degrees in 15-degree increments (Fig.~\ref{subfig7_2}), dispersion relative to the average force value did not change. All studied fingertip orientation angles show practically the same characteristics, and the only ones that differ are the orientations of 0 and 90 degrees. This is due to the fact that they occupy the ultimate orientation, while at an angle of 0~degrees from the very beginning, the displacement has the largest contact area and the corresponding largest forces, and at an angle of 90~degrees, it has the smallest contact area and force. However, even at the smallest possible contact area, a sufficient force of 18~N is provided at a displacement of 10~mm. This shows that fingertips~1 and 2 used in ARTiS provide high robustness (hundreds of uses without loss of performance) and can be used to hold instruments of various shapes by adapting to their orientation and surface shape.

\begin{figure}[!t]
\centering
 	\begin{subfigure}[b]{0.6\linewidth}
         \centering
    \includegraphics[width=1\linewidth, clip, trim=0pt 0pt 0pt 0pt]{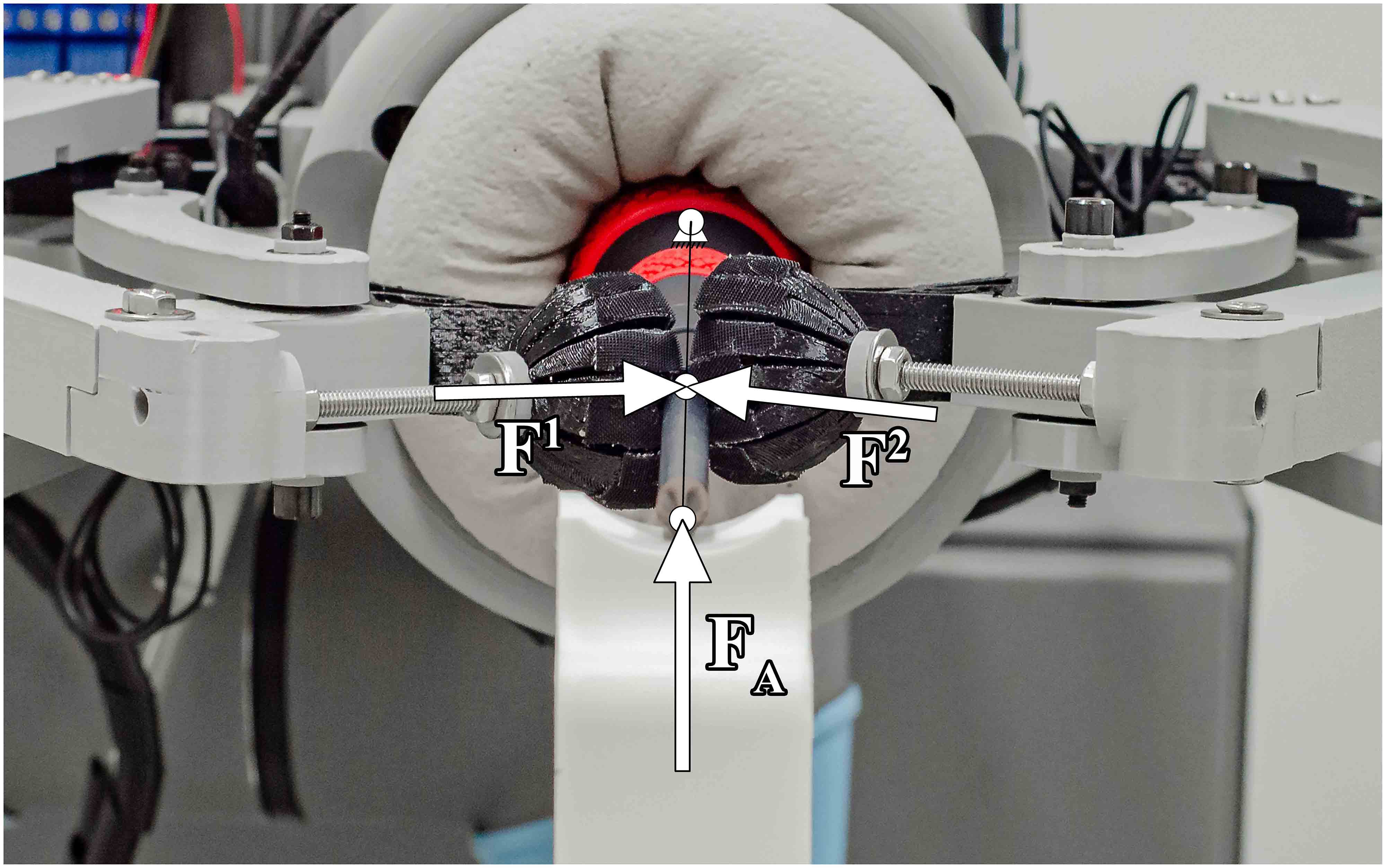}
        \vspace{-5.5mm}
        \caption{}
        \label{subfig7_3} 
    \end{subfigure} 
        \begin{subfigure}[b]{0.385\linewidth}
         \centering
    \includegraphics[width=1\linewidth, clip, trim=0pt 0pt 0pt 0pt]{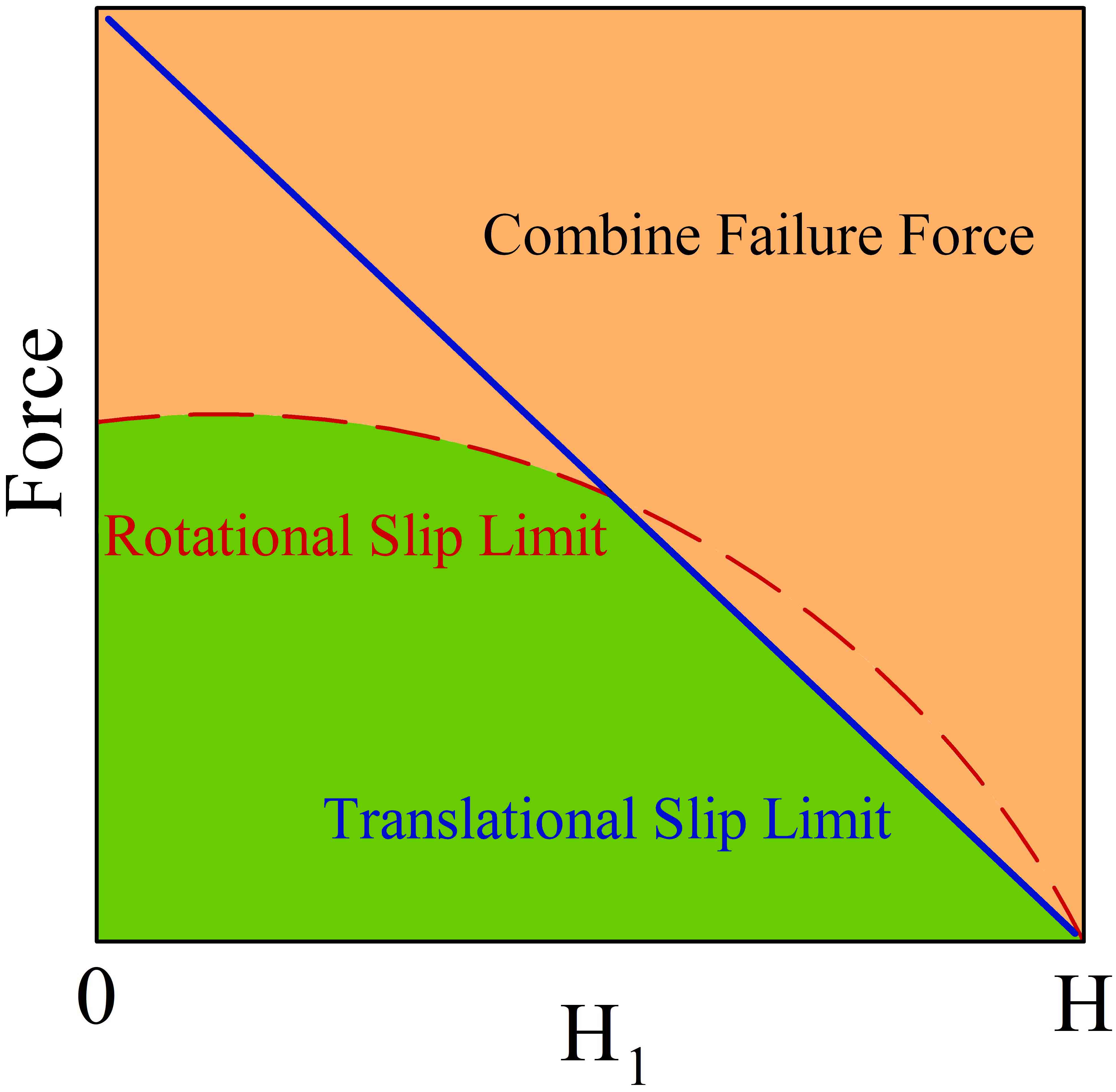}
        \vspace{-5.5mm}
        \caption{}
        \label{subfig7_4} 
    \end{subfigure} 
  	\begin{subfigure}[b]{1\linewidth}
         \centering
    \includegraphics[width=0.73\linewidth, clip, trim=0pt 0pt 0pt 0pt]{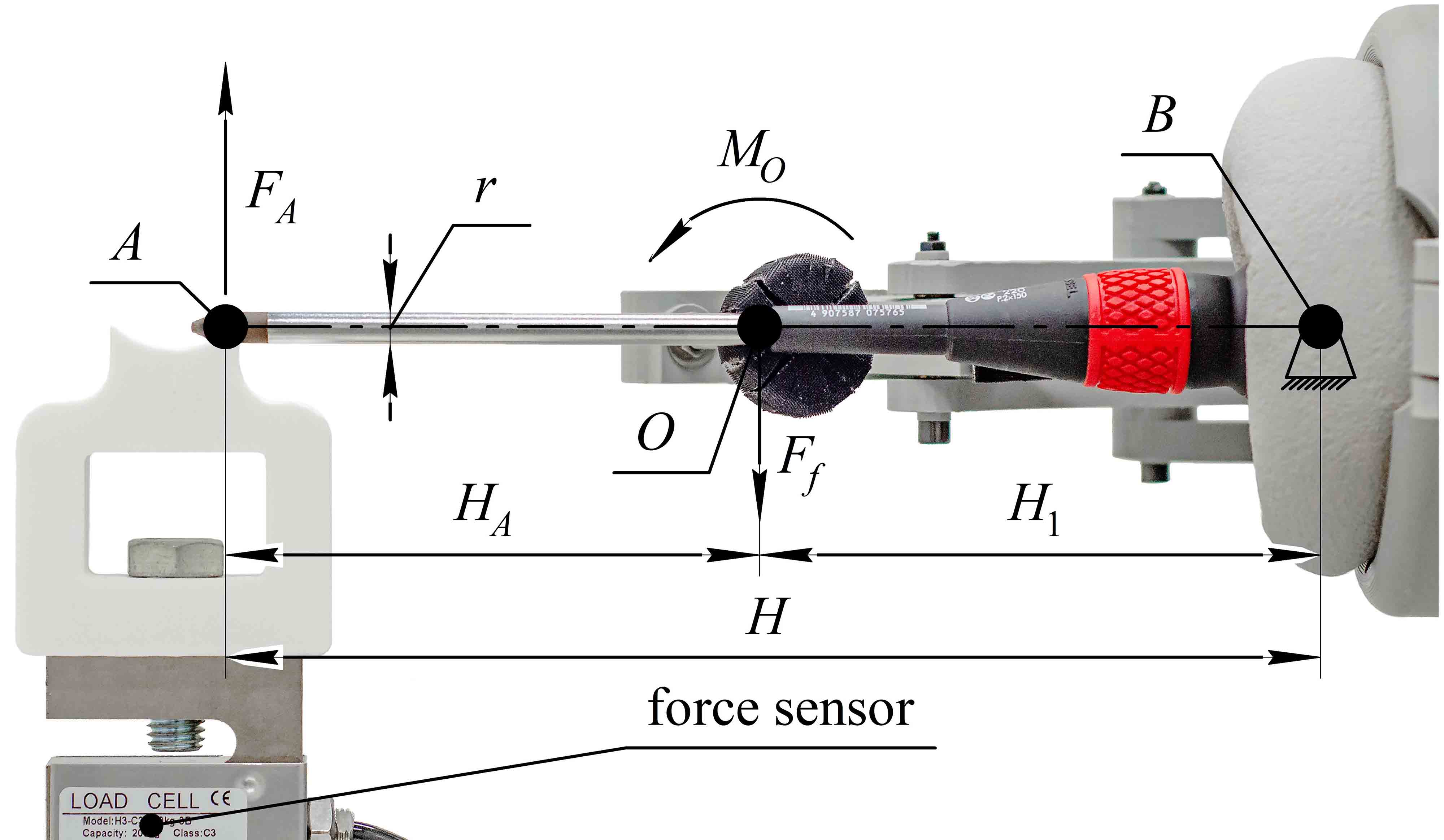}
    \includegraphics[width=0.23\linewidth, clip, trim=0pt 0pt 0pt 0pt]{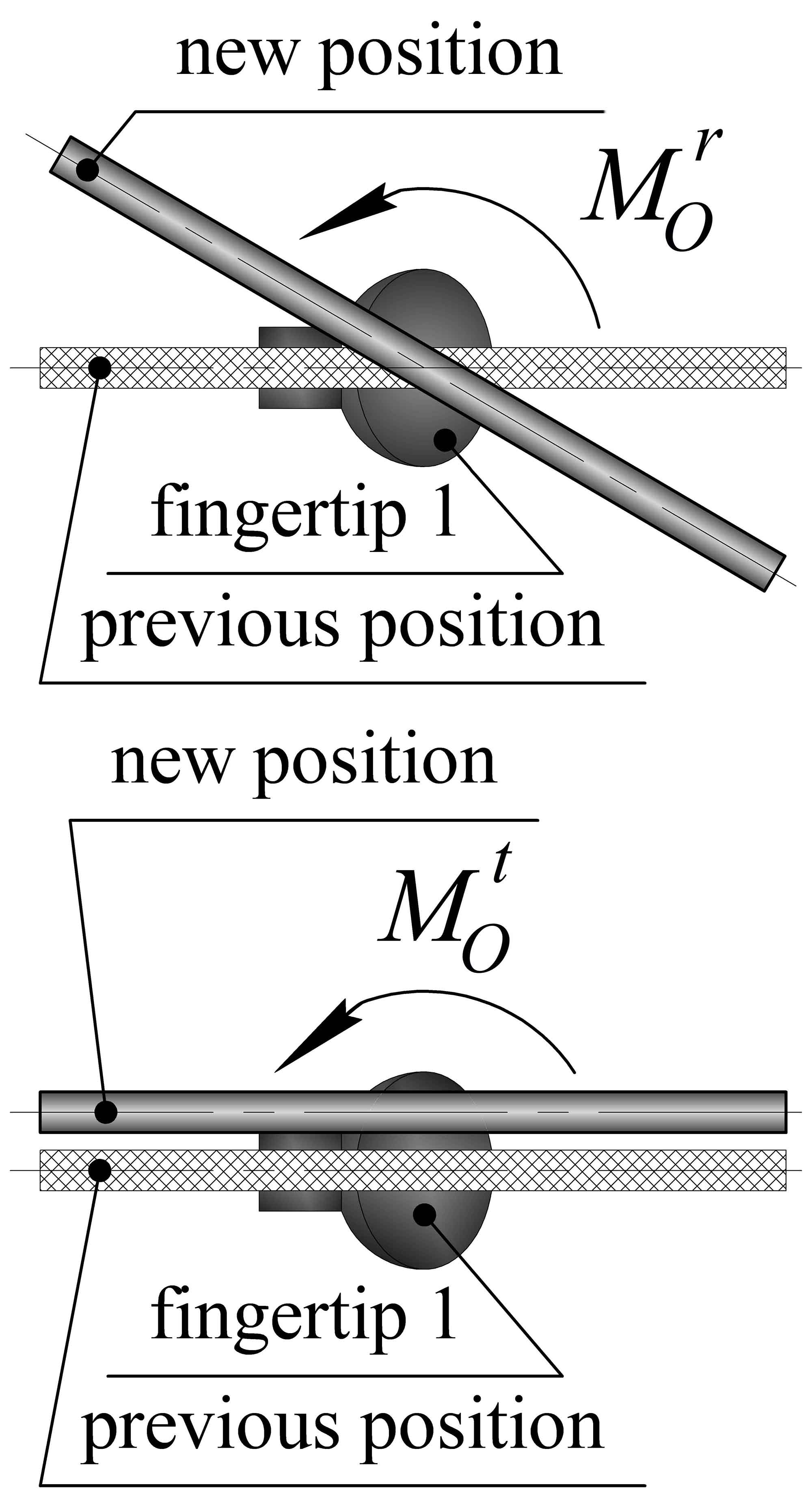}
         \vspace{-2.5mm}
        \caption{}
        \label{subfig7_5}
	\end{subfigure}\\
        \vspace{-1.5mm}
 \caption{Holding a screwdriver by a gripper when interacting with an object: (a) the most advantageous position that can lead to the screwdriver slipping in the fingers; (b)  demonstration of the schematic value of the force at the beginning of slipping; (c) slip calculation diagram.}
 \label{fig_71}
\end{figure}

\subsubsection{Holding/Force Study} When using a tool held by a gripper, the robustness and stability of its holding are important, and they can be disturbed by interaction with the external environment. The most unfavorable moment occurred when grasping the power sphere method with two fingers; the force applied to the tip of the tool is directed perpendicular to the grasping plane with two fingers (Fig.~\ref{subfig7_3}). Using this method, screwdrivers can be held such that, with sufficient applied force, they can slip out of the fingers, causing an inability to work correctly. This interaction is reduced to a problem similar to a lever, where a tool with a length of $H$ and a radius of $r$ plays the role of a lever that is fixed at one end (at point B using the jamming palm). At a certain distance $H_1$ from point B, it is clamped by two fingers (at point O) with a force $F_f$, forming a friction force and coefficient of friction $\mu$ (Fig.~\ref{subfig7_5}). Perpendicular to the tool at the tool tip - at point A (distance from the fingers $H_A=H-H_1$), a force is applied (measured using a force sensor $F_A$ when the tool moves perpendicular to the sensor), leading to the loss of contact between the fingers and the tool. When force $F_A$ is applied at point A, a torque $M_A = F_A \cdot H_A = F_A \cdot (H-H_1)$ arises. At the same time, depending on the size of the tool and the position of the finger ($H_1$), the clamped fingers (at point O) provide translational ($M_O^t = \mu \cdot F_f \cdot H_1$) and rotational ($M_O^r = \mu \cdot F_f \cdot r$) resistances. Depending on the system parameters, it will lead to the tool slipping parallel to the fingers (translational) or scrolling between the fingers (rotational). Slippage/scrolling occurs when $M_A \geq M_O$, respectively:

\vspace{-4mm}
\begin{align}
\label{deqn_ex2}
M_A \geq M_O^{t} \Rightarrow F_A \cdot (L-L_1) \geq \mu\cdot F_f \cdot H_1, \\%
M_A \geq M_O^{r} \Rightarrow F_A \cdot (L-L_1) \geq \mu\cdot F_f \cdot r,
\end{align}

\noindent and by calculating the equations for $F_A$ taking into account the conditions for both translational and rotational resistance, it is possible to obtain the overall resistance as the minimum of the two:

\vspace{-4mm}
\begin{equation}
\label{deqn_ex3}
F_A=\min \bigg(\frac{\mu\cdot F_f \cdot H_1}{H-H_1},\frac{\mu\cdot F_f \cdot r}{H-H_1}\bigg)\\%
\end{equation}

If presenting a graph of the beginning of the tool slipping depending on the distance ($H_1$) from the palm to the location of grasping the tool with the fingers at point O (Fig.~\ref{subfig7_4}), obtain:
\begin{itemize}
    \item When grasping the tool with the fingers closer to the palm ($H_1 \rightarrow 0$), rotational slip prevails due to an increase in the torque arm.
    \item When grasping the tool with the fingers closer to the tip of the tool ($H_1 \rightarrow H$), translation prevails and becomes the dominant factor.
\end{itemize}

\begin{figure}[!t]
\centering
 	\begin{subfigure}[b]{0.605\linewidth}
         \centering
         \includegraphics[width=1\linewidth, clip, trim=0pt 0pt 0pt 0pt]{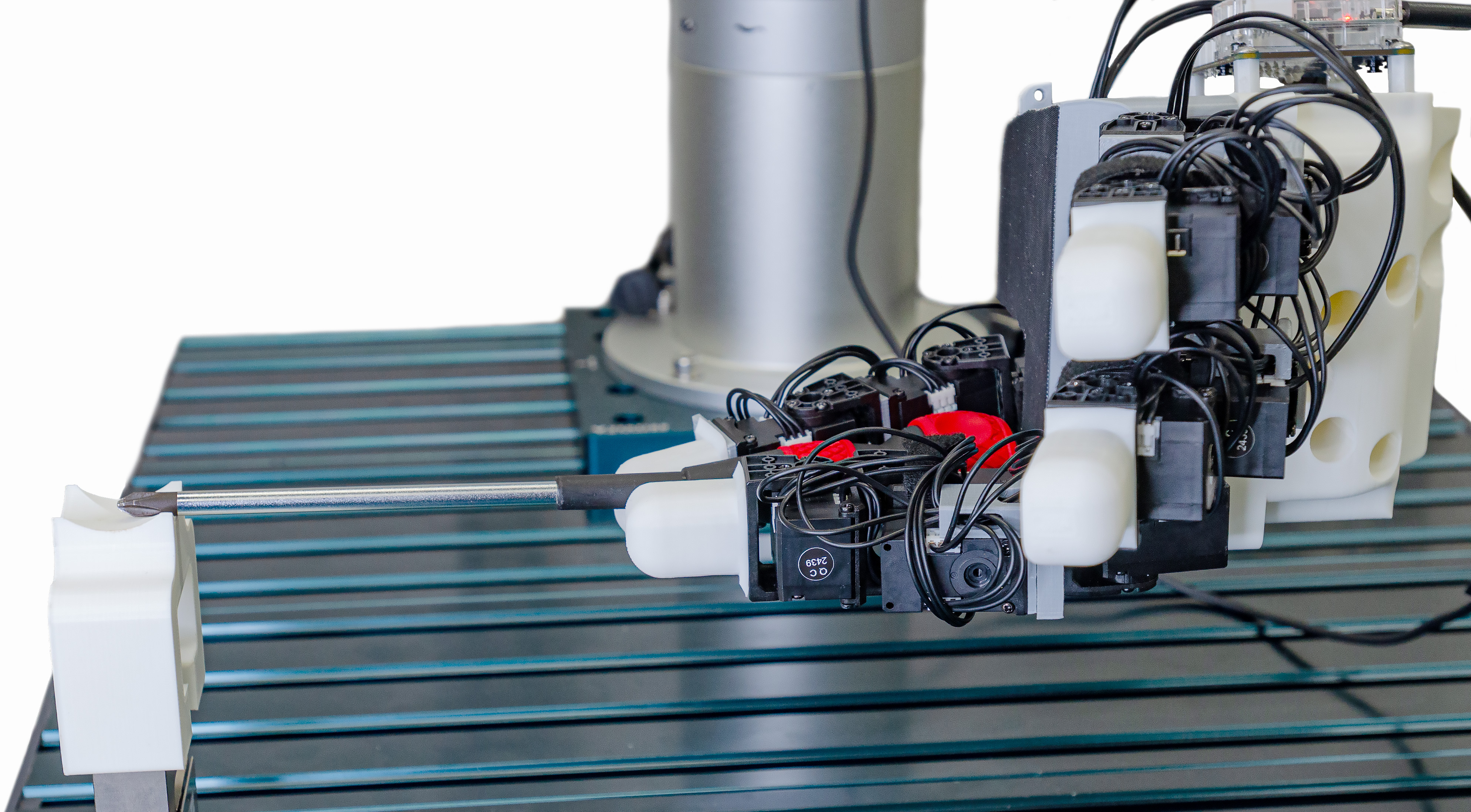}    
        \caption{}
        \label{subfig_leap_2} 
    \end{subfigure} 
  	\begin{subfigure}[b]{0.375\linewidth}
         \centering
         \includegraphics[width=1\linewidth, clip, trim=0pt 0pt 0pt 0pt]{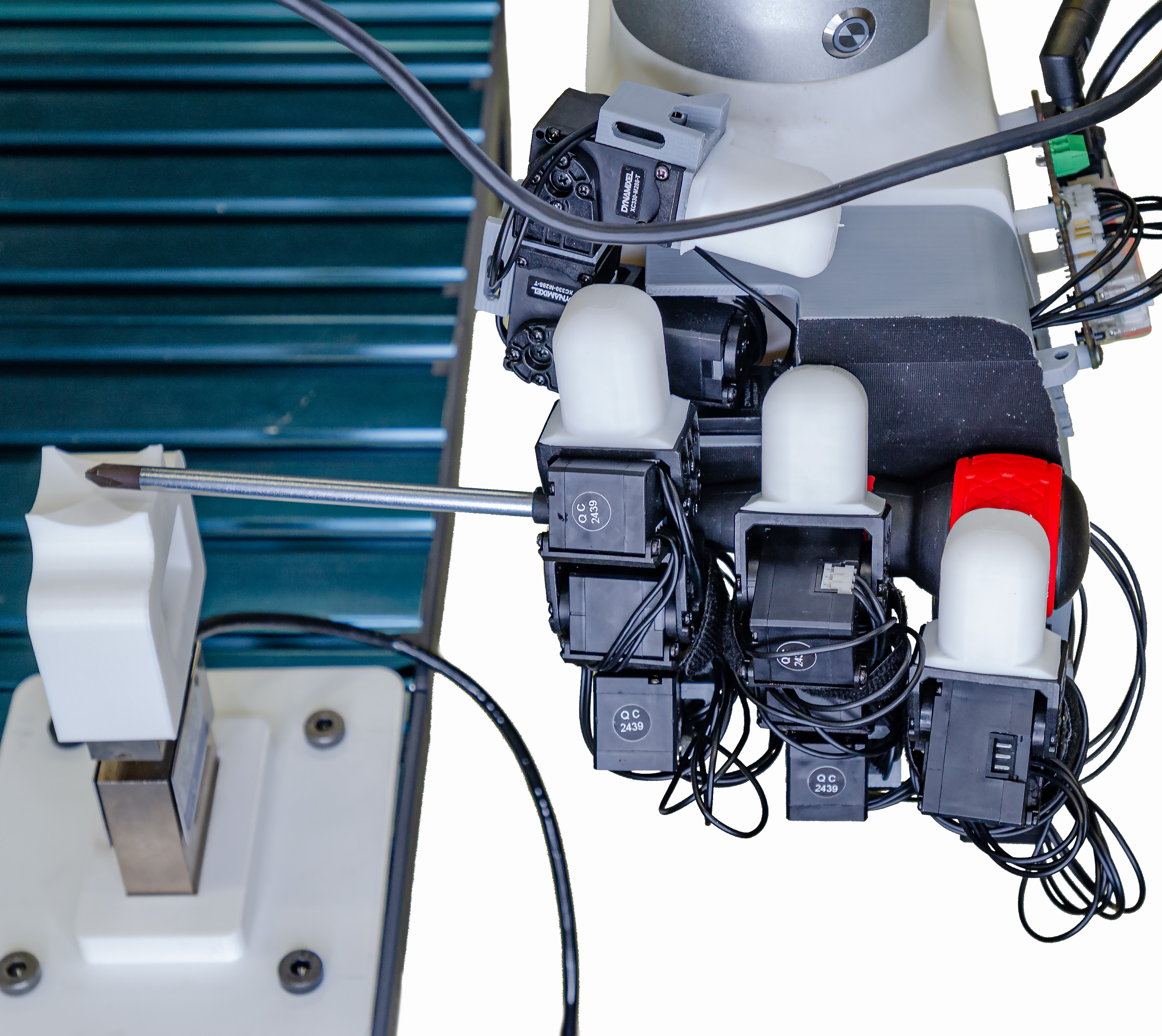}         
        \caption{}
        \label{subfig_leap27}
	\end{subfigure}\\
        \vspace{-1.5mm}
 \caption{Screwdrivers grasping methods with LEAP Hand gripper~\cite{Shaw2023LEAPHL}: (a) with 2 fingers; (b) with 3 fingers.}
 \label{fig_Leap}
\end{figure}

\begin{figure}[!t]
\centering
 	\begin{subfigure}[b]{0.90\linewidth}
         \centering
         \vspace{-4.5mm}    
         \includegraphics[width=0.8\linewidth, clip, trim=0pt 0pt 0pt 0pt]{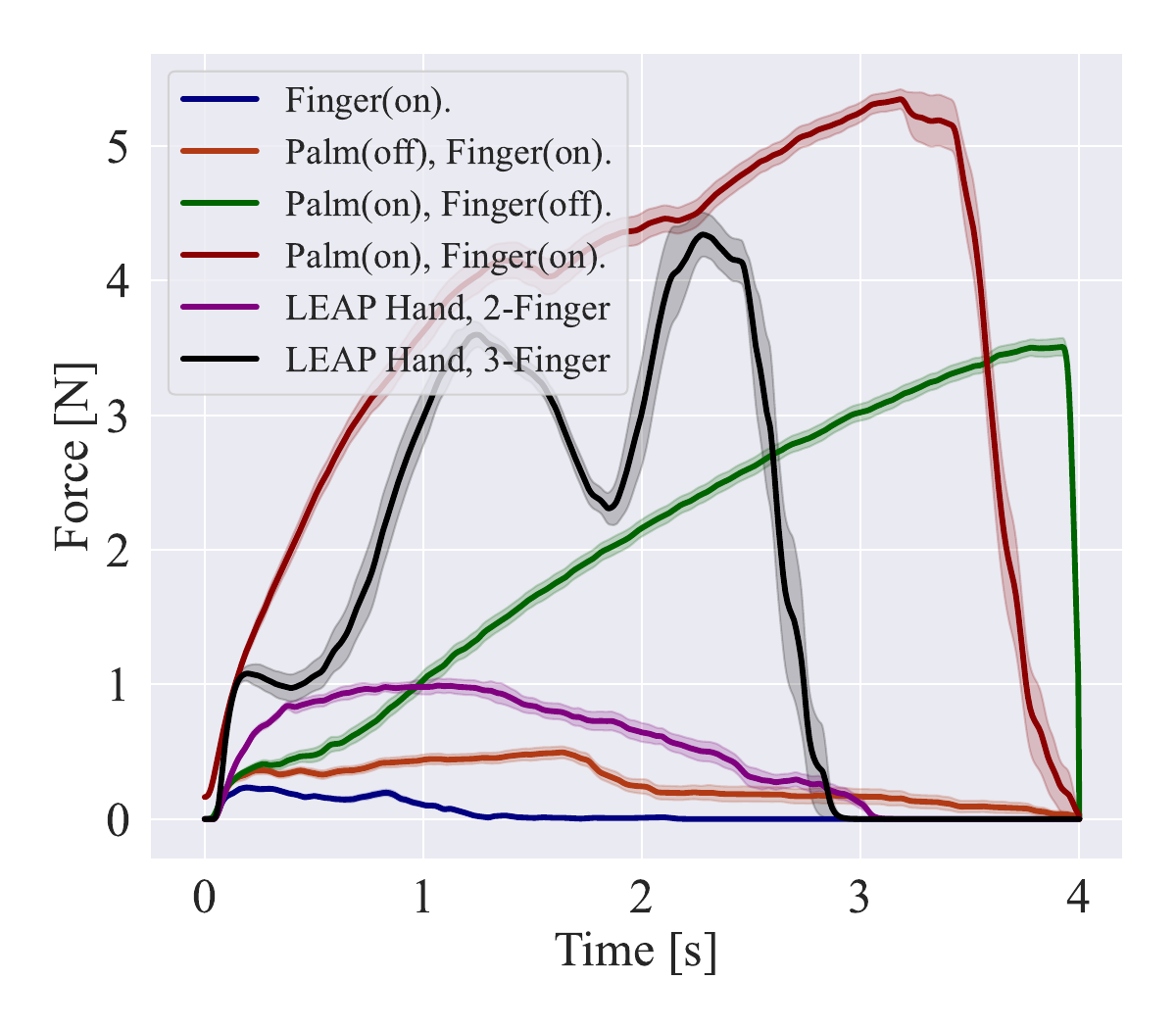}    \includegraphics[width=0.17\linewidth, clip, trim=0pt 0pt 0pt 0pt]{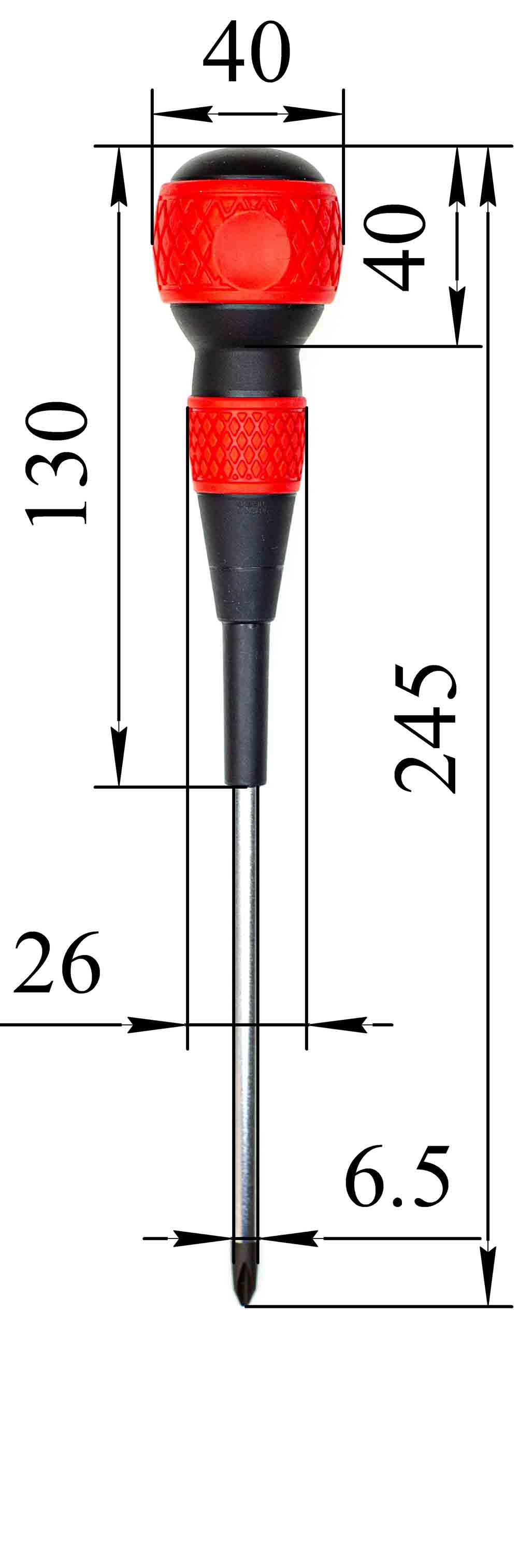}
        \vspace{-4.5mm}
        \caption{}
        \label{subfig7_6} 
    \end{subfigure} \\
  	\begin{subfigure}[b]{0.90\linewidth}
         \centering
         \includegraphics[width=0.8\linewidth, clip, trim=0pt 0pt 0pt 20pt]{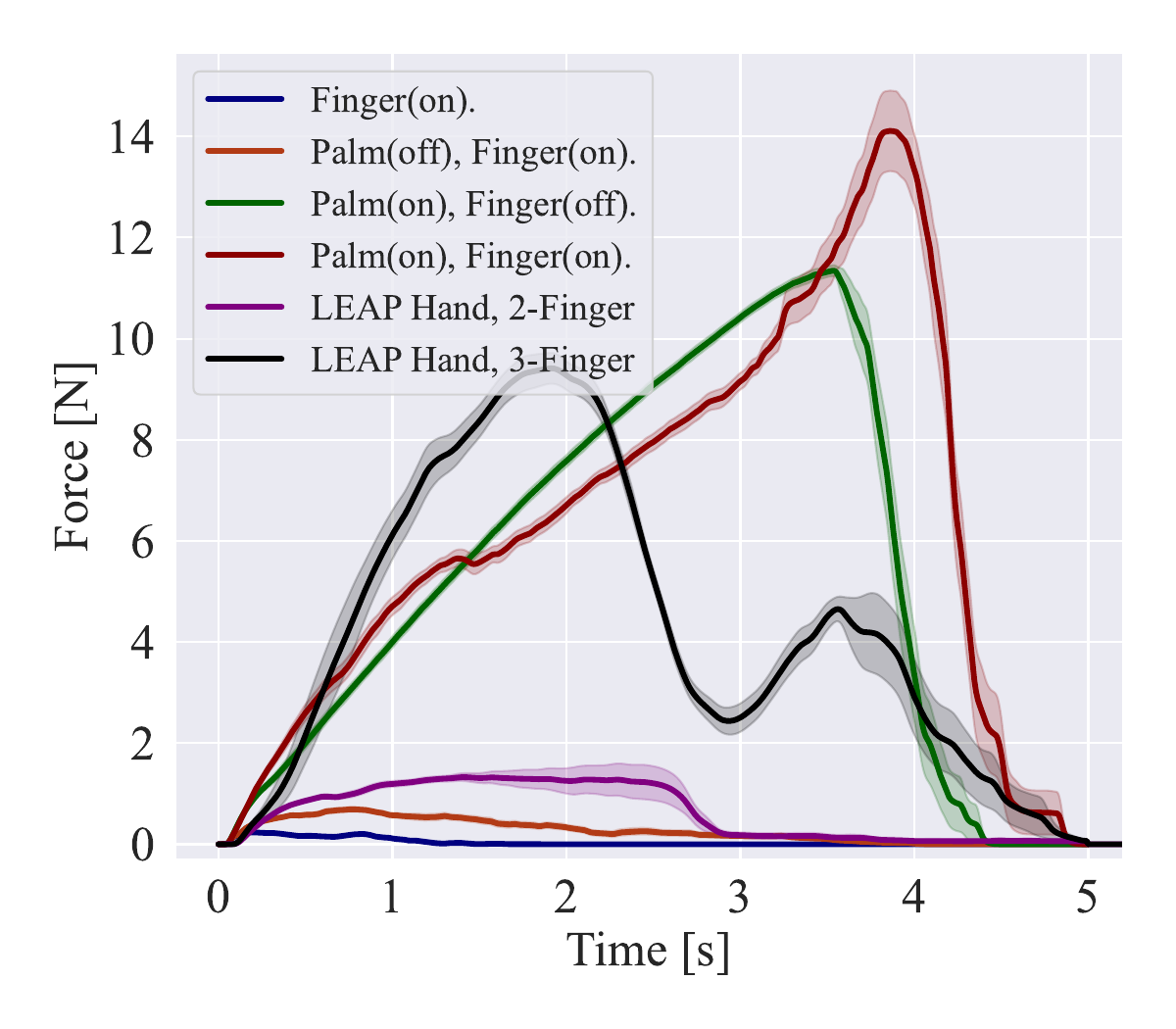}         \includegraphics[width=0.17\linewidth, clip, trim=0pt 0pt 0pt 0pt]{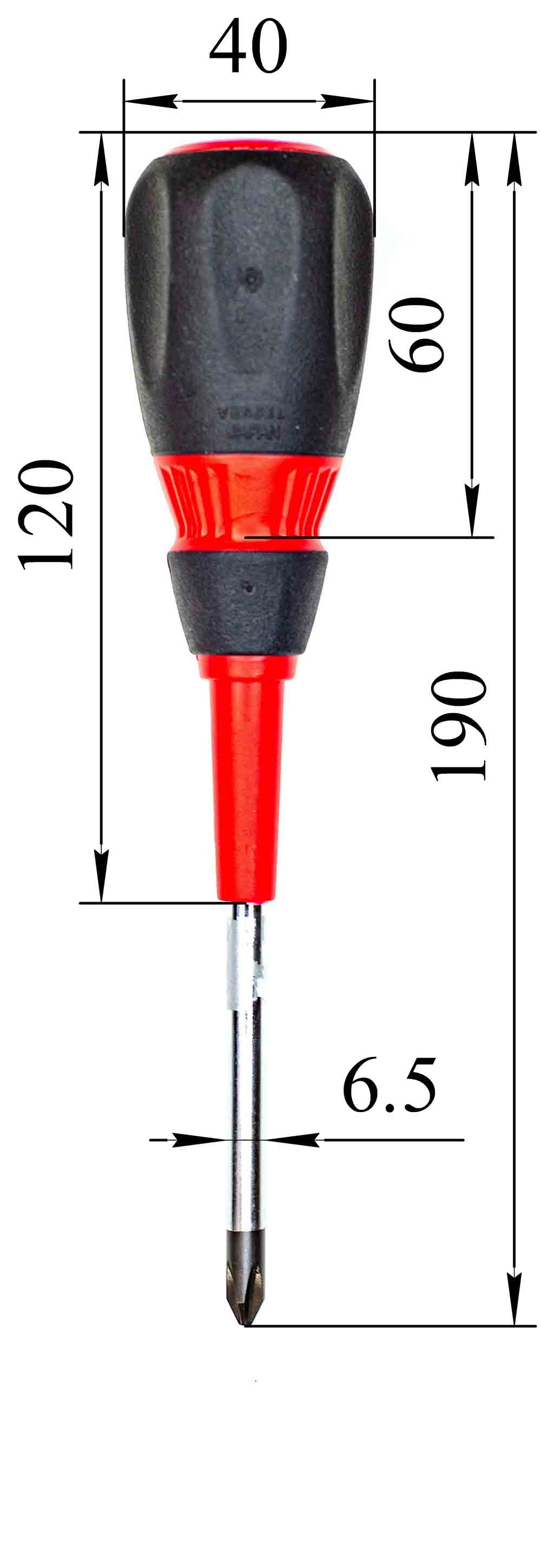}
         \vspace{-4.5mm}
        \caption{}
        \label{subfig7_7}
	\end{subfigure}\\
        \vspace{-1.5mm}
 \caption{Force applied perpendicular to the tool tip during ARTiS gripper holding by four different methods, LEAP Hand holding by two methods, and two screwdrivers (all dimensions in mm): (a) screwdriver \textnumero 2; (b) screwdriver \textnumero 4.}
 \label{fig_72}
\end{figure}

This scheme of holding the tool is identical to that when a human holds a screwdriver with one hand (Fig.~\ref{fig2_7}) or with two hands (fixing the screwdriver with one hand at point B and clamping with the other hand fingers at point O) and is the most effective in terms of robustness. To demonstrate the influence of the method of holding the tool with a gripper on the slipping force, an experiment was conducted according to the scheme (Fig.~\ref{subfig7_5}). The LEAP Hand gripper~\cite{Shaw2023LEAPHL}, which has a similar component cost and is also open access, was used as a baseline (Fig.~\ref{fig_Leap}). For this experiment, screwdrivers of two different shapes (spherical and conical) were used (Fig.~\ref{fig_72}), with a detailed description of all tools on the project page~\cite{Project_Artis}. When grasping (each method test was repeated 30 times to obtain an average force value): 
\begin{itemize}
    \item using only fingers - Finger(on);
    \item the palm is present but excluded; only fingers are actively used - Palm(off), Finger(on);
    \item with the use of the palm to actively fix the tool and without the use of fingers - Palm(on), Finger(off);
    \item with the use of the palm to actively fix the tool and with the use of fingers - Palm(on), Finger(on);
    \item LEAP Hand using 2 fingers perpendicularly pressing against the palm with friction coating (sphere) -  LEAP Hand, 2-Fingers;
    \item LEAP Hand using 3 fingers parallel pressing against the palm with friction coating (wrap) - LEAP Hand, 3-Fingers.
\end{itemize}

The results obtained for different screwdriver holding methods using ARTIS demonstrate (Fig.~\ref{fig_72}) that the highest robustness is achieved when the gripper uses a combination of palm and finger. The visible jumps in the force characteristics of the palm-finger holding method are due to the design of the fingertip, which consists of flexible ribbed elements. These jumps occur when contact with one rib is lost, and the transition to another occurs. Such jumps are characteristic of translation slip, while for the other two holding methods, where fingers are used, rotation slip without jumps is evident. A more detailed quantitative analysis of the impact of combinations of finger and palm use (palm and finger contributions in each failure mode, the governing mode of slip, and ablation-based safety factors) is presented in Appendix~\ref{A1}. From the statistical results, it was found that, on average, more than 90\% of the contributions to holding screwdrivers are from the palm contribution.

When comparing the results of the force that can be applied to the tip of the tool for ARTiS and the LEAP gripper during sphere (two-finger) holding confirm that for both screwdrivers, ARTiS shows a greater force of 5.4 and 7.8 times. Even when using the LEAP Hand with a power grasp (wrap - 3 fingers), it is possible to achieve 20.4\% and 32.8\% more force for ARTiS. At the same time, with the LEAP Hand, it loses the ability to control the tip of the tool, whereas for ARTiS, it is preserved. Therefore, the combined configuration of the active palm with adaptive fingers provides more robustness than predicted by isolated experiments, indicating a nonlinear benefit.

\begin{figure*}[!t]
    \centering
    \includegraphics[width=1\linewidth,clip ,trim=0pt 0pt 0pt 0pt]{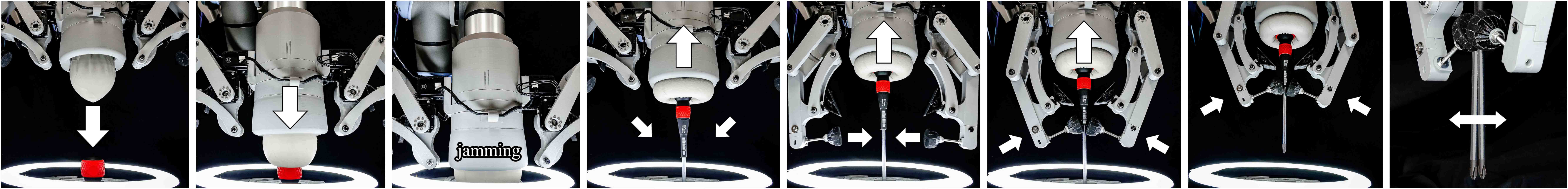}
    \vspace{-1.5mm}
    \caption{The process of precisely grasping screwdriver \textnumero 2 from the tool holder: palm pushing, jamming, lifting the tool, gripping it with two fingers, reorienting (using) the tool.}
    \label{fig_8}
\end{figure*}


\subsection{Evaluating Grasping Versatility}

\noindent To test the ability of ARTiS to grasp and hold various tools (versatility) during disassembly, three of the most widely used tools were selected (detailed specifications can be found on the project page~\cite{Project_Artis}): a screwdriver (9 types), a hammer (2 types), and a drill (2 types). Compared to other tests for object grasping for pick-and-place operations, during disassembly, it is important not only to determine whether the grasping is successful but also to assess whether it is possible to reorient/use the tool and then release it in the original location. To do this, an ARTiS gripping device with a control box was attached to a collaborative 6 DoF robot arm UR-5e (Fig.~\ref{fig1}), which is installed at a 45-degree angle to the worktable to which the tool holders are attached. The precise grasping consists of fixing the tool handle in the jamming palm and grasping the other end of the tool with fingertip~1 (Fig.~\ref{fig_8}). The first stage is jamming: the fingers are fully opened; then, moving vertically downwards, the palm is deformed to the shape of the tool. Upon reaching the maximum depth, air is sucked out of the palm, which increases the density of the palm and fixes the tool. Subsequently, the palm is lifted together with the fixed tool, and in the process, the fingers begin to close and fix the other part of the tool with the help of fingertip~1. After full grasping, depending on the need, the tool can be reoriented by changing the orientation of the fingers, or the tool can be used directly. Consequently, the tool is released in the reverse order of the grasping process. In the case of power grasping, everything happens identically to precise grasping; fingertip~2 is used, and the tool is located on the table, not in the tool holder.

\begin{table}[t]
\centering
\vspace{-3mm}
\caption{Scoring Table for Grasping Assessment}
\label{tab2}
\begin{threeparttable}
\scriptsize
\resizebox{\columnwidth}{!}{%
\begin{tabular}{cccccccc}
\hline
\textbf{Tool} &
\textbf{No.} &
\makecell{\textbf{Size}\\\textbf{(mm)}} &
\makecell{\textbf{Weight}\\\textbf{(kg)}} &
\makecell{\textbf{Jamming}\\\textbf{+ Lifting}} &
\makecell{\textbf{Grasping}\\\textbf{+ Releasing}} &
\makecell{\textbf{Grasping +}\\\textbf{Reor./Using}\\\textbf{+ Releasing}} &
\makecell{\textbf{Total}\\\textbf{success}\\\textbf{rate (\%)}} \\
\hline
\multirow{9}{*}{Screwdriver}
& 1 & $36\times36$ & 0.148 & 30/30 & 30/30 & 27/30 & 96.7 \\
& 2 & $\varnothing~40$ & 0.108 & 30/30 & 30/30 & 30/30 & 100 \\
& 3 & $\varnothing~32$ & 0.099 & 30/30 & 30/30 & 0/30 & 66.7 \\
& 4 & $\varnothing~40$ & 0.127 & 30/30 & 30/30 & 30/30 & 100 \\
& 5 & $\varnothing~30$ & 0.0585 & 17/30 & 0/30 & 0/30 & 18.9 \\
& 6 & $\varnothing~32$ & 0.056 & 30/30 & 30/30 & 30/30 & 100 \\
& 7 & $\varnothing~25$ & 0.0625 & 30/30 & 30/30 & 30/30 & 100 \\
& 8 & $\varnothing~26$ & 0.0605 & 30/30 & 30/30 & 0/30 & 66.7 \\
& 9 & $71\times20$ & 0.0855 & 30/30 & 30/30 & 30/30 & 100 \\
\hline
\multirow{2}{*}{Hammer}
& 1 & $\varnothing~20$ & 0.230 & 30/30 & 30/30 & 30/30 & 100 \\
& 2 & $\varnothing~32$ & 0.742 & 30/30 & 30/30 & 30/30 & 100 \\
\hline
\multirow{2}{*}{Drill}
& 1 & $\varnothing~60$ & 1.040 & 30/30 & 30/30 & 0/30 & 66.7 \\
& 2 & $\varnothing~55$ & 0.284 & 30/30 & 30/30 & 30/30 & 100 \\
\hline
\end{tabular}%
}
\end{threeparttable}
\end{table}

\begin{figure}[!t]
\centering
 	\begin{subfigure}[b]{0.22\linewidth}
         \centering
        \includegraphics[width=1\linewidth, clip, trim=0pt 0pt 0pt 0pt]{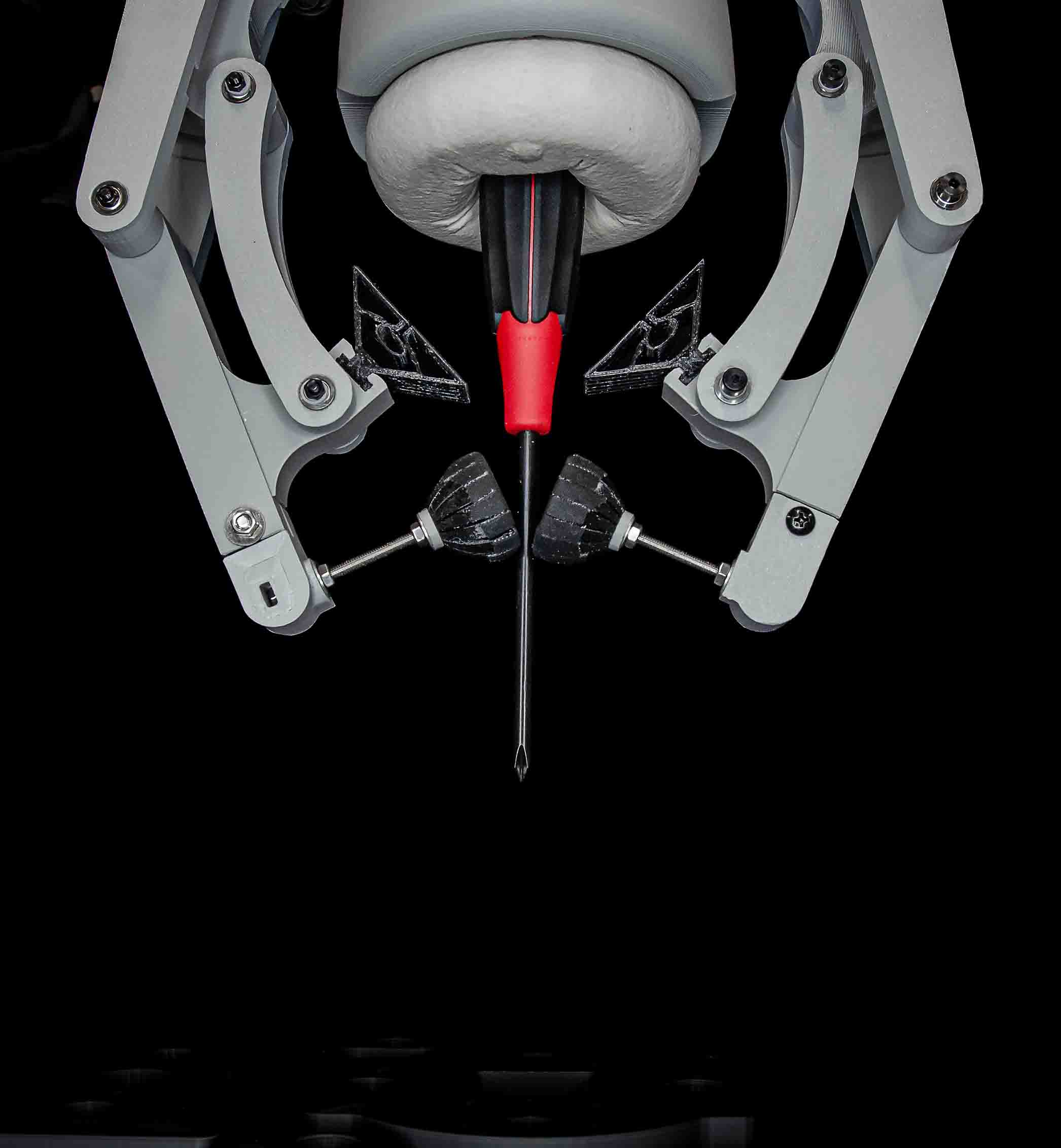}
        \vspace{-5.5mm}
        \caption{}
        \label{subfig11_1}
    \end{subfigure}
    ~
  	\begin{subfigure}[b]{0.22\linewidth}
         \centering
         \includegraphics[width=1\linewidth, clip, trim=0pt 0pt 0pt 0pt]{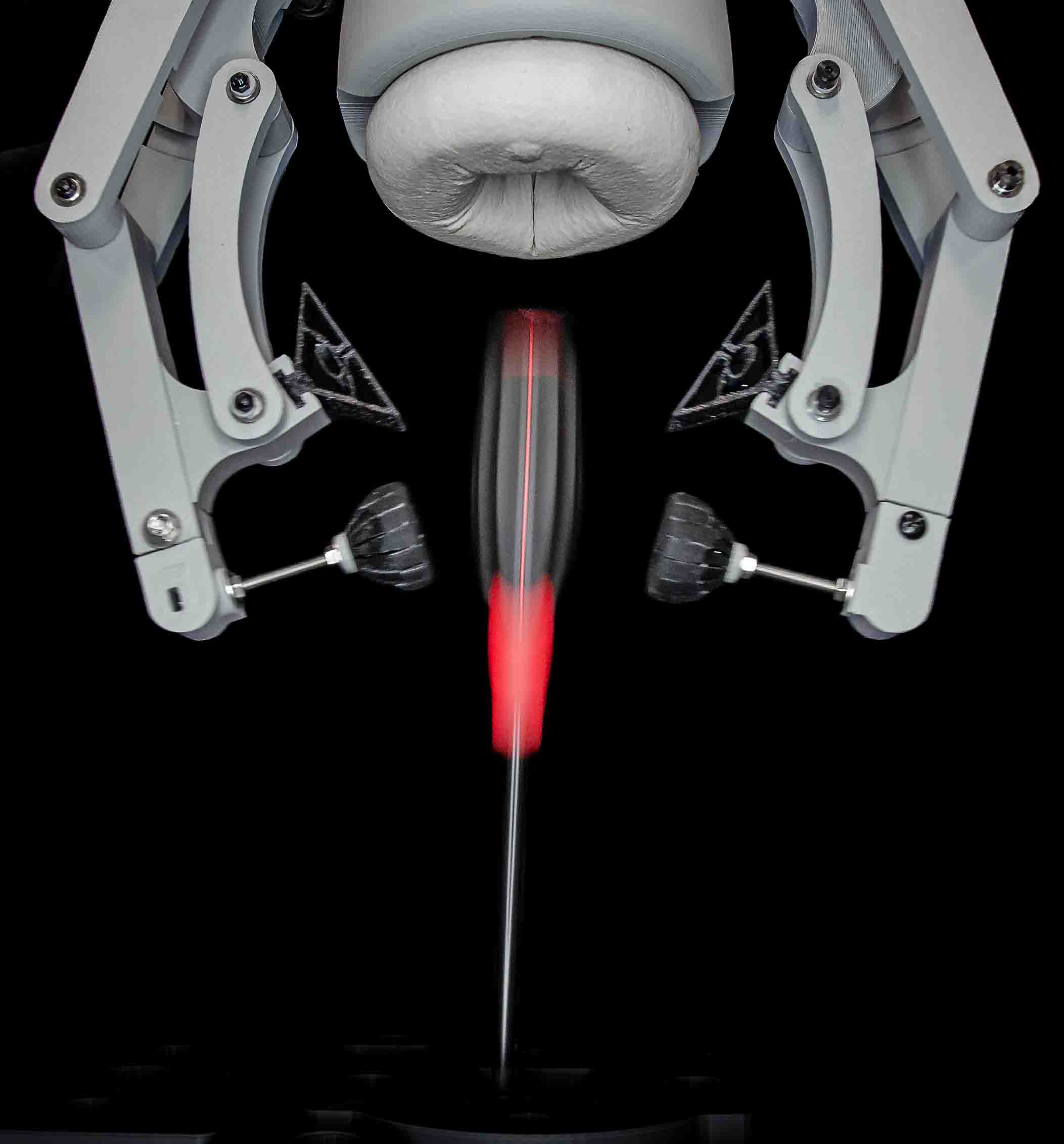}
         \vspace{-5.5mm}
        \caption{}
        \label{subfig11_2}
	\end{subfigure}
    ~
 	\begin{subfigure}[b]{0.22\linewidth}
         \centering
        \includegraphics[width=1\linewidth, clip, trim=0pt 0pt 0pt 0pt]{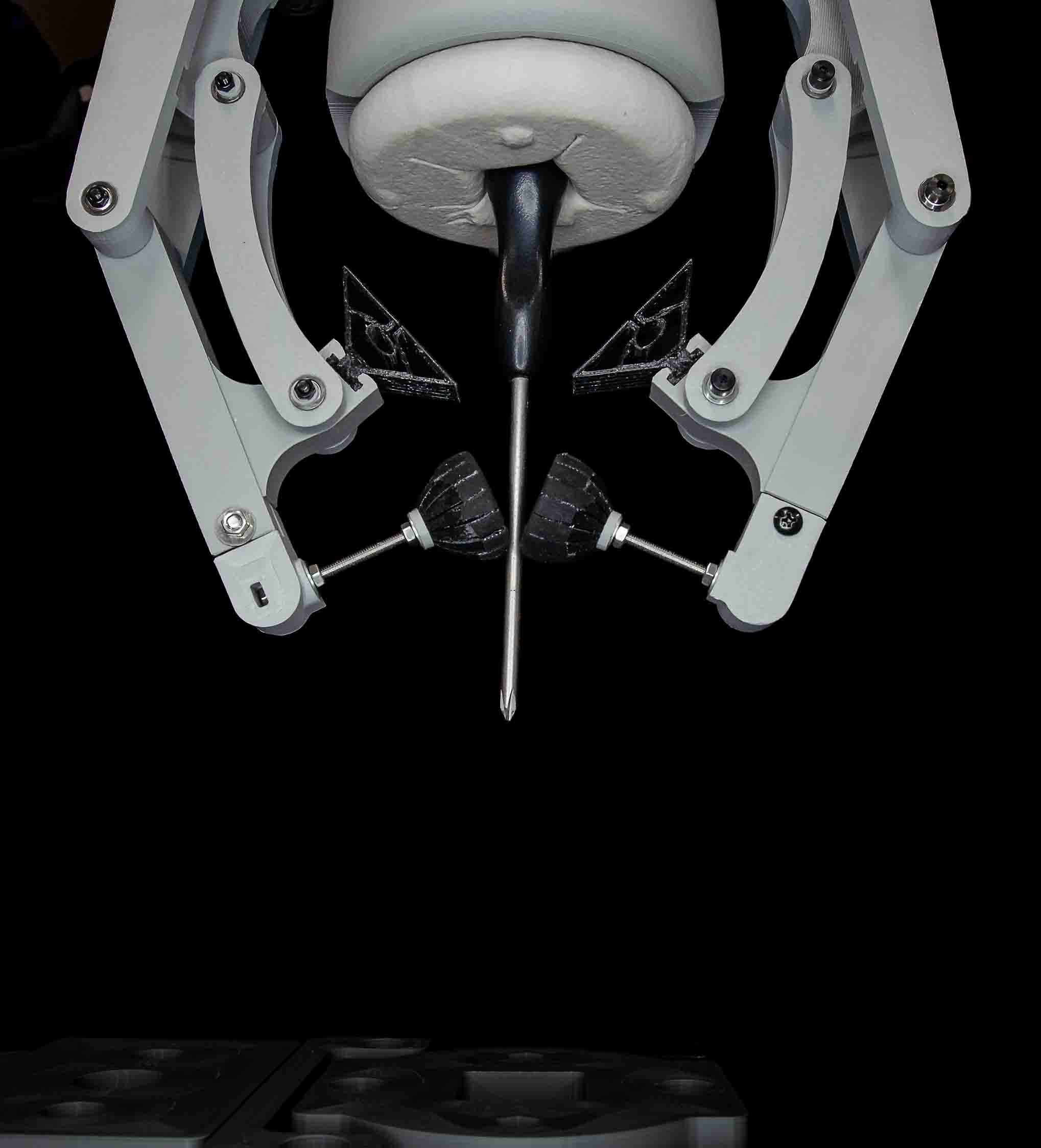}
        \vspace{-5.5mm}
        \caption{}
        \label{subfig11_3}
    \end{subfigure}
    ~
  	\begin{subfigure}[b]{0.22\linewidth}
         \centering
         \includegraphics[width=1\linewidth, clip, trim=0pt 0pt 0pt 0pt]{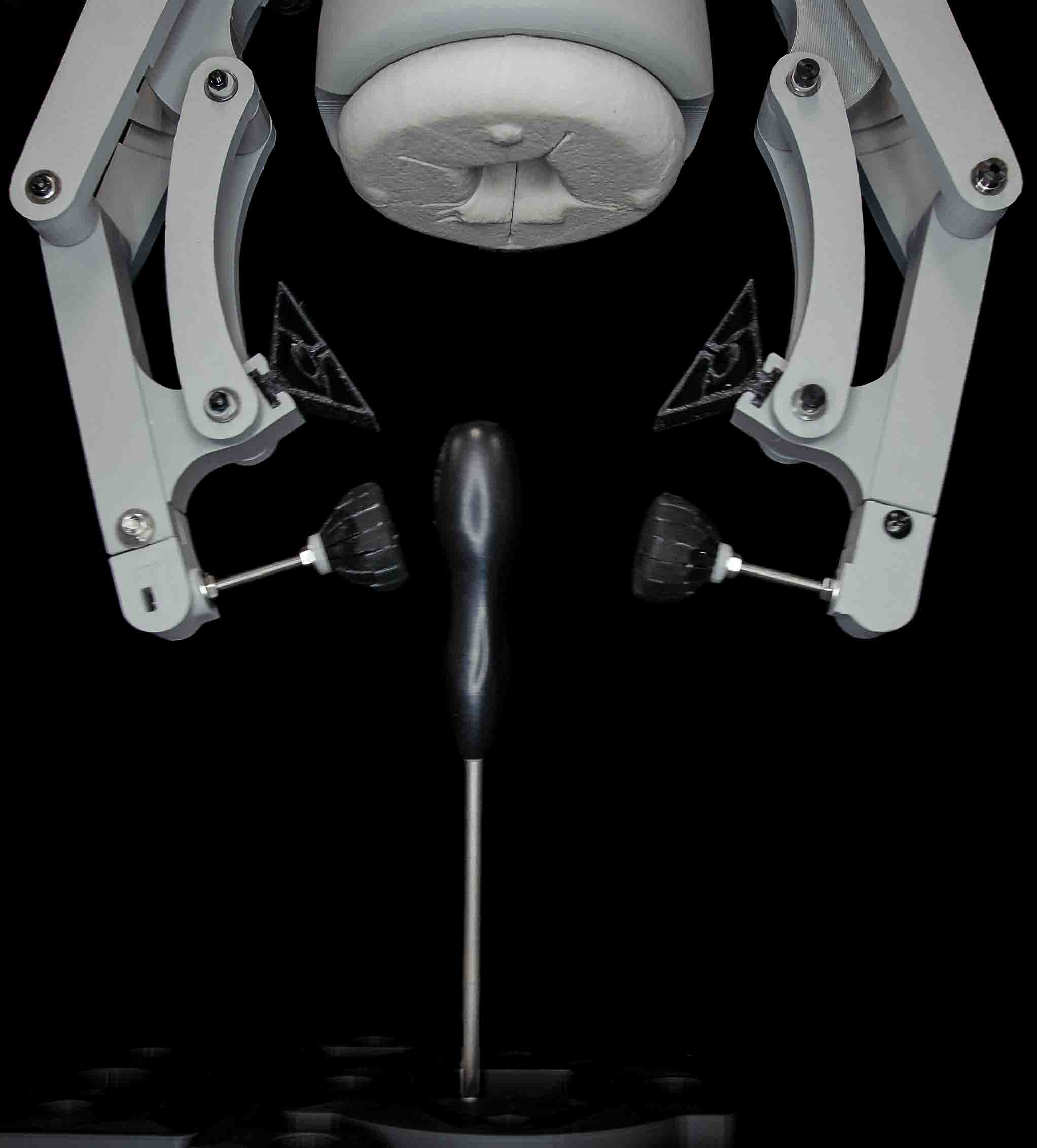}
         \vspace{-5.5mm}
        \caption{}
        \label{subfig11_4}
	\end{subfigure}\\
        \vspace{-1mm}
 \caption{Loss of contact with screwdrivers under releasing: (a) holding \textnumero 3; (b) finger opening and falling \textnumero 3; (c) holding \textnumero 8; (d) finger opening and falling \textnumero 8.}
 \label{fig11}
\end{figure}

To assess the stages of the grasping process and tool use by ARTiS, a success score was calculated (30 times for each tool, Table~\ref{tab2}): jamming with lifting, grasping with releasing, grasping with reorientation/using, and releasing. For this application and device, it has been determined that the failure criterion is the loss of fixation/contact between the tool and the palm. 

Analyzing the grasping "\textit{Jamming + Lifting}" scoring (Table~\ref{tab2}), results show a one hundred percent success rate for all tools except the screwdriver \textnumero5, which is 57\%, while the overall success rate is 97\%. This indicates the effectiveness of the jamming palm for grasping tools of various shapes. The geometry of \textnumero5~\cite{Project_Artis}, has the smoothest surface of all screwdrivers; there are no protrusions, and the smooth plastic used makes it easy to slip through. Because of this combination of characteristics, it is very difficult to grab such a handle surface with a jamming palm. The solution may be to increase the depth of the jamming palm so that the entire screwdriver handle can be completely immersed in the palm. However, such an increase will lead to doubling the outer shell of the palm and the amount of filler material (granules), which is very heavy. This would lead to the need to fabricate a special shape and thickness of the jamming palm surface and add complexity to the existing design. An identical situation occurs for "\textit{Grasping with Releasing}", where all tools achieve one hundred percent success, except for the screwdriver \textnumero5 (0\%). In this case, even when the jamming palm has successfully grasped the screwdriver and lifted it, the process of fixing the tool with the fingers occurs with 100\% success. However, when pressing fingertip 1 on the screwdriver, the screwdriver slips in the middle of the palm, and therefore, when the fingers are opened, the screwdriver simply falls out of the palm (Fig.~\ref{subfig11_4}). In this case, the solution may be identical to the previous one, increasing the contact area and, accordingly, the frictional properties between the tool handle and the palm.

\begin{figure}[!t]
    \centering
    \includegraphics[width=1\linewidth,clip ,trim=0pt 0pt 0pt 0pt]{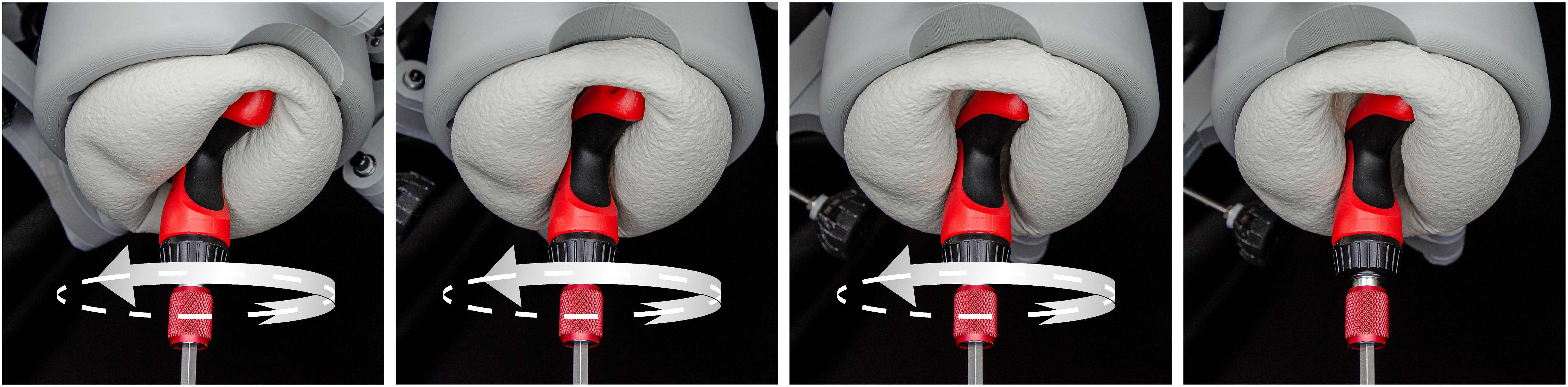}
    \vspace{-1.5mm}
    \caption{Tightening a screw to measure the torque that a gripping device can generate, in this case with a \textnumero 9 screwdriver.}
    \label{fig_8_1}
\end{figure}

The most difficult and most important operation for robotic disassembly is the "\textit{Grasping with Reorientation/Using and Releasing}" operation, which is 100\% successful for only 5 out of 9 screwdrivers, 2 out of 2 hammers, and 1 out of 2 drills. In the case of screwdrivers \textnumero3, 5, and 8, after successful jamming, lifting, and fixation (at this stage, screwdriver \textnumero5 slipped in the middle of the palm), tool tip reorientation was performed using fingertip 1, which, for \textnumero3 and 8, caused a slip in the middle of the palm. The slipping of screwdrivers in the mid-palm during reorientation is caused by insufficient frictional properties between the screwdriver handle and the palm surface. As a result of the slip in the palm at the release stage (after opening the fingers), the screwdrivers fall out of the palms (Fig.~\ref{fig11}). This situation is due to the smooth shape of screwdrivers \textnumero5 and 8, as they easily lose contact with the palm because they have no protrusions to grab onto. Another tool that has a 0\% success rate for use is the drill \textnumero1. When using the drill \textnumero1 (video~\cite{Project_Artis}), it can be clearly seen that ARTiS allows grasping and bringing the drill to the working area. However, during drilling, vibrations of the drill occur, which are transmitted to the fingertips and palm. For the fingertips, there are no changes and contact is not lost, but for the palm, there is the effect of compacting the granules inside the palm. A similar effect of granule occurs when tightening a screw with a screwdriver \textnumero9 (Fig.~\ref{fig_8_1}), which increases the gap between the tool and the palm, and leads to loss of contact. In the case of the drill \textnumero1, this effect is caused by vibration and spreads over the entire contact plane. This causes the loss of contact with the palm, at which point all the weight and effort fall on the fingers, which are unable to fully hold the working drill. Unlike drill 1, drill 2 has a 100\% success rate, which is due to significantly less vibration that occurs during the drilling. One of the options for ensuring successful drilling with drill \textnumero1 is to minimize vibration. Therefore, one of the methods for avoiding unwanted vibrations can be to control the feed force and drilling speed of the tool. The success rate of "\textit{Grasping with Reorientation/Using and Releasing}," taking into account all tools, is 68\%, and the total for all three stages is 85\%, which is a strong indicator of the versatility of grasping and using tools. 


\begin{figure}[!t]
\centering
 	\begin{subfigure}[b]{0.68\linewidth}
         \centering
        \includegraphics[width=1\linewidth, clip, trim=0pt 0pt 0pt 0pt]{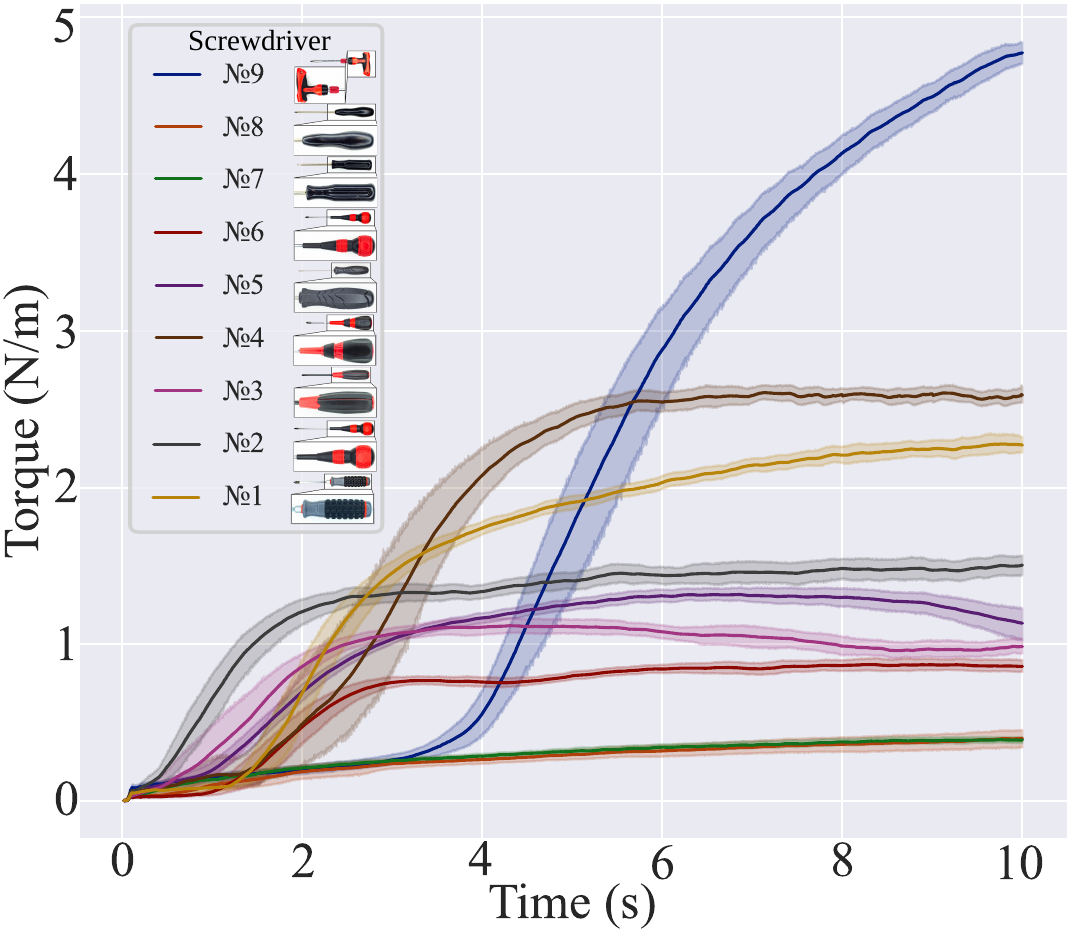}
        \vspace{-5.5mm}
        \caption{}
        \label{subfig10_1}
    \end{subfigure} \\
  	\begin{subfigure}[b]{0.68\linewidth}
         \centering
         \vspace{2mm}
         \includegraphics[width=1\linewidth, clip, trim=0pt 0pt 0pt 0pt]{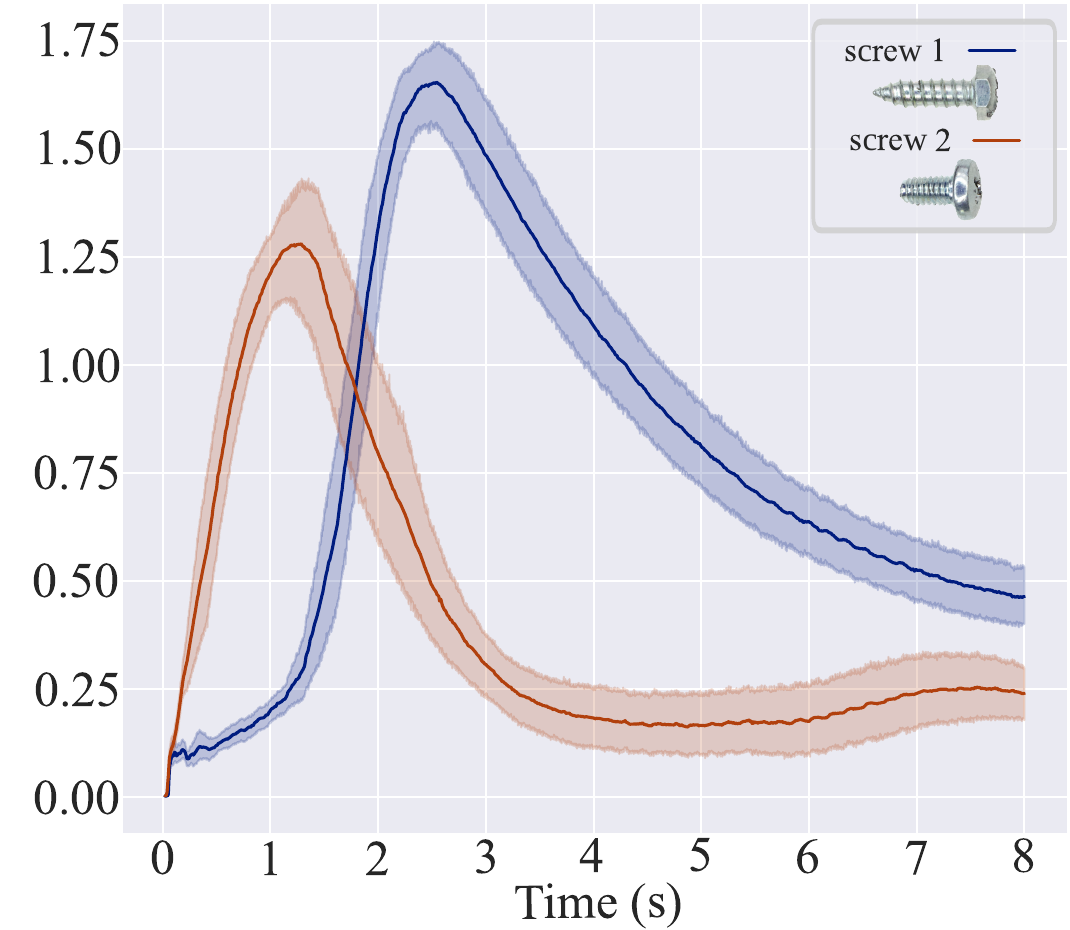}
         \vspace{-5.5mm}
        \caption{}
        \label{subfig10_2}
	\end{subfigure}\\
        \vspace{-1mm}
 \caption{The torque obtained: (a) with different screwdrivers and grippers when tightening a screw; (b) when unscrewing.}
 \label{fig10_1}
\end{figure}

Also, there are two standard methods of using screwdrivers in assembly and disassembly: tightening and unscrewing. The torque was found by taking data (30 times for each) from the 6 DoF force/torque sensor on the robot's wrist (Fig.~\ref{fig_8_1}), while tightening (3 deg/sec) with all the studied screwdrivers (Fig.~\ref{subfig10_1}) and while unscrewing (5 deg/sec) two types of screws used in the design of the studied car's air conditioner (Fig.~\ref{subfig10_2}). The screw tightening data obtained allows us to estimate the torque that different screwdriver handle shapes can provide when interacting with a jamming palm. The torque generated by the gripper with nine screwdrivers was statistically evaluated Table~\ref{tab:sd_torque_statistics}. The main performance indicators were maximum torque, torque at 10~s, mean torque, RMS torque, torque-time integral (area under the curve (AUC), which was used to quantify the accumulated torque output over the 10~s interval), and the time required to reach the maximum torque. A higher AUC indicates that the screwdriver maintained a higher torque level over time, rather than only producing a short-duration peak.

\begin{table}[t]
\centering
\caption{Torque performance statistics for different screwdrivers when tightening a screw. Values are reported as mean $\pm$ SD.}
\label{tab:sd_torque_statistics}
\begin{threeparttable}
\scriptsize
\setlength{\tabcolsep}{6.5pt}
\renewcommand{\arraystretch}{0.95}
\begin{tabular}{ccccc}
\hline
\textbf{Screwdriver} &
\makecell{\textbf{Peak / 10 s}\\\textbf{(N$\cdot$m)}} &
\makecell{\textbf{Mean / RMS}\\\textbf{(N$\cdot$m)}} &
\makecell{\textbf{AUC}\\\textbf{(N$\cdot$m$\cdot$s)}} &
\makecell{\textbf{$t_{\max}$}\\\textbf{(s)}} \\
\hline
1 &
\makecell{2.29 $\pm$ 0.10\\2.27 $\pm$ 0.09} &
\makecell{1.57 $\pm$ 0.07\\1.75 $\pm$ 0.06} &
15.66 $\pm$ 0.72 &
9.63 $\pm$ 0.39 \\

2 &
\makecell{1.53 $\pm$ 0.11 \\1.50 $\pm$ 0.11} &
\makecell{1.26 $\pm$ 0.08 \\1.31 $\pm$ 0.08} &
12.57 $\pm$ 0.76 & 9.39 $\pm$ 0.70 \\

3 &
\makecell{1.14 $\pm$ 0.08\\0.99 $\pm$ 0.08} &
\makecell{0.91 $\pm$ 0.06\\0.96 $\pm$ 0.06} &
9.05 $\pm$ 0.59 &
5.04 $\pm$ 0.94 \\

4 &
\makecell{2.64 $\pm$ 0.09\\2.59 $\pm$ 0.08} &
\makecell{1.80 $\pm$ 0.19\\2.06 $\pm$ 0.14} &
18.01 $\pm$ 1.89 &
7.52 $\pm$ 1.33 \\

5 &
\makecell{1.34 $\pm$ 0.07\\1.14 $\pm$ 0.17} &
\makecell{1.01 $\pm$ 0.05\\1.09 $\pm$ 0.06} &
10.14 $\pm$ 0.46 &
6.97 $\pm$ 0.86 \\

6 &
\makecell{0.90 $\pm$ 0.06\\0.86 $\pm$ 0.07} &
\makecell{0.67 $\pm$ 0.05\\0.73 $\pm$ 0.05} &
6.72 $\pm$ 0.49 &
8.28 $\pm$ 1.09 \\

7 &
\makecell{0.40 $\pm$ 0.03\\0.39 $\pm$ 0.03} &
\makecell{0.29 $\pm$ 0.02\\0.31 $\pm$ 0.03} &
2.89 $\pm$ 0.25 &
9.30 $\pm$ 0.73 \\

8 &
\makecell{0.41 $\pm$ 0.09\\0.40 $\pm$ 0.09} &
\makecell{0.27 $\pm$ 0.05\\0.29 $\pm$ 0.05} &
2.70 $\pm$ 0.52 &
9.11 $\pm$ 1.76 \\

9 &
\makecell{4.79 $\pm$ 0.13\\4.77 $\pm$ 0.12} &
\makecell{2.05 $\pm$ 0.21\\2.72 $\pm$ 0.20} &
20.51 $\pm$ 2.05 &
9.92 $\pm$ 0.12 \\
\hline
\end{tabular}
\end{threeparttable}
\end{table}

\begin{table}[t] 
\centering 
\caption{Peak torque statistics for two screws during unscrewing. Values are reported as mean $\pm$ SD.} \label{tab:two_screw_peak_torque} 
\scriptsize 
\resizebox{\columnwidth}{!}{%
\begin{tabular}{ccccc} 
\hline 
\textbf{Screw} & \makecell{\textbf{Peak torque}\\\textbf{ (N$\cdot$m)}} & \makecell{\textbf{SEM (N$\cdot$m)}} & \makecell{\textbf{95\% CI (N$\cdot$m)}} & \makecell{\textbf{Time to peak (s)}} \\ 
\hline 
1 & 1.33 $\pm$ 0.22 & 0.07 & [1.17, 1.49] & 1.29 $\pm$ 0.20 \\ 
2 & 1.71 $\pm$ 0.15 & 0.05 & [1.60, 1.81] & 2.36 $\pm$ 0.23 \\ 
\hline 
\end{tabular}} 
\end{table}

As can be seen from the results obtained in Table~\ref{tab:sd_torque_statistics}, among the tested screwdrivers, screwdriver \textnumero 9 demonstrated the highest overall torque performance. It achieved the largest maximum torque, 4.79 $\pm$ 0.13~Nm, and the highest torque at 10~s, 4.77 $\pm$ 0.12~Nm. Screwdriver \textnumero 9 also showed the highest mean torque, RMS torque, and AUC, indicating that it not only reached a high peak torque but also maintained a higher torque level throughout the 10~s interval.

The second-best configuration in terms of maximum torque was screwdriver \textnumero 4, with 2.64 $\pm$ 0.09~Nm. However, screwdriver \textnumero 9 exceeded \textnumero 4 by approximately 81.6\% in maximum torque and by approximately 84.6\% in torque at 10~s. Compared with screwdriver \textnumero 1, 9 produced approximately 109.0\% higher maximum torque and approximately 110.1\% higher torque at 10~s. These results indicate a substantial improvement in torque transmission for screwdriver \textnumero 9.

Although screwdriver \textnumero 3 reached its maximum torque the fastest, with a time-to-maximum torque of 5.04 $\pm$ 0.94~s, its maximum torque was relatively low at 1.14 $\pm$ 0.08~Nm. Therefore, fast torque saturation did not correspond to superior torque capacity. In contrast, screwdriver \textnumero 2 reached its maximum torque later, at 9.36 $\pm$ 0.7~s, but achieved a relatively high torque level of 1.53 $\pm$ 0.11~Nm.

Overall, the most advanced and hand-friendly screwdriver designs (\textnumero 1, 2, 4, and 9) showed the highest torques in this test. They provided the highest peak torque, the highest final torque after 10~s, and the greatest accumulated torque over time. Screwdriver \textnumero 4 was the second strongest configuration, while \textnumero 7 and 8 showed the lowest torque performance. The geometry and dimensions of the tool affect the torque the most, and this can be easily proven by analyzing screwdrivers with the same design, \textnumero 2 and 6, with different diameters, showing parallel lines (Fig.~\ref{subfig10_1}) with a constant deviation caused by the increase in the diameter of the contact surface. All hand-friendly screwdriver designs (\textnumero 1, 2, 4, and 9) allow unscrewing the screws used in a car air conditioner (Fig.~\ref{subfig10_1}).



\begin{figure}[!t]
    \centering
    \vspace{-3mm}
    \includegraphics[width=1\linewidth, clip, trim=0pt 0pt 0pt 0pt]{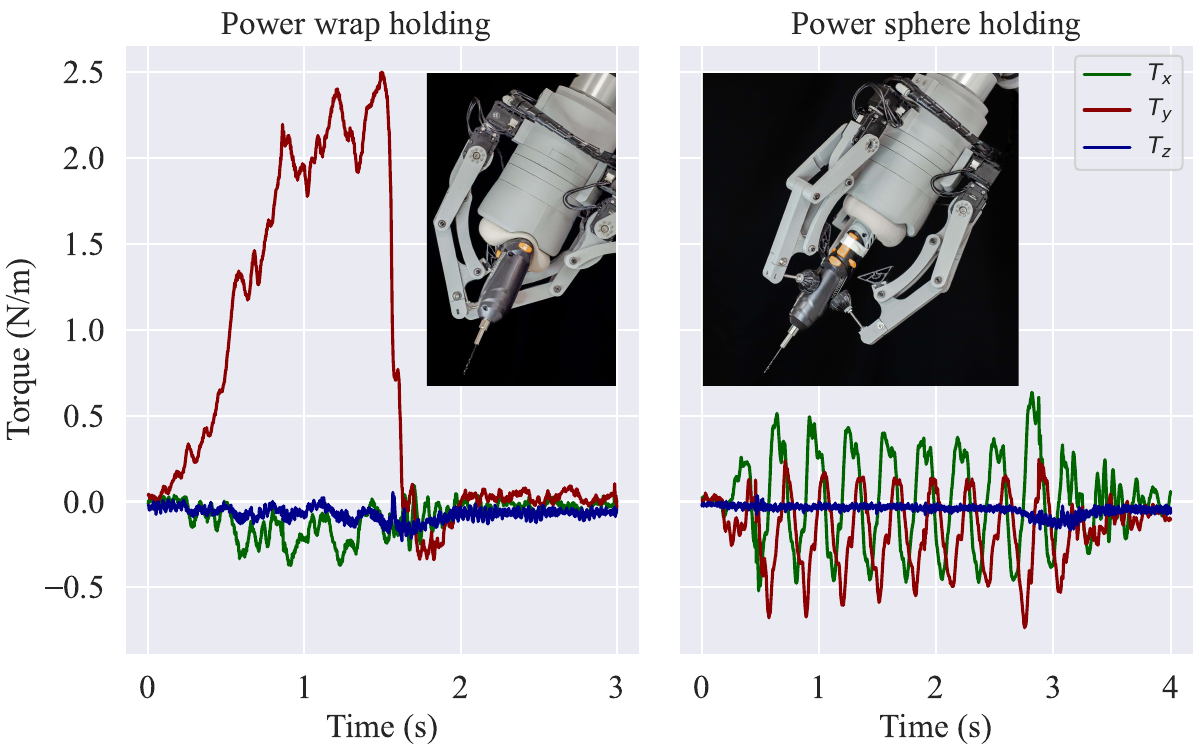}
    \vspace{-5.5mm}
    \caption{Torque when drilling plastic elements of a car air conditioner with different holding configurations.}
    \label{fig_12}
\end{figure}

The peak torque required during the unscrewing process was evaluated for two screws (Fig.~\ref{subfig10_2}, 30 trials for each screw). The main evaluated metrics were the peak torque, the standard error of the mean (SEM), and the time required to reach the peak torque. Table~\ref{tab:two_screw_peak_torque} summarizes the statistical results. The screw 1 condition required a higher peak torque than screw 2. The mean peak torque for screw 2 was 1.71 $\pm$ 0.15~Nm, with an SEM of 0.05~Nm, whereas screw 1 showed a mean peak torque of 1.33 $\pm$ 0.22~Nm, with an SEM of 0.07~Nm. Thus, the peak torque for screw 2 was higher by 0.38~Nm, corresponding to an increase of approximately 28.2\% compared with screw 1. In addition to the higher torque magnitude, screw 2 also reached its peak torque later than screw 1. The time to peak torque was 2.36 $\pm$ 0.23~s for screw 2 and 1.29 $\pm$ 0.20~s for screw 1. Overall, these results show that screw 2 required both a higher peak torque and a longer time to reach this peak compared with screw 1. Therefore, screw 2 can be considered the more demanding screw condition in terms of torque requirement during unscrewing. 


\begin{figure*}[!t]
    \centering
    \includegraphics[width=1\linewidth,clip ,trim=0pt 0pt 0pt 0pt]{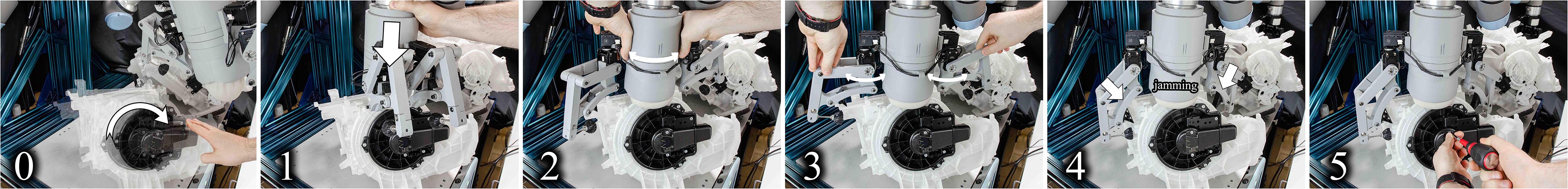}
    \vspace{-2mm}
    \caption{Collaborative usage of a gripping device ARTiS, in this case for fixing (to ensure stability of position and orientation) a part of the mechanism during disassembly: (0) - demonstration of the problem of part movement during interaction with a human; (1) - positioning the ARTiS palm on the surface of the part, pressing it against the table; (2) - correction of the position of the fingers relative to the gripper axis; (3) - correcting the orientation of the fingers relative to the finger axis; (4) - using palm jamming and finger clamping to fix the part; (5) - human work on unscrewing screws with a fixed part.}
    \label{fig_14}
\end{figure*}

During disassembly, a problem often arises from the inability to unscrew or disconnect certain elements or fasteners; therefore, it is common to drill out or destroy the connections. Power and precise holding methods were used to obtain the torque that occurs when a plastic air conditioner fastener is drilled (Fig.~\ref{fig_12}). The obtained data show significant fluctuations during precise holding, which are due to both the small area of contact with the palm and the significant distance to the tool tip (which creates a larger force lever). Instead, power holding performs much better and shows smaller fluctuations due to greater contact with the palm and a smaller force lever. The jump in torque relative to the y-axis during a power wrap holding operation is caused by the pressure on the drill, which is at an angle to the axis of the gripper (sensor). The implementation of various robotic grasping and holding techniques for tools, along with their successful application, demonstrates the substantial versatility of ARTiS.

To further evaluate the grasping generalization of ARTiS beyond the objects used in the disassembly experiments, an additional grasping experiment was conducted using objects from the YCB benchmark set~\cite{calli2015ycb}. The purpose of this experiment was specifically to assess whether the adaptive grasping architecture could acquire and securely lift objects with geometries and physical characteristics different from those of the original disassembly tools; functional tool use or task-level manipulation of these objects was not evaluated.
Three object subsets were selected: food items, kitchen items, and tool items. Sharp or puncturing objects were excluded to avoid damaging the jamming palm. For each object, ARTiS performed 30 grasping attempts, and a trial was counted as successful when the object was securely grasped and lifted without dropping or unstable slipping.  

\begin{table}[t] 
\centering 
\caption{Grasping Success Rate of ARTiS on Selected YCB Object Subsets} \label{tab:ycb_generalization} 
\scriptsize 
\resizebox{\columnwidth}{!}{
\begin{tabular}{l c c c c c c} 
\toprule 
\textbf{YCB subset} & \textbf{Objects} & \textbf{Attempts} & \textbf{Success} & \makecell{\textbf{Success}\\\textbf{rate}} & \textbf{95\% CI} & \makecell{\textbf{100\%} \\\textbf{success} \\\textbf{objects}} \\ 
\midrule Food items & 18 & 540 & 526 & 97.41\% & 95.7-98.4\% & 15/18 \\ 
Kitchen items & 12 & 360 & 321 & 89.17\% & 85.5-91.9\% & 7/12 \\ Tool items & 13 & 390 & 382 & 97.95\% & 96.0-98.9\% & 11/13 \\ 
\midrule \textbf{Total} & \textbf{43} & \textbf{1290} & \textbf{1229} & \textbf{95.27\%} & \textbf{93.9-96.3\%} & \textbf{33/43} \\ 
\bottomrule 
\end{tabular} } 
\vspace{1mm} 
\begin{flushleft} 
\scriptsize
\vspace{-2mm}
The confidence intervals were computed using the Wilson score interval over the number of successful grasping trials. 
\end{flushleft} 
\end{table}

Table~\ref{tab:ycb_generalization} summarizes the grasping success rate over 43 evaluated objects and 1290 total grasping attempts. ARTiS achieved an overall success rate of 95.27\% across all selected YCB objects. The highest performance was obtained for tool items and food items, with success rates of 97.95\% and 97.41\%, respectively. This result indicates that the combination of the jamming palm and adaptive fingertips provides stable grasping over objects with diverse dimensions, surface properties, and shapes, including cylindrical cans, boxes, fruit-like objects, hand tools, and small tool-related components. The kitchen subset resulted in a lower success rate of 89.17\%, mainly due to several objects with challenging geometries and a displaced center of mass. In particular, thin or flat objects, such as the metal plate and cooking skillet, produced lower grasping reliability than bulky or rounded objects. Nevertheless, ARTiS successfully grasped most objects in this subset, including the plastic wine glass, metal bowl, metal mug, glass lid, spoon, fork, and spatula. Across all subsets, 33 out of 43 objects were grasped with 100\% success, and 36 objects achieved a success rate of at least 90\%. These results complement the tool-specific grasping and holding experiments by showing that the adaptive palm-fingertip architecture is not limited to the geometries represented by the original disassembly tool set. Rather, ARTiS can establish stable grasps across a broader range of object shapes and physical characteristics. It should be emphasized, however, that these experiments evaluate grasp acquisition and lifting only; they do not demonstrate generalized functional tool use or task-level manipulation with previously unseen tools. Detailed per-object results, including the number of failures and Wilson 95\% confidence intervals, are provided in Appendix~\ref{A3}.

\subsection{Evaluating Task Versatility}
\noindent A qualitative assessment of ARTiS's ability to perform more than just standard tasks while holding and using tools during disassembly was conducted. Since human-robot collaboration is very common during disassembly, it is important to ensure such cooperation. Most components and parts during disassembly have irregular surface that provide poor stability during handling (Fig.~\ref{fig_14} - slide 0, by pressing the air conditioner with a hand, it starts to move downwards, which does not allow for applying force to unscrew the screws from the part). Therefore, a person often has to hold the part with one hand and work with the tool with the other, which is very inconvenient in most cases (method 1 - $M_1$). To solve this problem, ARTiS was reprogrammed for collaborative tasks, allowing a person to position the gripper on the part and activate individual parts (palm and fingers) for holding (Fig.~\ref{fig_14}). First, the "teaching mode robot" button of the collaborative arm is pressed, which allows the gripper to move to a position where the palm presses the part to the table (Fig.~\ref{fig_14} - slide 1). After that, slide 2 shows how the position of the fingers changes around the gripper axis. Later, on slide 3, the adjustment of the orientation of the fingers is shown. Subsequently, after obtaining the desired configuration of the gripper around the part (if necessary, repeat steps 1-3, Fig.~\ref{fig_14}), data is read from the encoders of the motors about the position of all links. Knowing the configurations of all links that were set by the human ($J_0$, $J_1$, $J_2$, $J_3$), they are fixed in this position. At the same time, due to the jamming of the palm and the actuation of the fingers ($J_4$, $J_5$, $J_6$  until the maximum torque is reached), the part is fixed on the table (Fig.~\ref{fig_14} - slide 4). A study was conducted with 12 subjects, and the average time of fixation of the part using a gripper (after training) was 8 seconds. Having a fully fixed part with the help of ARTiS, a human can easily unscrew all the screws using two hands to hold both the tool and the screw (Fig.~\ref{fig_14} - slide 5 - method 2 ($M_2$)). 

A within-subject human study was conducted to evaluate the effect of gripper air conditioner fixation on the time required for the unscrewing task (using a \textnumero 2 screwdriver). Twelve participants performed the task using two different methods, $M_1$ and $M_2$. Since the robot-gripper fixation introduces an additional preparation step, a fixation time of 8~s was added to the $M_2$ unscrewing time: $T_{\mathrm{M_2, total}} = T_{\mathrm{M_2}} + 8$. Each participant completed six trials for each method, and the participant-level mean was used for statistical comparison in order to avoid treating repeated trials from the same participant as independent samples.

The participant-level results are summarized in Table~\ref{tab:bolt_unscrewing_results}. For unscrewing three screws, $M_1$ required 39.12 $\pm$ 9.28~s, whereas $M_2$ required 25.23 $\pm$ 6.50~s for the unscrewing operation alone. After including the fixed 8~s robot-gripper setup time, the total $M_2$ task time was 33.23 $\pm$ 6.50~s. Thus, even when the setup time was included, $M_2$ reduced the total task completion time by an average of 5.89~s compared with $M_1$, corresponding to a mean reduction of approximately 13.77\%.

\begin{table}[t]
\centering
\caption{Task Completion Time for Unscrewing Three Screws}
\label{tab:bolt_unscrewing_results}
\begin{threeparttable}
\scriptsize
\resizebox{\columnwidth}{!}{%
\begin{tabular}{c c c c c c}
\toprule
\textbf{Part.} &
\textbf{$M_1$ (s)} &
\textbf{$M_2$ (s)} &
\textbf{$M_{2,\mathrm{total}}$ (s)} &
\textbf{Saving (s)} &
\textbf{Saving (\%)} \\
\midrule
$P_1$  & 51.46 & 28.52 & 36.52 & 14.94 & 29.04 \\
$P_2$  & 60.35 & 42.01 & 50.01 & 10.34 & 17.14 \\
$P_3$  & 34.46 & 20.12 & 28.12 &  6.34 & 18.40 \\
$P_4$  & 29.54 & 21.73 & 29.73 & -0.19 & -0.65 \\
$P_5$  & 34.92 & 23.51 & 31.51 &  3.41 &  9.77 \\
$P_6$  & 43.77 & 21.49 & 29.49 & 14.28 & 32.63 \\
$P_7$  & 30.72 & 19.49 & 27.49 &  3.23 & 10.50 \\
$P_8$  & 35.81 & 21.65 & 29.65 &  6.16 & 17.21 \\
$P_9$  & 41.05 & 25.67 & 33.67 &  7.38 & 17.98 \\
$P_{10}$ & 34.36 & 23.43 & 31.43 &  2.93 &  8.52 \\
$P_{11}$ & 30.61 & 22.24 & 30.24 &  0.38 &  1.23 \\
$P_{12}$ & 42.39 & 32.89 & 40.89 &  1.50 &  3.53 \\
\midrule
\textbf{Mean}   & \textbf{39.12} & \textbf{25.23} & \textbf{33.23} & \textbf{5.89} & \textbf{13.77} \\
\textbf{SD}     & 9.28 & 6.50 & 6.50 & 5.08 & 10.34 \\
\textbf{Median} & 35.37 & 22.84 & 30.84 & 4.79 & 13.82 \\
\bottomrule
\end{tabular}}
\begin{tablenotes}
\scriptsize
\item $M_1$: one hand holds the workpiece and the other hand operates the screwdriver. 
$M_2$: unscrewing time using both hands after the workpiece is fixed by the robot gripper.
\end{tablenotes}
\end{threeparttable}
\end{table}

Because the same participants performed both methods, the data were analyzed using paired statistical tests. A paired $t$-test showed a statistically significant reduction in task completion time for $M_2$ compared with $M_1$, $t(11)=4.02$, $p=0.0020$, with a large paired-effect size of Cohen's $d_z=1.16$. In addition, a Wilcoxon signed-rank test was performed as a non-parametric confirmation, which also indicated a significant difference between the two methods ($W=1.00$, $p=0.00098$). These results indicate that the reduction in task time was consistent across participants and was not only due to random trial-to-trial variation.

The results also show that the benefit of $M_2$ depends on the number of bolts to be unscrewed. The robot-gripper setup time is a fixed cost, whereas the time saved during unscrewing accumulates with each additional bolt. Based on the observed mean task times, the average per-bolt time was 13.04~s/bolt for $M_1$ and 8.41~s/bolt for $M_2$ during the unscrewing operation. Therefore, the average per-bolt saving, before considering the setup time, was approximately 4.63~s/bolt. The expected time saving for $n$ bolts can be expressed as $\Delta T(n) = 4.63n - 8.00$, where $\Delta T(n)>0$ indicates that $M_2$ is faster than $M_1$. From this model, the break-even point is obtained as $n = 8.00/4.63 \approx 1.73$. This means that $M_2$ is not expected to be advantageous for a single screw because the fixed robot-gripper setup time dominates the task duration. However, with two or more bolts, the accumulated per-screw saving compensates for the setup cost. In the tested case of three bolts, $M_2$ already provided a measurable and statistically significant reduction in total task time.

\begin{figure}[!t]
\centering
 	\begin{subfigure}[b]{1\linewidth}
         \centering
    \includegraphics[width=1\linewidth, clip, trim=0pt 0pt 0pt 0pt]{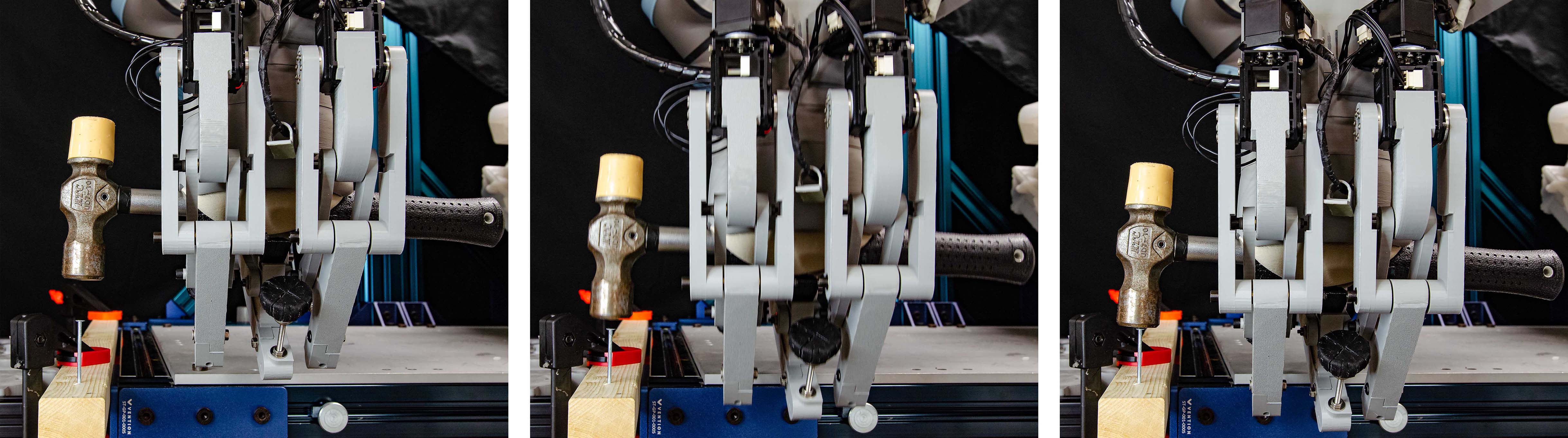}
        \vspace{-5.5mm}
        \caption{}
        \label{fig15_1}
    \end{subfigure} 
  	\begin{subfigure}[b]{1\linewidth}
         \centering
         \vspace{1mm}
    \includegraphics[width=1\linewidth, clip, trim=0pt 0pt 0pt 0pt]{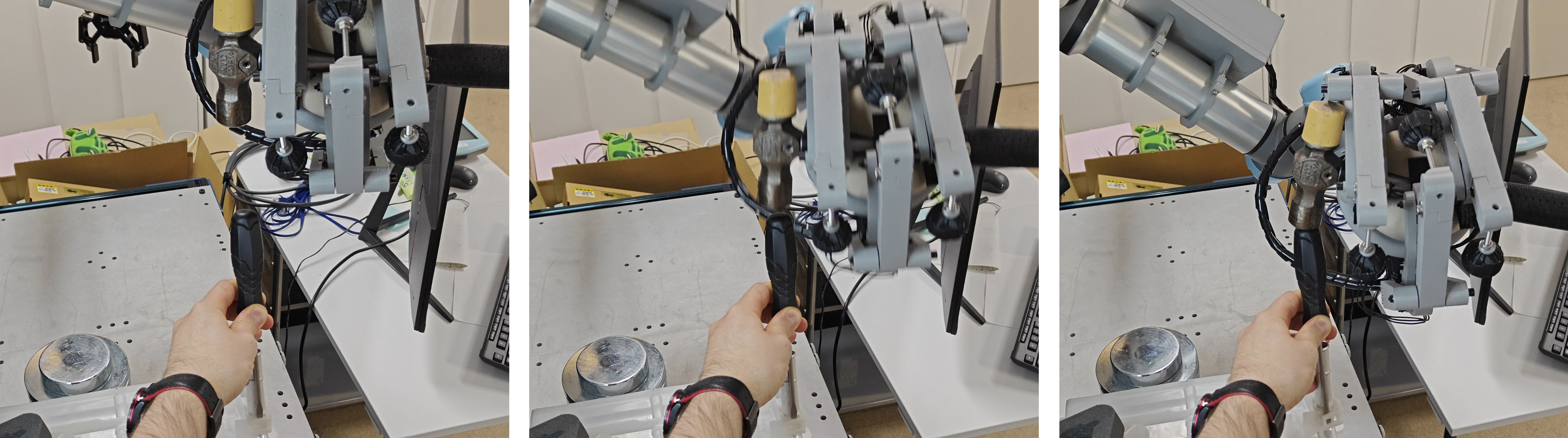}
        \vspace{-5.5mm}
        \caption{}
        \label{fig15_3}
	\end{subfigure}
    \begin{subfigure}[b]{1\linewidth}
         \centering
         \vspace{1mm}
    \includegraphics[width=1\linewidth, clip, trim=0pt 0pt 0pt 0pt]{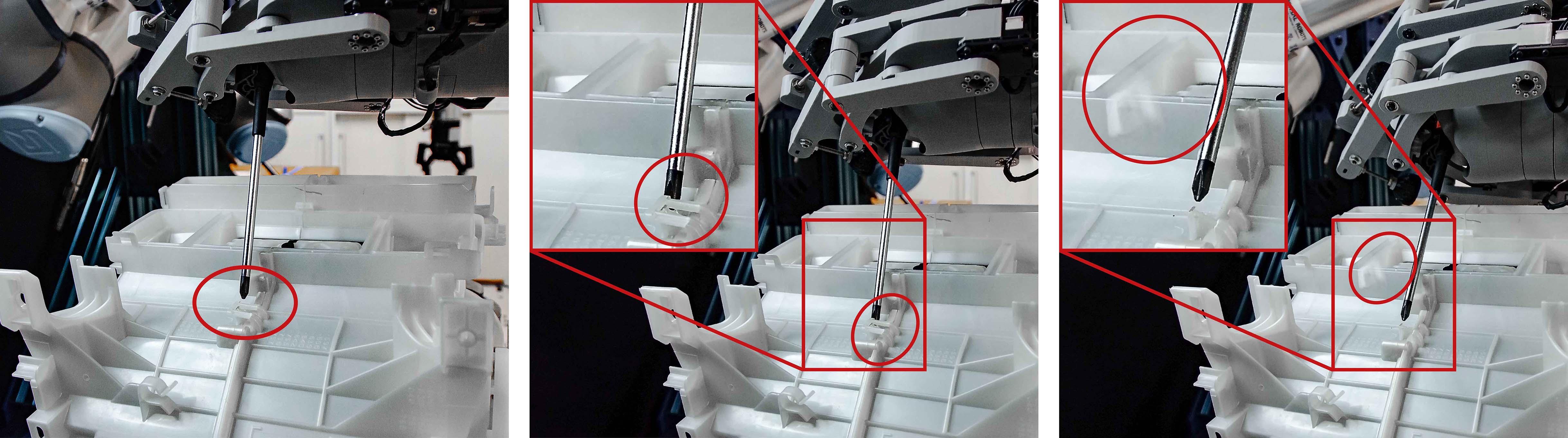}
        \vspace{-5.5mm}
        \caption{}
        \label{fig15_2}
	\end{subfigure}\\
        \vspace{-1.5mm}
\caption{Using ARTiS for power wrap holding/using tools: (a) hammer when driving a nail into wood;  (b) hammer hits the screwdriver to wedge it into the gap; (c) a screwdriver to break composite fasteners during disassembly.}
\label{fig_15}
\end{figure}

Overall, the results suggest that ARTiS part fixation is beneficial when the task involves repeated unscrewing operations. Although $M_2$ introduces an initial time cost, this cost is compensated as the number of screws increases. Therefore, the proposed ARTiS fixation strategy is particularly suitable for tasks in which multiple fasteners must be removed from the same workpiece.

In addition to the usual use of tools (screwdriver, hammer, and their combinations) are often used in disassembly to break structural components. The power wrap holding hammer can be used both to impact/break components (Fig.~\ref{fig15_1}) and to hit other tools (screwdrivers) to break or wedge into gaps (Fig.~\ref{fig15_3}). Another example is the use of screwdrivers with the ARTiS power wrap holding to break fasteners (Fig.~\ref{fig15_2}). In the presented variant, the screwdriver is inserted under the composite fastener, after which, leaving the position of the screwdriver tip unchanged, the orientation of the screwdriver is changed until it contacts the protrusion on the part. As a result, it gets a lever from the screwdriver, and when it is pressed sharply in the same direction, the fastener breaks and detaches from the part. This is analogous to how people use a screwdriver as a lever to increase the force transmitted to an element, allowing them to easily break off the necessary parts. These examples are only a few of the possibilities of using tools with ARTiS, indicating significant task versatility for various robotics disassembly applications.

\section{Conclusion and Future Work}
\noindent In this paper, a new robotic gripper design was proposed for common tools used in disassembly applications. The Adaptive Robotic Tool gripper in disassembly Systems (ARTiS) demonstrates that robotic tool handling should be considered not only a grasping problem but also a tool-fixation, tool-use, and human-teaching problem. By combining an adaptive jamming palm, pose-specific fingertips, and active finger reconfiguration, ARTiS provides a mechanically simple but functionally versatile approach for using human-designed tools in robotic disassembly. The additional YCB experiments demonstrate grasping generalization beyond the original set of disassembly tools, with ARTiS achieving stable grasp acquisition and lifting across objects with diverse geometries and physical characteristics. Generalized YCB functional tool use was not evaluated and remains an important direction for future work. The human study further demonstrates the potential of the teaching mode for human-robot collaboration and data collection.

Success rates were obtained for all stages: grasping 97\%, and grasping and tool reorientation by ARTiS 85\%. However, some tools were difficult to hold during operation due to their shape/material and the oscillation of the tool, which led to a loss of contact, so the success rate for using different tools was 68\%. Therefore, the main limitation of the gripper is the influence of tool oscillations on holding the tool in the palm and controlling its stiffness. There are several methods to deal with this: using stiffer palm materials, increasing the vacuum in the palm, or integrating active side stops into the palm. A secondary limitation of the gripper is its dimensions, which are due to the fabrication method of 3D printing. Using a different manufacturing method and structural optimization can solve this problem. Also, future studies will focus on implementing learning using human visual learning, ARTiS, with various tools.



%


\bibliographystyle{IEEEtran}
\IEEEtriggercmd{\enlargethispage{2in}}
\bibliography{Dex}

\vfill

\begin{IEEEbiography}[{\includegraphics[width=1in,height=1.25in,clip,keepaspectratio, trim=5pt 6pt 5pt 2pt]{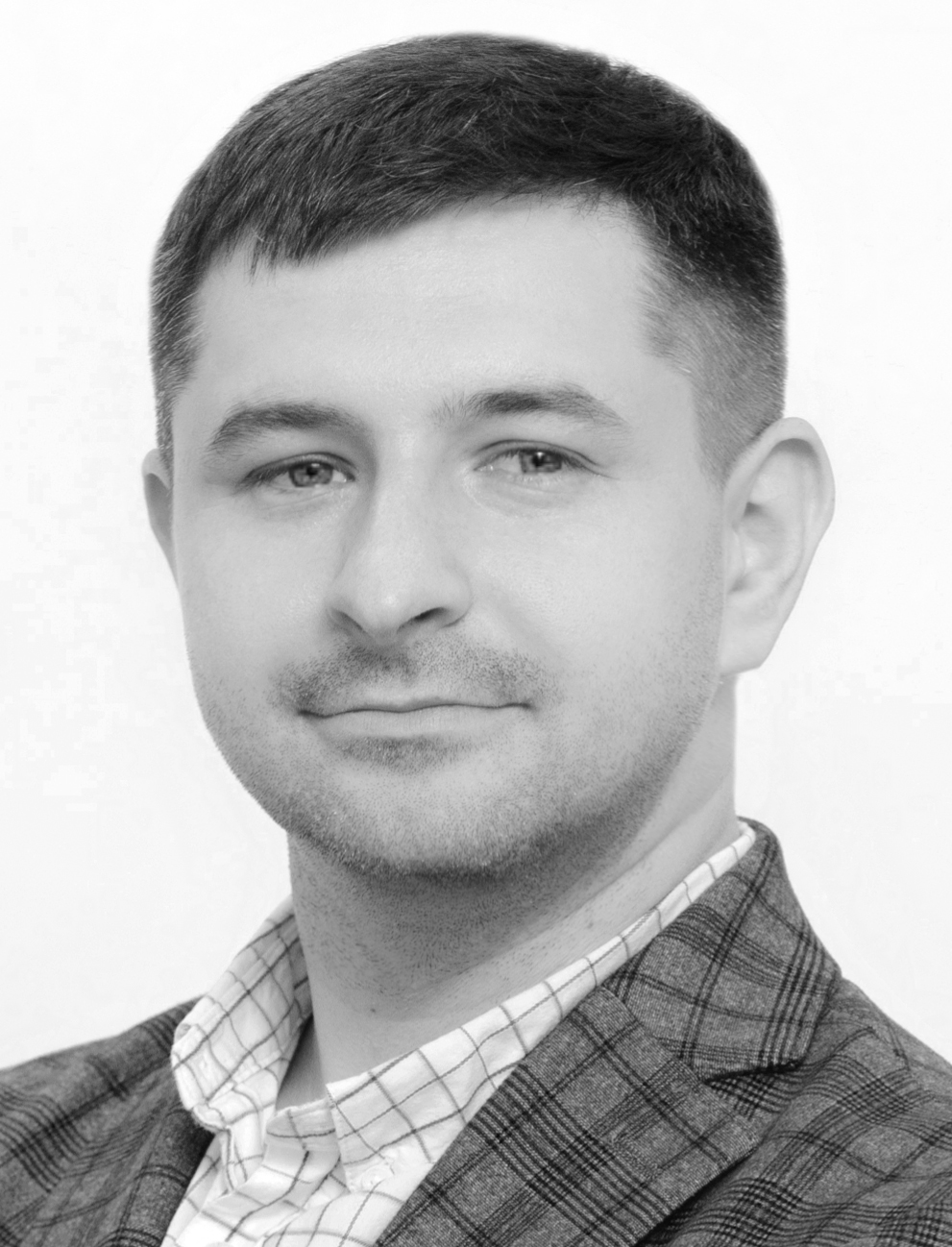}}]{Roman Mykhailyshyn}
(Senior Member, IEEE) received the B.Sc., M.Sc., and Ph.D. degrees in Engineering Sciences from Ternopil Ivan Puluj National Technical University, Ternopil, Ukraine, in 2013, 2014, and 2018, respectively.

He is currently a Senior Research Scientist at the National Institute of Advanced Industrial Science and Technology (AIST), Japan, and an Adjunct Professor since 2023 at the American University Kyiv (Powered by Arizona State University),  Ukraine.  He was an Associate Professor with the Department of Automation of Technological Processes and Manufacturing, Ternopil Ivan Puluj National Technical University, Ukraine, from 2019 to 2023. From 2021 to 2022, he was a Fulbright Visiting Scholar with the Department of Robotics Engineering, Worcester Polytechnic Institute (WPI), USA.  From 2022 to 2024, he was a Research Fellow (Provost's Early Career Fellow) in the Texas Robotics and Walker Department of Mechanical Engineering at The University of Texas at Austin, USA. From 2024 to 2025, he was a Research Associate Professor at The University of Osaka, Japan.  His research interests include robotics, grasping, manipulation, haptics, gripper design, and rapid prototyping. 

Dr. Mykhailyshyn was the recipient of the 2020 - 2023 Fellowships of the Cabinet of Ministers of Ukraine for Young Scientists and the 2022 Machines Young Investigator Award.
\end{IEEEbiography}


\begin{IEEEbiography}[{\includegraphics[width=1in, height=1.25in, clip, keepaspectratio, decodearray={0 1 0 1 0 1}]{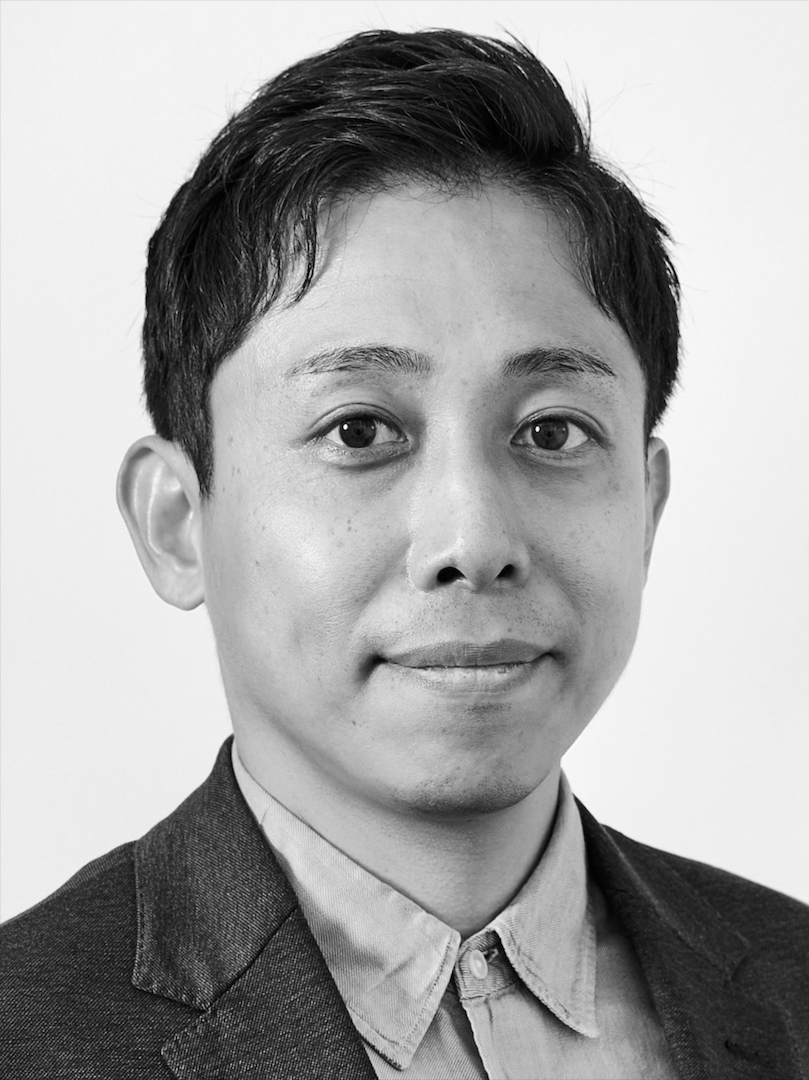}}]{Yukiyasu Domae}
(Senior Member, IEEE) received the B.E., M.E., and Ph.D. degrees from Hokkaido University, Japan, in 2004, 2006, and 2012, respectively. From 2008 to 2018, he was with Mitsubishi Electric Corporation, where he served as a Researcher and then as Principal Researcher in the Image Recognition Systems Group, Advanced Technology Research and Development Center. In 2018, he joined the National Institute of Advanced Industrial Science and Technology (AIST) as Group Leader of the Manipulation Research Group. Since 2019, he has led the Automation Research Team within the Embodied AI Research Team at the Artificial Intelligence Research Center, focusing on machine vision, robotic manipulation, and data-driven learning for robotics.
\end{IEEEbiography}


\begin{IEEEbiography}[{\includegraphics[width=1in,height=1.25in,clip,keepaspectratio]{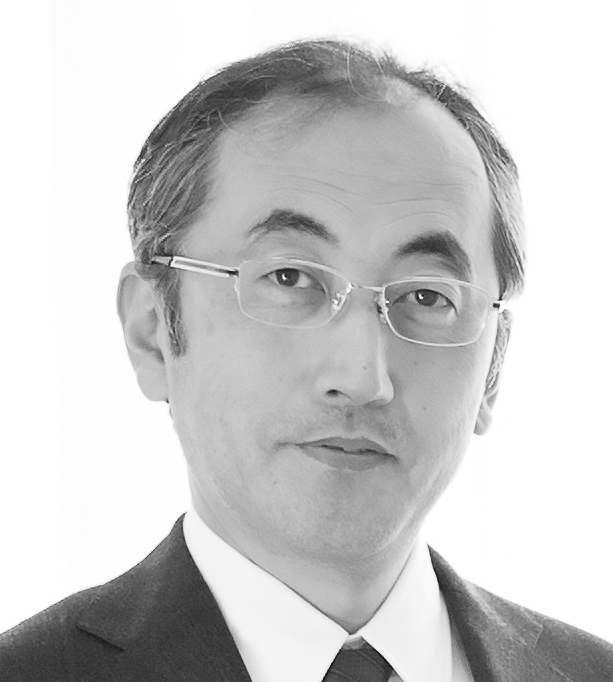}}]{Kensuke Harada}
(Fellow, IEEE) received the B.Sc., M.Sc., and Ph.D. degrees in mechanical engineering from Kyoto University, Kyoto, Japan, in 1992, 1994, and 1997, respectively. From 1997 to 2002, he was a Research Associate with Hiroshima University, Hiroshima, Japan. Since 2002, he has been with the National Institute of Advanced Industrial Science and Technology (AIST). From 2005 to 2006, he was a Visiting Scholar with the Department of Computer Science, Stanford University, Stanford, CA, USA, and the Leader of the Manipulation Research Group, AIST, from 2013 to 2015. He is currently a Professor with the Graduate School of Engineering Science, The University of Osaka, Toyonaka, Japan. His research interests include mechanics and control of robot manipulators and robot hands, biped locomotion, and motion planning of robotic systems.
\end{IEEEbiography}

\appendices

\section{Control Algorithm and Operation of ARTiS}\label{A2}

\subsection{Control Algorithm and Autonomous Operation}
\label{subsec:artis_control_algorithm}

\noindent To improve the practical applicability of ARTiS, the gripper was designed to support not only direct manual operation but also autonomous execution and teaching-based playback. The developed Python/ROS~2 compatible API provides a hardware abstraction layer for controlling the main ARTiS components: the base rotation joint $J_0$, the finger-axis orientation joints $J_1$-$J_3$, the four-bar closure joints $J_4$-$J_6$, and the jamming-palm pneumatic valve. Therefore, the gripper can be commanded using high-level actions such as opening the fingers, clearing the palm workspace, activating the jamming palm, closing the fingertips, and replaying a taught grasping sequence. The API is provided with the project repository~\cite{Project_Artis}.

The ARTiS command vector can be written as
\begin{equation}
\label{eq:artis_command_vector}
\mathbf{u}_{ARTiS}
=
\left[
q_0, q_1, q_2, q_3, q_4, q_5, q_6, s_p
\right]^T ,
\end{equation}
where $q_0$-$q_6$ are the commanded joint positions of $J_0$-$J_6$, and $s_p \in \{0,1\}$ is the jamming-palm state. Here, $s_p=0$ corresponds to the unjammed state, while $s_p=1$ corresponds to the jammed state.

For autonomous operation, ARTiS can be integrated with external vision sensing and a wrist-mounted force-torque sensor. The vision system detects the target tool, estimates its pose, identifies the handle or contact regions, and tracks the tool tip during task execution. The wrist force-torque sensor is used during tool insertion into the jamming palm to regulate the contact force, minimize undesired torques, and verify stable tool seating before jamming. The measured wrench is defined as
\begin{equation}
\label{eq:wrench_vector}
\mathbf{w}
=
\left[
F_x, F_y, F_z, M_x, M_y, M_z
\right]^T ,
\end{equation}
where $F_z$ is treated as the main palm-contact force, and $M_x$ and $M_y$ indicate misalignment between the tool and the palm. During tool seating, the controller seeks to satisfy
\begin{equation}
\label{eq:palm_contact_goal}
F_z \rightarrow F_d, 
\qquad
M_x \rightarrow 0,
\qquad
M_y \rightarrow 0,
\end{equation}
where $F_d$ is the desired contact force. A stable palm contact is accepted when
\begin{equation}
\label{eq:palm_contact_threshold}
|F_d-F_z| < \epsilon_F,
\qquad
|M_x| < \epsilon_M,
\qquad
|M_y| < \epsilon_M .
\end{equation}

The autonomous sequence is summarized in Algorithm~\ref{alg:artis_autonomous_control}. First, the target tool is detected, and its relevant grasping features are estimated. The gripper then opens the four-bar mechanisms and rotates the finger-axis joints to move the fingers outside the jamming-palm workspace. After the robot wrist moves to the tool approach pose, the tool is pressed into the jamming palm under force control. Once stable palm contact is detected, the palm is jammed, the tool is slightly lifted, and grasp stability is verified. The fingertips are then closed around the handle or selected contact points to provide additional support. During tool use, the vision system tracks the tool tip and, using movements, orients the finger joints to correct the tool tip pose error. Finally, the tool is moved to a release area, the fingers are opened while the palm remains jammed, the tool is placed on the release surface, and the jamming palm is released.

\begin{algorithm}[t]
\small
\caption{Autonomous ARTiS Tool Grasping and Use}
\label{alg:artis_autonomous_control}
\begin{algorithmic}[1]

\State \textbf{Input:} vision data $\mathcal{I}$, wrist wrench $\mathbf{w}$, task command $\mathcal{T}$
\State \textbf{Output:} tool grasping, tool use, and tool release

\State Detect the target tool and estimate its pose using vision sensing
\State Identify the tool handle, contact regions, and tool-tip pose
\State Select the grasping method according to the tool and task

\State Set the jamming palm to the unjammed state, $s_p \gets 0$
\State Open $J_4$-$J_6$ and rotate $J_1$-$J_3$ to clear the palm workspace
\State Move the robot wrist to the tool approach pose

\State Press the tool into the jamming palm under force control
\State Regulate the palm contact using $\mathbf{w}$ so that $F_z \rightarrow F_d$ and $M_x,M_y \rightarrow 0$

\If{stable palm contact is detected}
    \State Activate the jamming palm, $s_p \gets 1$
\Else
    \State Stop and report failure
\EndIf

\State Lift the tool slightly (depending on the tool size) and verify grasp stability using vision and force-torque feedback

\If{the tool is unstable or slipping}
    \State Disengage the jamming palm and report failure
\EndIf

\State Orient $J_1$-$J_3$ toward the selected handle/contact regions
\State Close $J_4$-$J_6$ around the tool for additional support

\While{task $\mathcal{T}$ is not complete}
    \State Track the tool-tip pose using vision
    \State Correct the robot wrist motion and finger orientation to reduce the tool-tip error
    \State Monitor force-torque safety limits and tool stability
\EndWhile

\State Move the tool to the release area
\State Open $J_4$-$J_6$ while keeping the jamming palm active
\State Place the tool on the release surface using force-torque feedback
\State Release the jamming palm, $s_p \gets 0$
\State Verify release and lift the gripper away

\end{algorithmic}
\end{algorithm}

\subsection{Teaching-Based Gripper Programming}
\label{subsec:artis_teaching_mode}

\noindent In addition to autonomous sensor-guided operation, ARTiS supports a teaching-based programming mode for repetitive tool-grasping tasks Algorithm~\ref{alg:artis_teach_replay}. In this mode, a human operator manually configures the gripper for a specific tool while the system records the ARTiS joint positions, jamming-palm state, timing information, and optional sensing data. The saved sequence can then be replayed automatically for repeated execution of the same grasping process or used as demonstration data for robot imitation learning.

Each recorded teaching step is represented as
\begin{equation}
\label{eq:teaching_step}
\mathcal{D}_k
=
\left\{
\mathbf{q}_k,
s_{p,k},
t_k,
\mathbf{w}_k,
{}^{C}\mathbf{T}_{tool,k}
\right\},
\end{equation}
where $\mathbf{q}_k=[q_{0,k},q_{1,k},...,q_{6,k}]^T$ is the ARTiS joint configuration at step $k$, $s_{p,k}$ is the jamming-palm state, $t_k$ is the timing information, $\mathbf{w}_k$ is the measured wrist wrench, and ${}^{C}\mathbf{T}_{tool,k}$ is the tool pose estimated by the vision system in the camera frame. If vision or force-torque sensing is not available, the corresponding data are omitted, and the step is recorded using only the gripper state.

A complete taught grasping primitive is stored as
\begin{equation}
\label{eq:teaching_sequence}
\mathcal{S}
=
\left[
\mathcal{D}_1,
\mathcal{D}_2,
...,
\mathcal{D}_K
\right],
\end{equation}
where $K$ is the number of recorded steps. During playback, the saved sequence is executed step by step. If vision sensing is available, the taught pose is adapted to the currently observed tool pose; otherwise, the recorded joint commands are replayed directly. If a wrist force-torque sensor is available, palm-contact steps can be refined by regulating the contact force before jamming. This allows the same teaching framework to operate either as a simple open-loop replay system or as a sensor-adaptive autonomous playback method.

\begin{algorithm}[t]
\small
\caption{Teaching and Playback of an ARTiS}
\label{alg:artis_teach_replay}
\begin{algorithmic}[1]

\State \textbf{Input:} manual operation, vision data $\mathcal{I}$, wrist wrench $\mathbf{w}$
\State \textbf{Output:} reusable ARTiS grasping sequence $\mathcal{S}$

\State Initialize an empty sequence $\mathcal{S}$

\While{teaching mode is active}
    \State Read current ARTiS state $\mathbf{q}_k$, palm state $s_{p,k}$, and time $t_k$
    \State Read sensing data $\mathbf{w}_k$ and ${}^{C}\mathbf{T}_{tool,k}$

    \If{operator requests recording}
        \State Store $\mathcal{D}_k=\{\mathbf{q}_k,s_{p,k},t_k,\mathbf{w}_k,{}^{C}\mathbf{T}_{tool,k}\}$
        \State Append $\mathcal{D}_k$ to $\mathcal{S}$
    \EndIf

    \If{operator requests jamming or release}
        \State Update jamming-palm state $s_p$
    \EndIf

    \If{operator requests manual repositioning}
        \State Disable motor torque
    \Else
        \State Enable torque and hold the current position
    \EndIf
\EndWhile

\If{$\mathcal{S}$ is empty}
    \State Stop and report failure
\Else
    \State Save $\mathcal{S}$ as a tool-grasping primitive
\EndIf

\Statex
\State \textbf{Playback:}
\State Load saved sequence $\mathcal{S}$

\If{target tool is detected}
    \State Adapt the taught approach pose to the current tool pose
\EndIf

\For{each step $\mathcal{D}_k$ in $\mathcal{S}$}
    \State Command ARTiS to the recorded configuration $\mathbf{q}_k$
    \State Set the jamming palm according to $s_{p,k}$

    \If{force-torque sensing is available during palm contact}
        \State Regulate contact so that $F_z \rightarrow F_d$ and $M_x,M_y \rightarrow 0$
    \EndIf

    \If{unsafe contact, slip, or unstable grasp is detected}
        \State Stop playback and report failure
    \EndIf
\EndFor

\State Verify the final tool state and report success or failure

\end{algorithmic}
\end{algorithm}

\section{Gripper Ablation Using Screwdrivers}\label{A1}

An analysis of the use of two screwdrivers under identical grasping conditions, but the screwdrivers have different geometric parameters: the first had a total length $L = 230$ (Screwdriver 2), while the second had $L = 180$ mm (Screwdriver 4). All other geometric and contact parameters, finger position $L_1 = 120$ mm, palm radius $r_p = 40$ mm, and tool radius $r_f = 3.25$ mm, were kept constant. For each configuration, the maximum applied force at point $A$ was used to compute the moment $M = F_A(L-L_1)$, and the equivalent friction in translation and rotation was obtained
\begin{align}
    (\mu F)_t = \frac{M_{O}^{t}}{H_1}, 
    ~ ~ ~ ~ ~ ~ ~
    (\mu F)_r = \frac{M_{O}^{r}}{r}.
\end{align}

The combined Palm+Finger configuration was compared against the Palm-only and Finger-only configurations to quantify, in Table~\ref{tab:tool_comparison}: (a) palm and finger contributions in each failure mode, (b) the governing mode of slip, and (c) ablation-based safety factors.

\paragraph{Palm and Finger Contributions}
The longer screwdriver 2 produces a larger moment $M$ for any given force $F_A$, and the measured moment at the slip was $\bar M_{\mathrm{comb}} = 0.603~\mathrm{N·m}$, with a corresponding translational friction equivalent of $(\mu F)_{\mathrm{trans}} = 5.022~\mathrm{N}$. For the shorter screwdriver 4, the arm $H_A = H-H_1$ (Fig.~\ref{subfig7_5}) decreases from $110$~mm to $60$~mm. As a result, for a comparable slip force $F_A$, the generated moment is smaller. The measured moment at slip is $\bar M_{\mathrm{comb}} = 0.705~\mathrm{N·m}$, yielding
$(\mu F)_{\mathrm{trans}} = 5.879~\mathrm{N}$, slightly higher than in screwdriver 2. Palm and finger contributions in each slip mode (translational) are computed as
\[
C_{\mathrm{p,trans}} 
= 
\frac{(\mu F)_{\mathrm{trans,p}}}
{(\mu F)_{\mathrm{trans,p}} + (\mu F)_{\mathrm{trans,f}}},
\]
\[
C_{\mathrm{f,trans}} 
= 
\frac{(\mu F)_{\mathrm{trans,f}}}
{(\mu F)_{\mathrm{trans,p}} + (\mu F)_{\mathrm{trans,f}}},
\]
and for rotation:
\[
C_{\mathrm{p,rot}} 
=
\frac{(\mu F)_{\mathrm{rot,p}}}
{(\mu F)_{\mathrm{rot,p}} + (\mu F)_{\mathrm{rot,f}}},
\qquad \]
\[
C_{\mathrm{f,rot}} 
=
\frac{(\mu F)_{\mathrm{rot,f}}}
{(\mu F)_{\mathrm{rot,p}} + (\mu F)_{\mathrm{rot,f}}}.
\]

\begin{table}[!t]
\centering
\caption{Comparison of slip mechanics, mode contributions, and safety factors for the tested screwdrivers}
\label{tab:tool_comparison}
\resizebox{\columnwidth}{!}{%
\begin{tabular}{lcc}
\hline
\textbf{Metric} & \textbf{Screwdriver 2} & \textbf{Screwdriver 4} \\
\hline

Translational friction $(\mu F)_{\mathrm{trans}}$ 
    & 5.022 N & 5.879 N \\

Moment at slip $\bar M_{\mathrm{comb}}$
    & 0.603 N·m & 0.705 N·m \\

Palm trans. contribution $C_{\mathrm{p,trans}}^\ast$
    & 92.07$\%$ & 97.4$\%$ \\

Finger trans. contribution $C_{\mathrm{f,trans}}^\ast$
    & 7.93$\%$ & 2.6$\%$ \\[1mm]

Palm rot. contribution $C_{\mathrm{p,rot}}^\ast$
    & 48.55$\%$ & 75.1$\%$ \\

Finger rot. contribution $C_{\mathrm{f,rot}}^\ast$
    & 51.45$\%$ & 24.9$\%$ \\[1mm]

Governing failure mode 
    & Trans. & Trans. \\

Global palm contribution $C_{\mathrm{p,global}}$
    & 92$\%$ & 97$\%$ \\

Global finger contribution $C_{\mathrm{f,global}}$
    & 8$\%$ & 3$\%$ \\[1mm]

Ablation trans. safety factor $\eta_T^{\mathrm{abl}}$
    & 0.724 & 1.02 \\

Ablation rot. safety factor $\eta_R^{\mathrm{abl}}$
    & 0.111 & 0.11 \\

Global ablation safety factor $\eta_{\mathrm{global}}^{\mathrm{abl}}$
    & 0.111 & 0.11 \\

\hline
\end{tabular}}
\end{table}

\paragraph{Governing mode}
The slip is governed if
\[
(\mu F)_{\mathrm{trans,comb}}
<
(\mu F)_{\mathrm{rot,p}} + (\mu F)_{\mathrm{rot,f}},
\]
and otherwise by rotation. The rotational scale $(\mu F)_{\mathrm{rot,p}} + (\mu F)_{\mathrm{rot,f}} \approx 20.7~\mathrm{N}$ is much larger than the translation requirement $5.022~\mathrm{N}$, indicating that screwdriver 2 fails translationally. For screwdriver 4, the rotational scale remains much larger than the translational requirement $(\mu F)_{\mathrm{rot,comb}}^{\mathrm{abl}} \approx 23.3~\mathrm{N} \gg (\mu F)_{\mathrm{trans,comb}} \approx 5.88~\mathrm{N}$; therefore, screwdriver 4 also fails by translation. The global contributions of the palm and finger follow the governing mode:
\[
C_{\mathrm{p,global}} = 
\begin{cases}
C_{\mathrm{p,trans}}^\ast & \text{if translational slip}, \\
C_{\mathrm{p,rot}}^\ast   & \text{if rotational slip},
\end{cases}
\]
with an analogous expression for $C_{\mathrm{f,global}}$.

\paragraph{Ablation safety factors}
Ablation compares the measured combined configuration to the linear sum  of isolated Palm-only and Finger-only abilities:
\[ \eta_{T}^{\mathrm{abl}} 
= 
\frac{(\mu F)_{\mathrm{p}}^{t} + (\mu F)_{\mathrm{f}}^{t}}
     {(\mu F)_{\mathrm{trans,comb}}},\]
\[ \eta_{R}^{\mathrm{abl}} 
= 
\frac{(\mu F)_{\mathrm{p}}^{r} + (\mu F)_{\mathrm{f}}^{r}}
     {(\mu F)_{\mathrm{rot,comb}}}, \]
and the global safety factor is
\[ \eta_{\mathrm{global}}
=
\min\bigl(\eta_{T}^{\mathrm{abl}},\, \eta_{R}^{\mathrm{abl}}\bigr). \]


Ablation of screwdriver 2 predicts $\eta_T^{\mathrm{abl}} = 0.724$ and $\eta_R^{\mathrm{abl}} = 0.111$, leading to a conservative global value $\eta_{\mathrm{global}} = 0.111$. Since $\eta_T^{\mathrm{abl}} < 1$, the combined configuration outperforms the linear superposition of isolated contacts, suggesting a synergistic palm–finger interaction. For screwdriver 4 $\eta_T^{\mathrm{abl}} \approx 1 \Rightarrow$ linear addition predicts translational capacity well. This suggests that the larger moment-arm geometry enables beneficial nonlinear interactions (micro-slip suppression, enhanced normal force distribution) that are less pronounced in the shorter screwdriver. Slip is governed by translation, palm dominance is strong but not absolute, and the combined configuration provides more resistance than predicted by isolated-contact tests, indicating nonlinear beneficial interactions.


\begin{figure}[!t]
    \centering
    \includegraphics[width=1\linewidth, clip, trim=0pt 0pt 0pt 0pt]{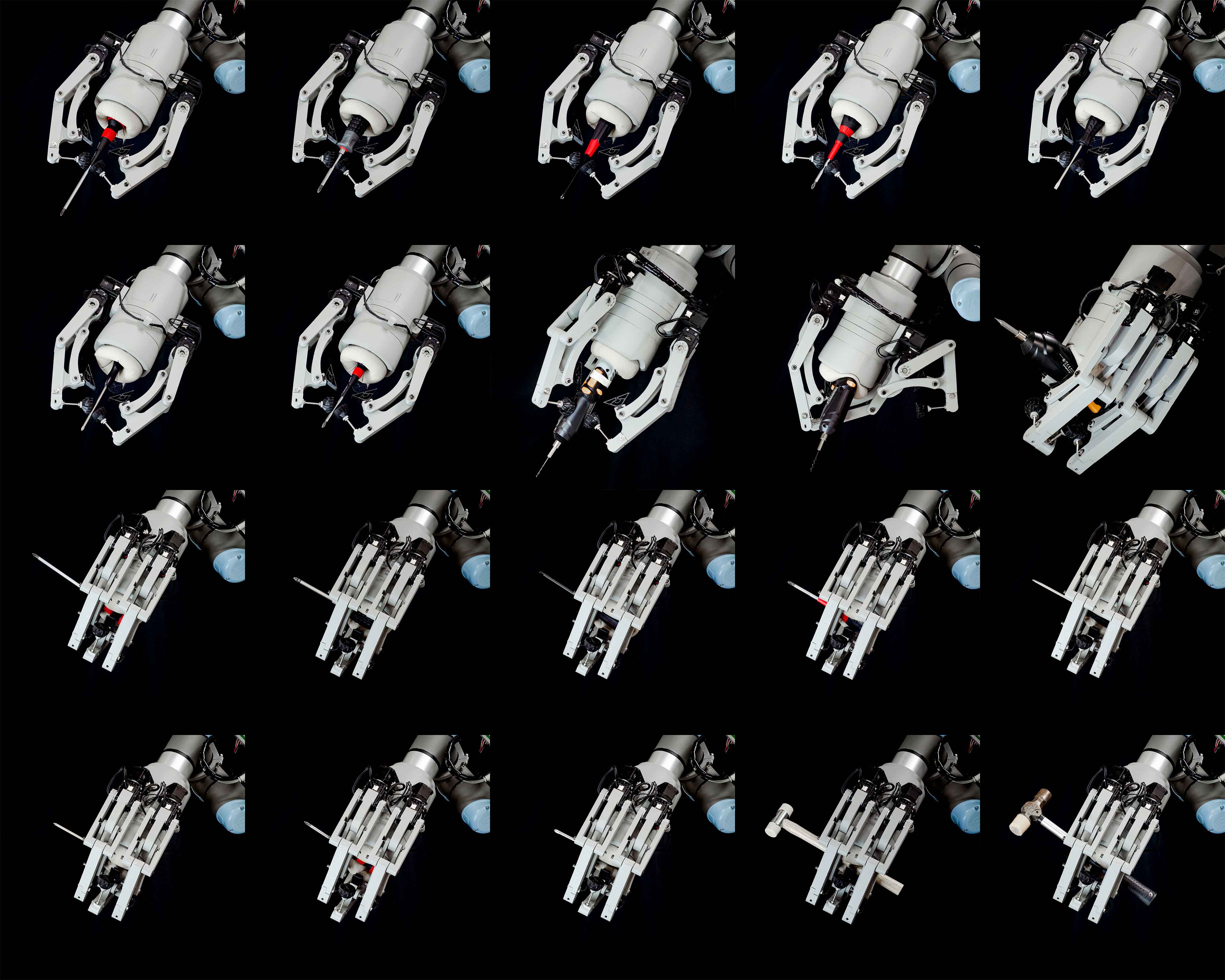}
    \caption{Power sphere and wrap holding the study tools by ARTiS}
    \label{fig_A2}
\end{figure}

\begin{figure}[!t]
\centering
        \includegraphics[width=0.19\linewidth, clip, trim=0pt 0pt 0pt 0pt]{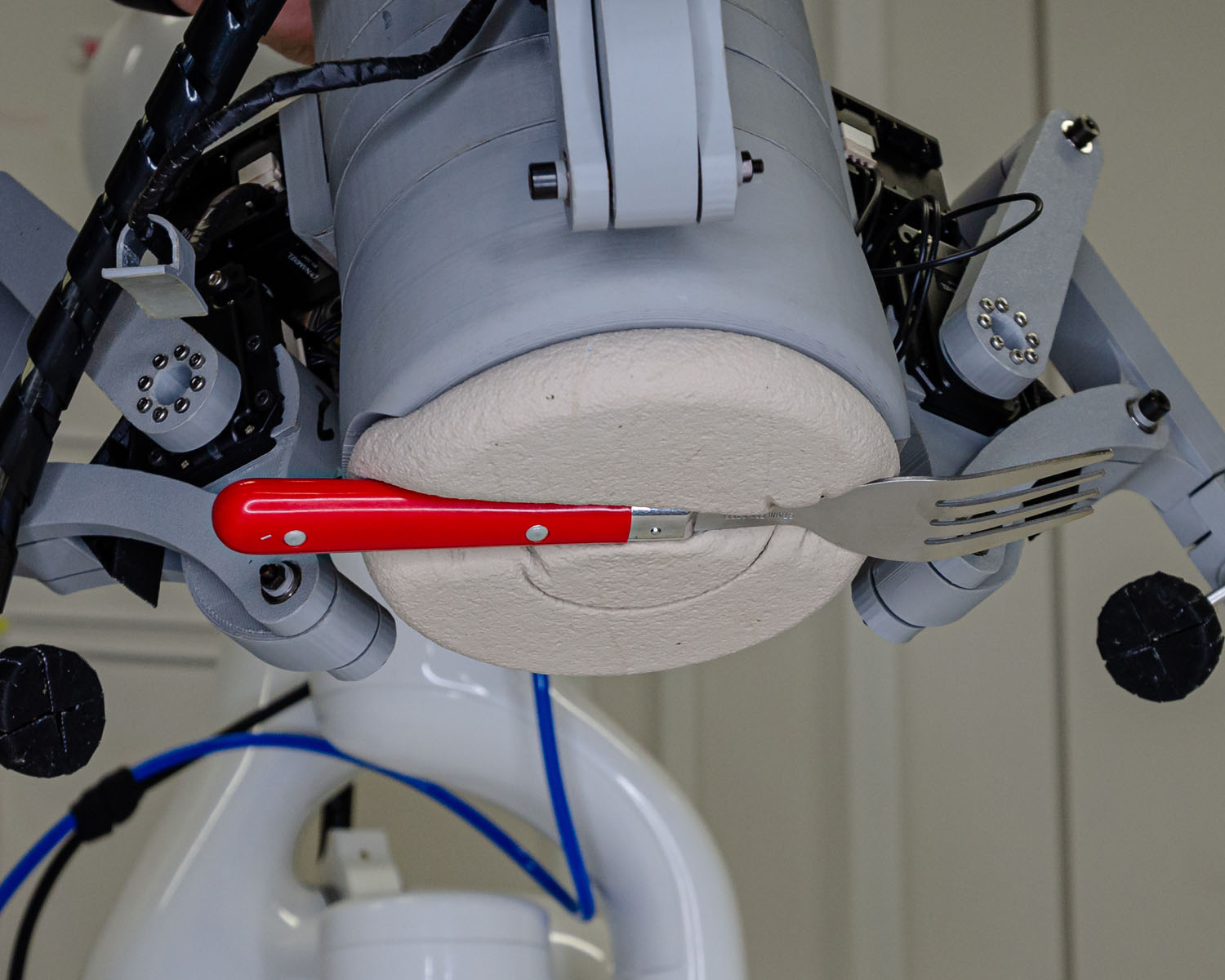}
        \includegraphics[width=0.19\linewidth, clip, trim=0pt 0pt 0pt 0pt]{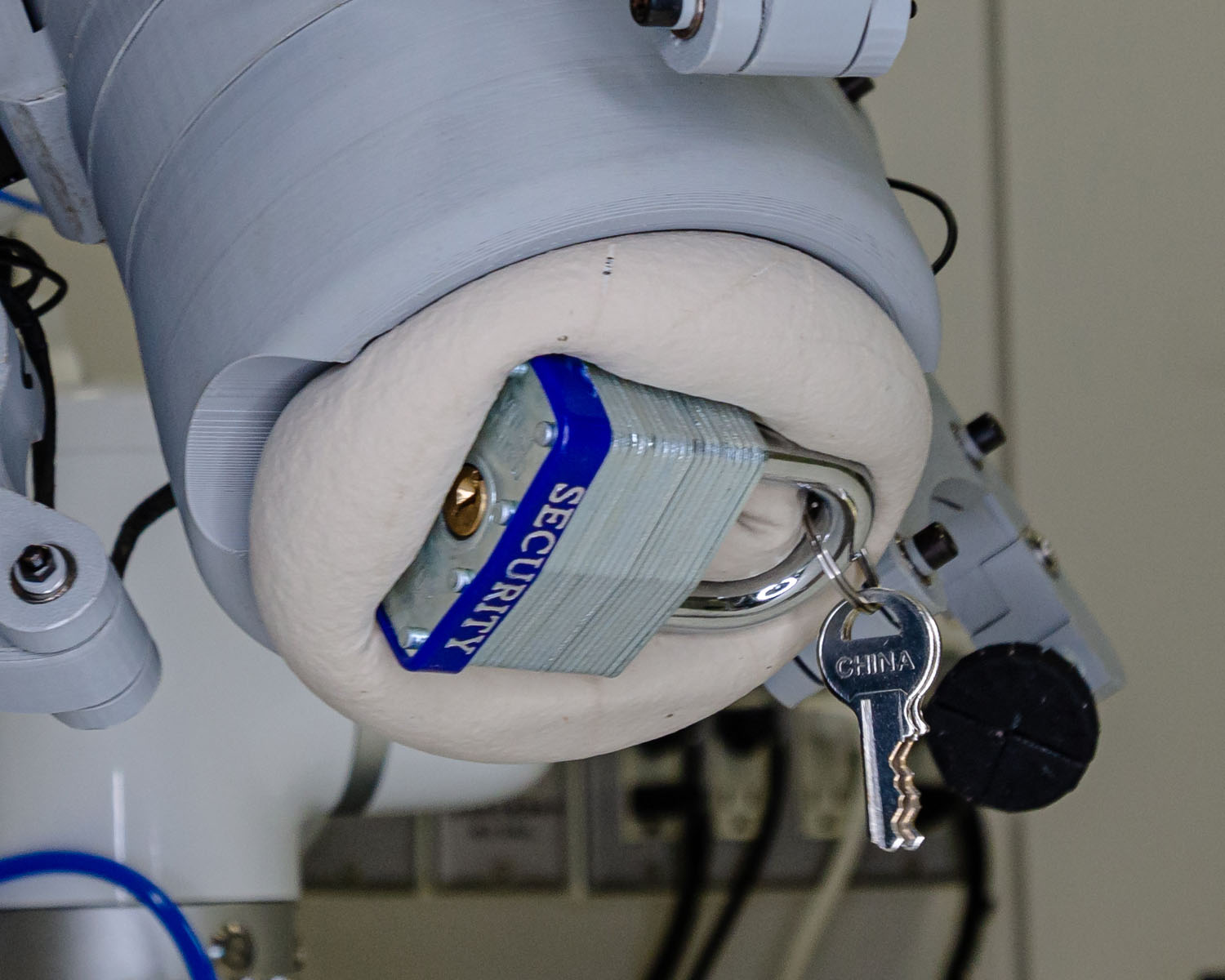}
        \includegraphics[width=0.19\linewidth, clip, trim=0pt 0pt 0pt 0pt]{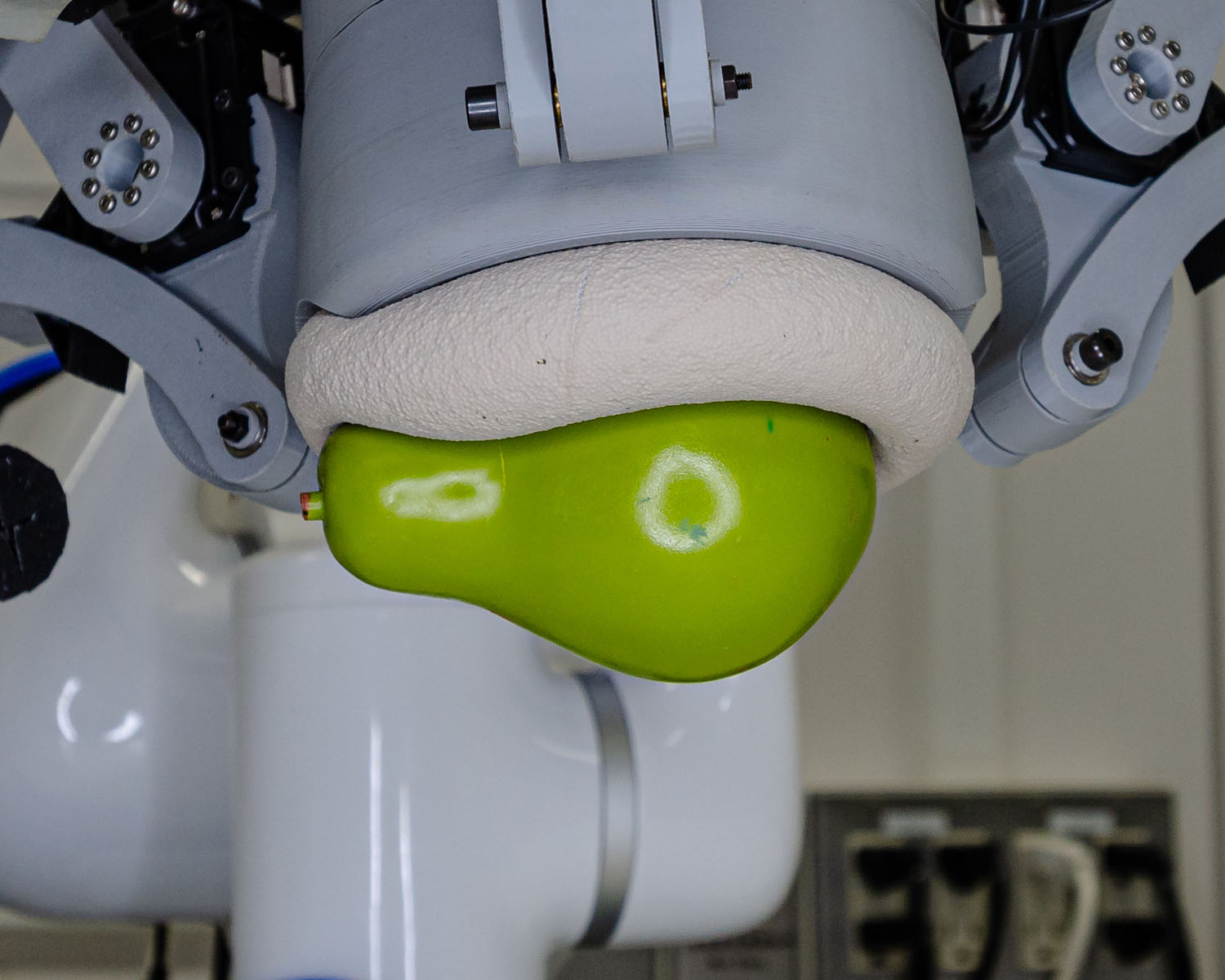}
        \includegraphics[width=0.19\linewidth, clip, trim=0pt 0pt 0pt 0pt]{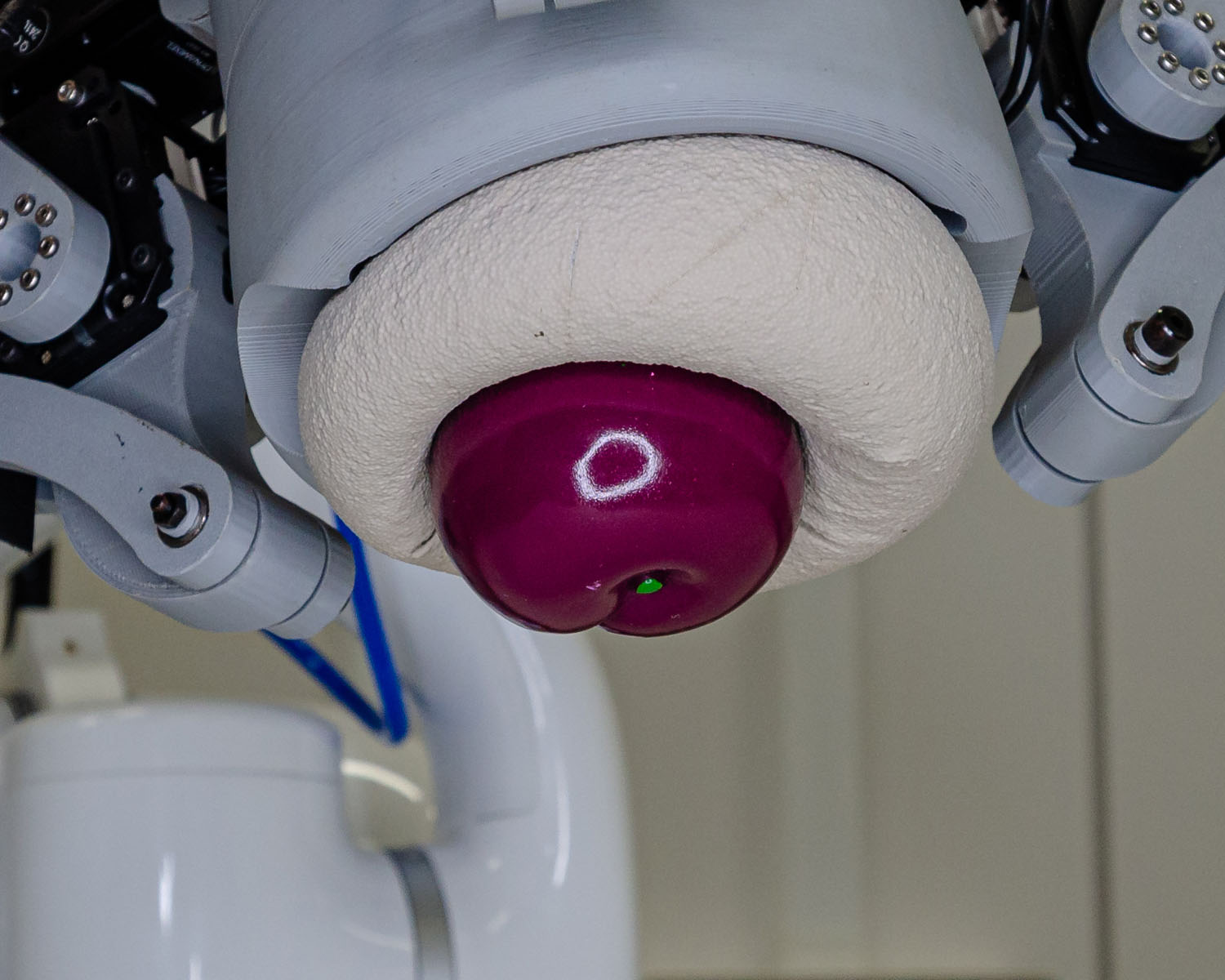}
        \includegraphics[width=0.19\linewidth, clip, trim=0pt 0pt 0pt 0pt]{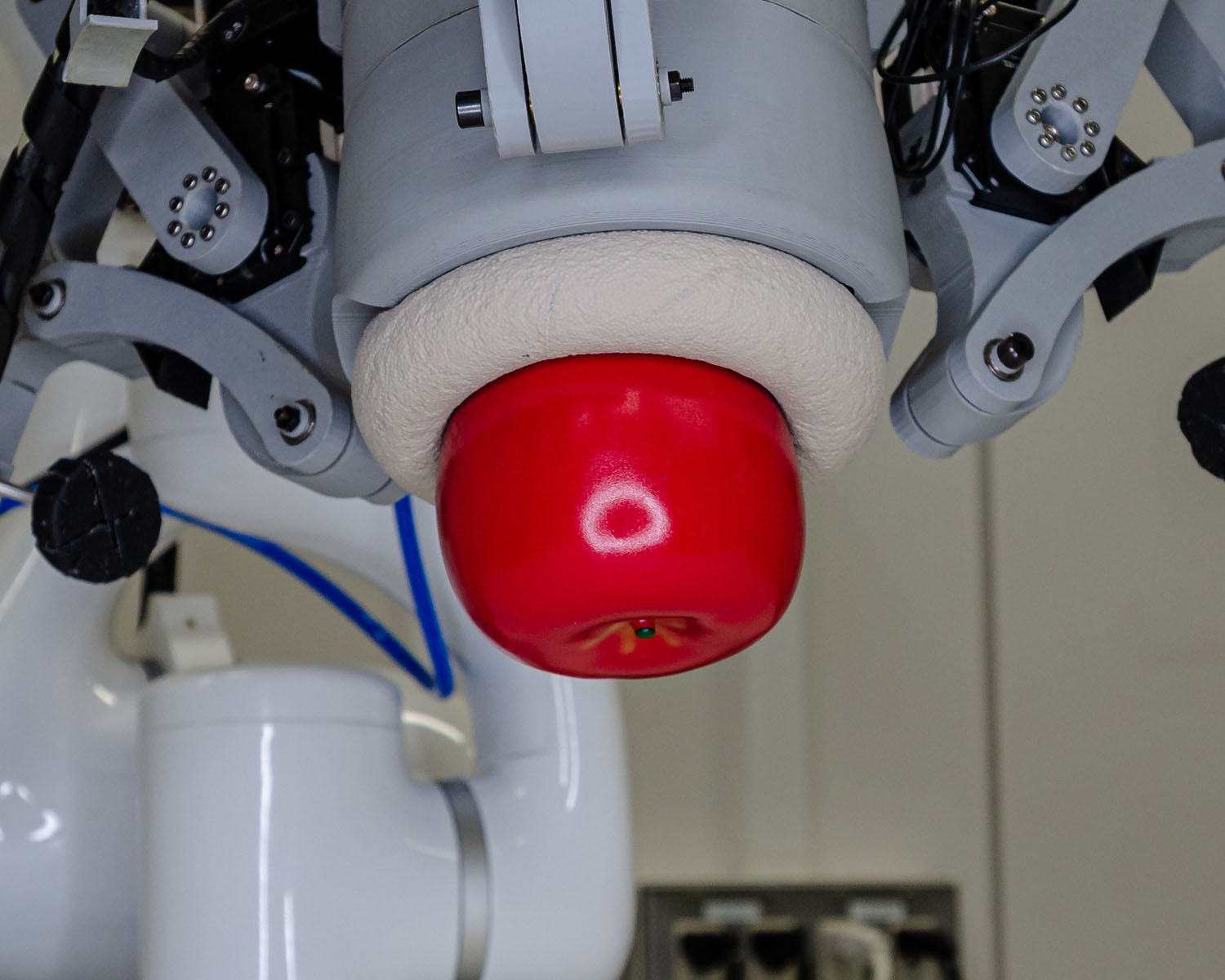}
        \\
        \vspace{1.2mm}
        \includegraphics[width=0.19\linewidth, clip, trim=0pt 0pt 0pt 0pt]{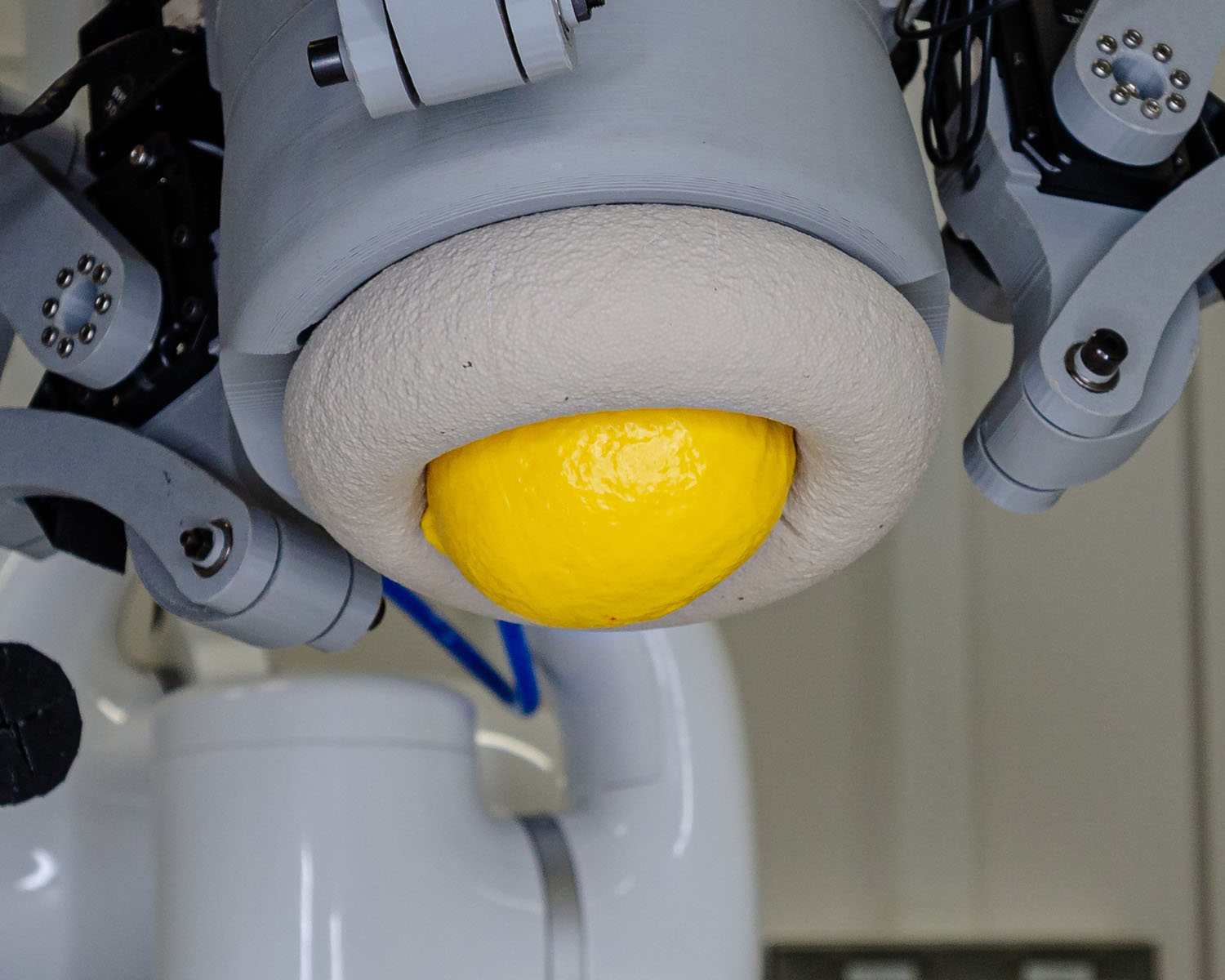}
        \includegraphics[width=0.19\linewidth, clip, trim=0pt 0pt 0pt 0pt]{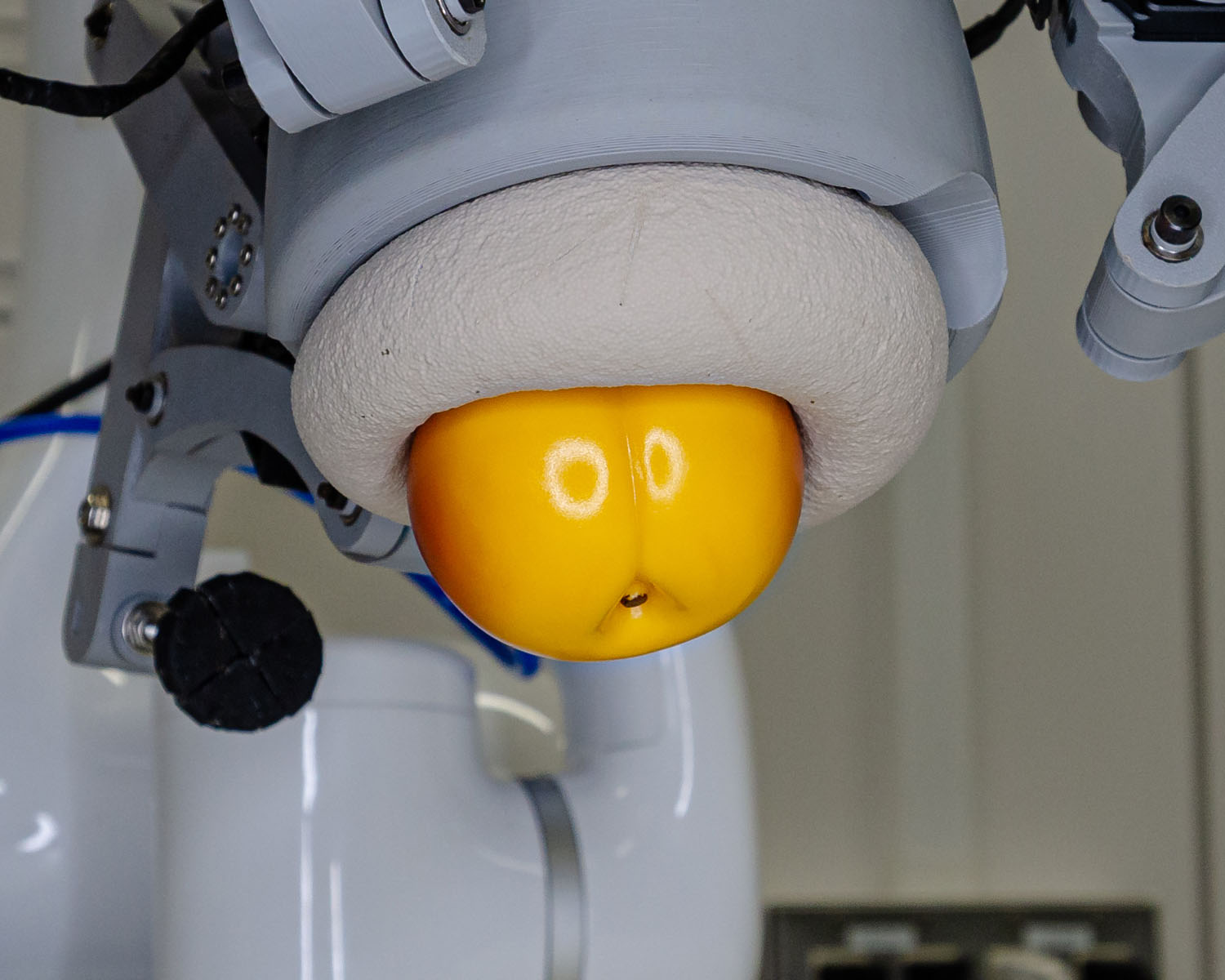}
        \includegraphics[width=0.19\linewidth, clip, trim=0pt 0pt 0pt 0pt]{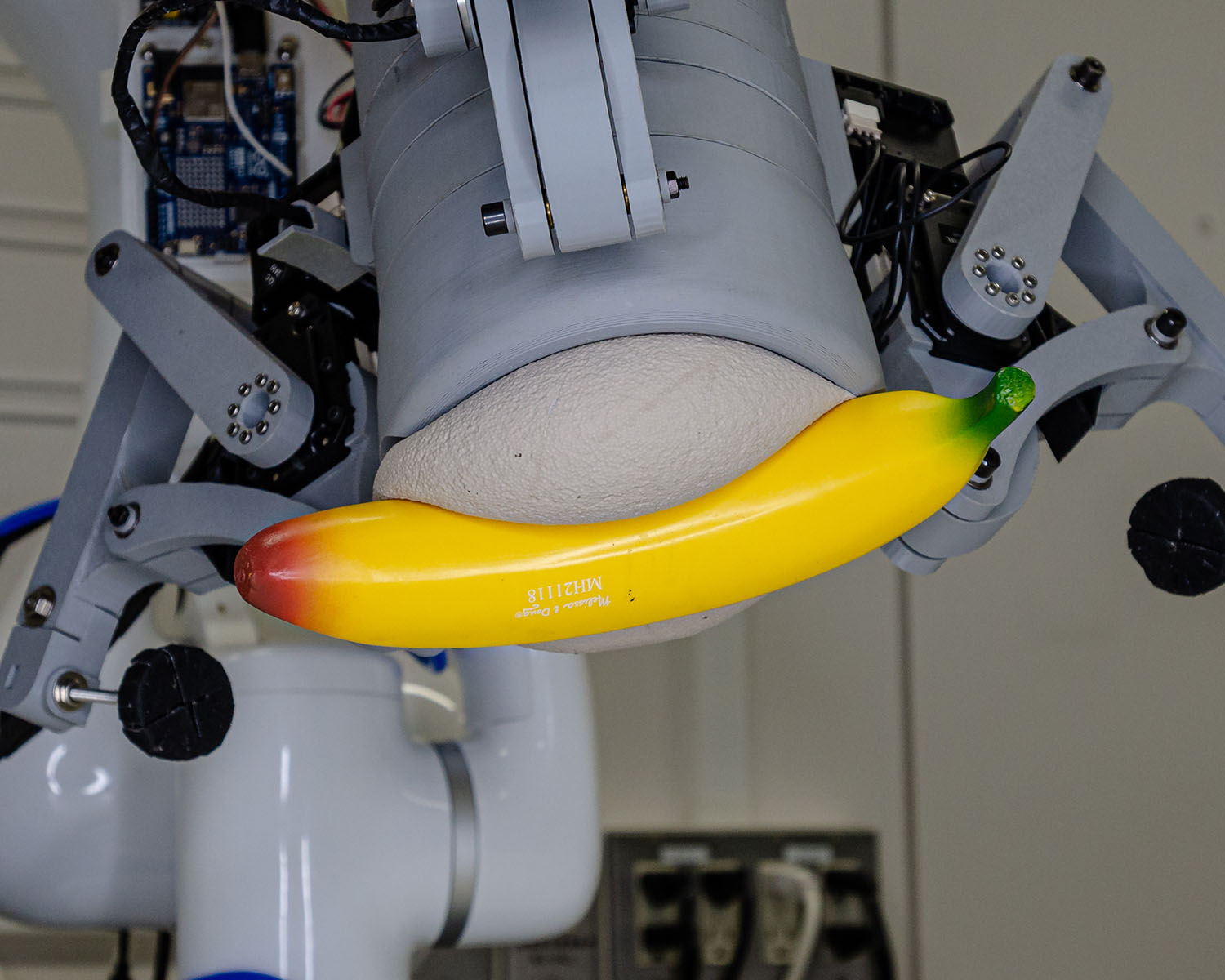}
        \includegraphics[width=0.19\linewidth, clip, trim=0pt 0pt 0pt 0pt]{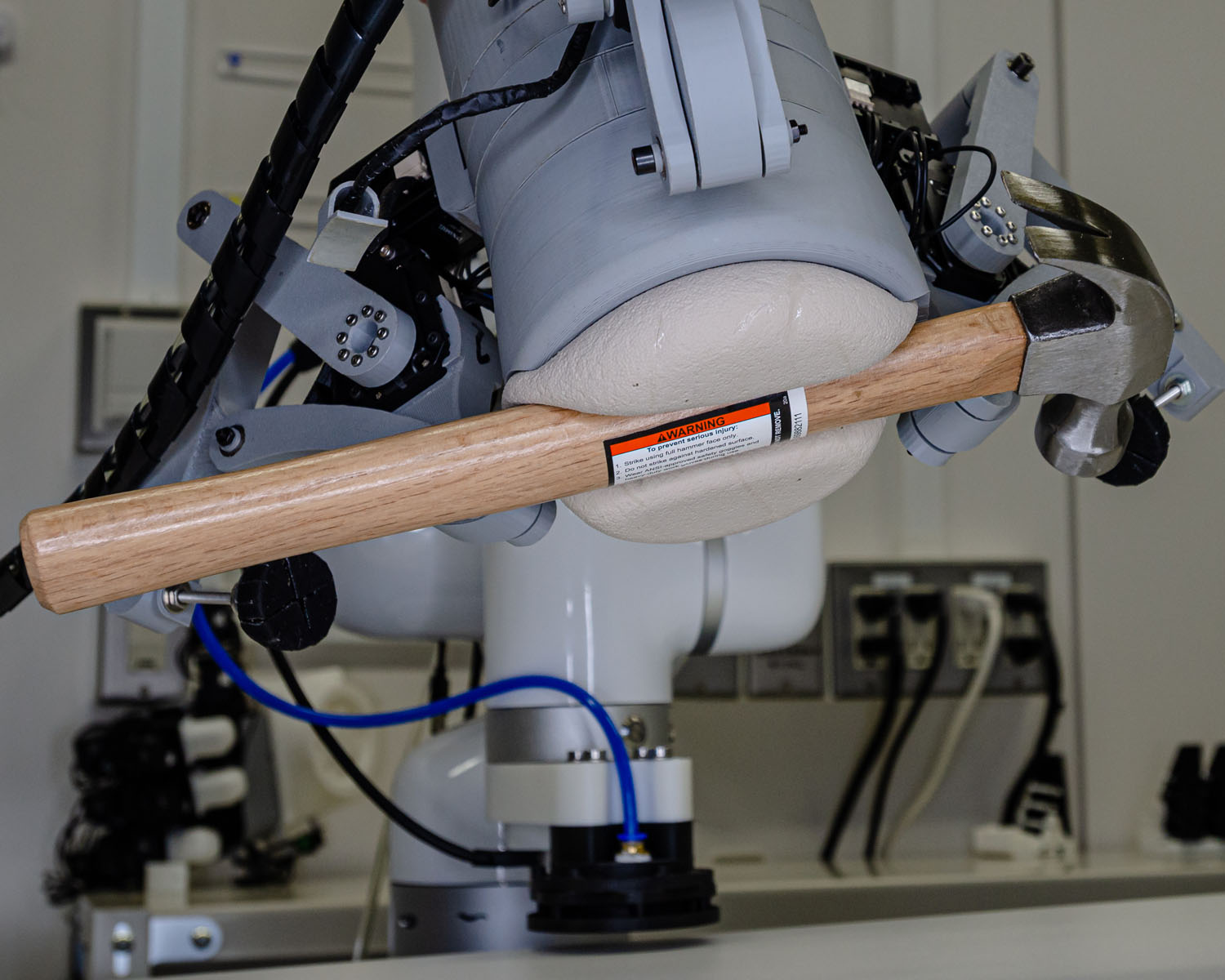}
        \includegraphics[width=0.19\linewidth, clip, trim=0pt 0pt 0pt 0pt]{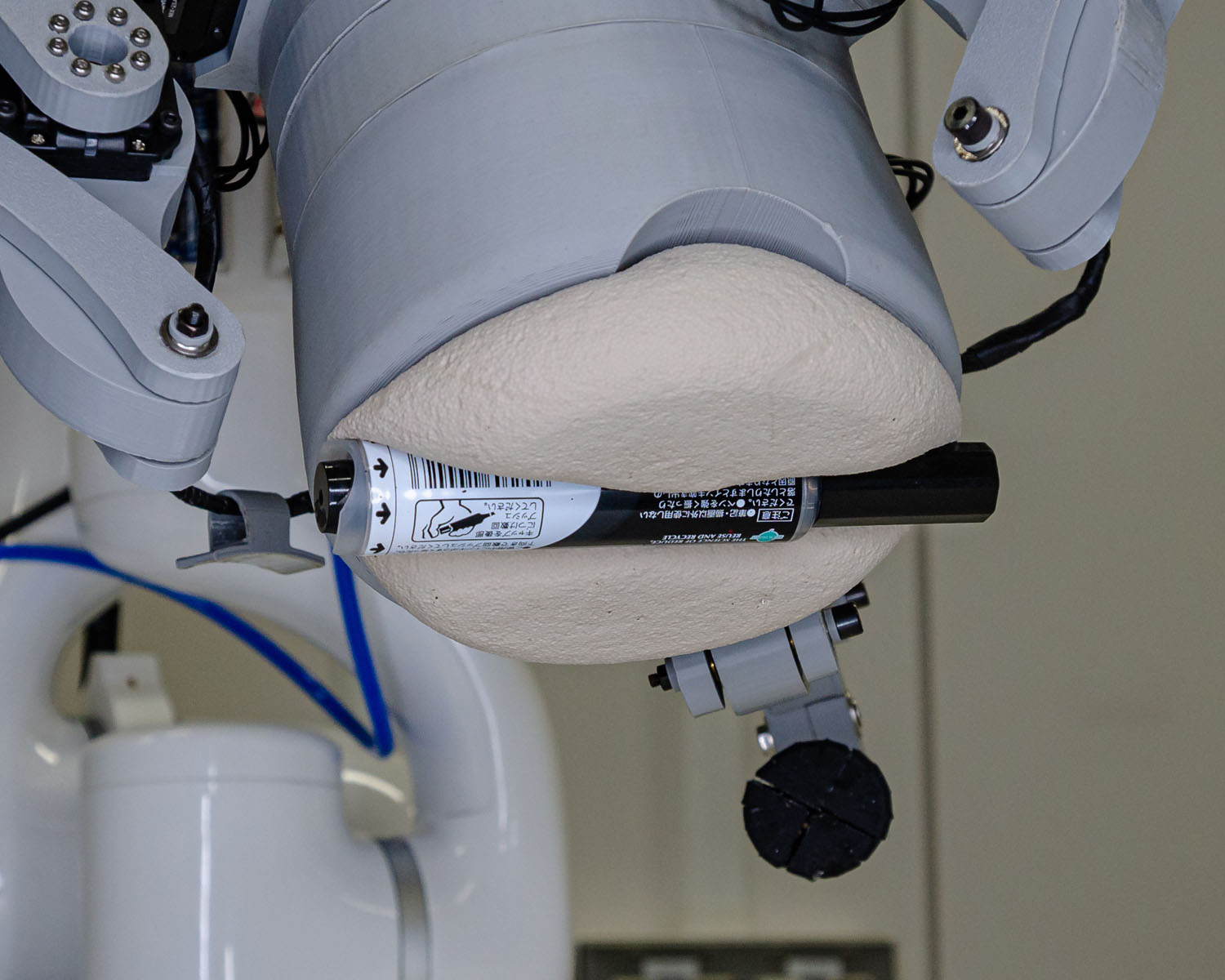}
        \\
        \vspace{1.2mm}
        \includegraphics[width=0.19\linewidth, clip, trim=0pt 0pt 0pt 0pt]{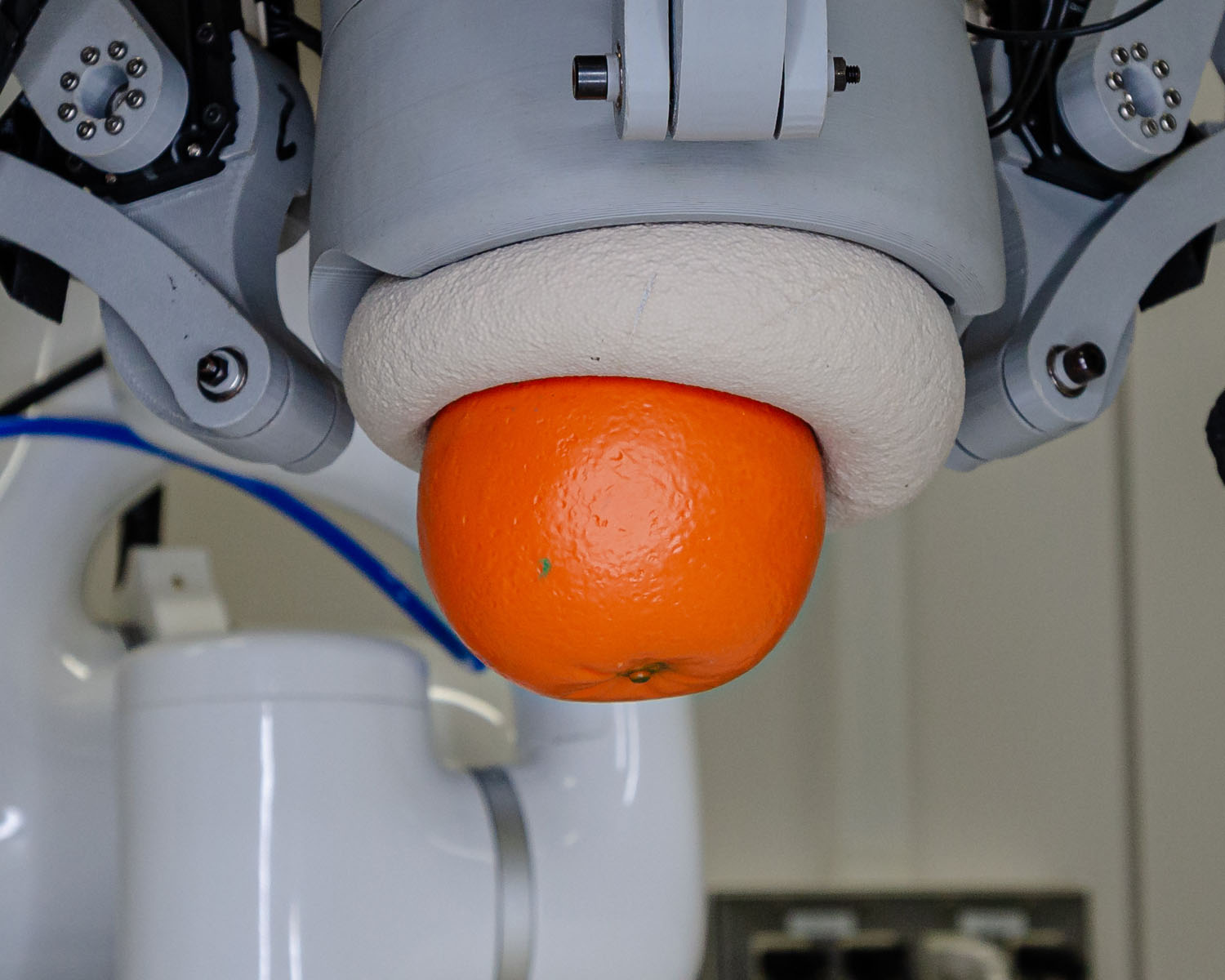}
        \includegraphics[width=0.19\linewidth, clip, trim=0pt 0pt 0pt 0pt]{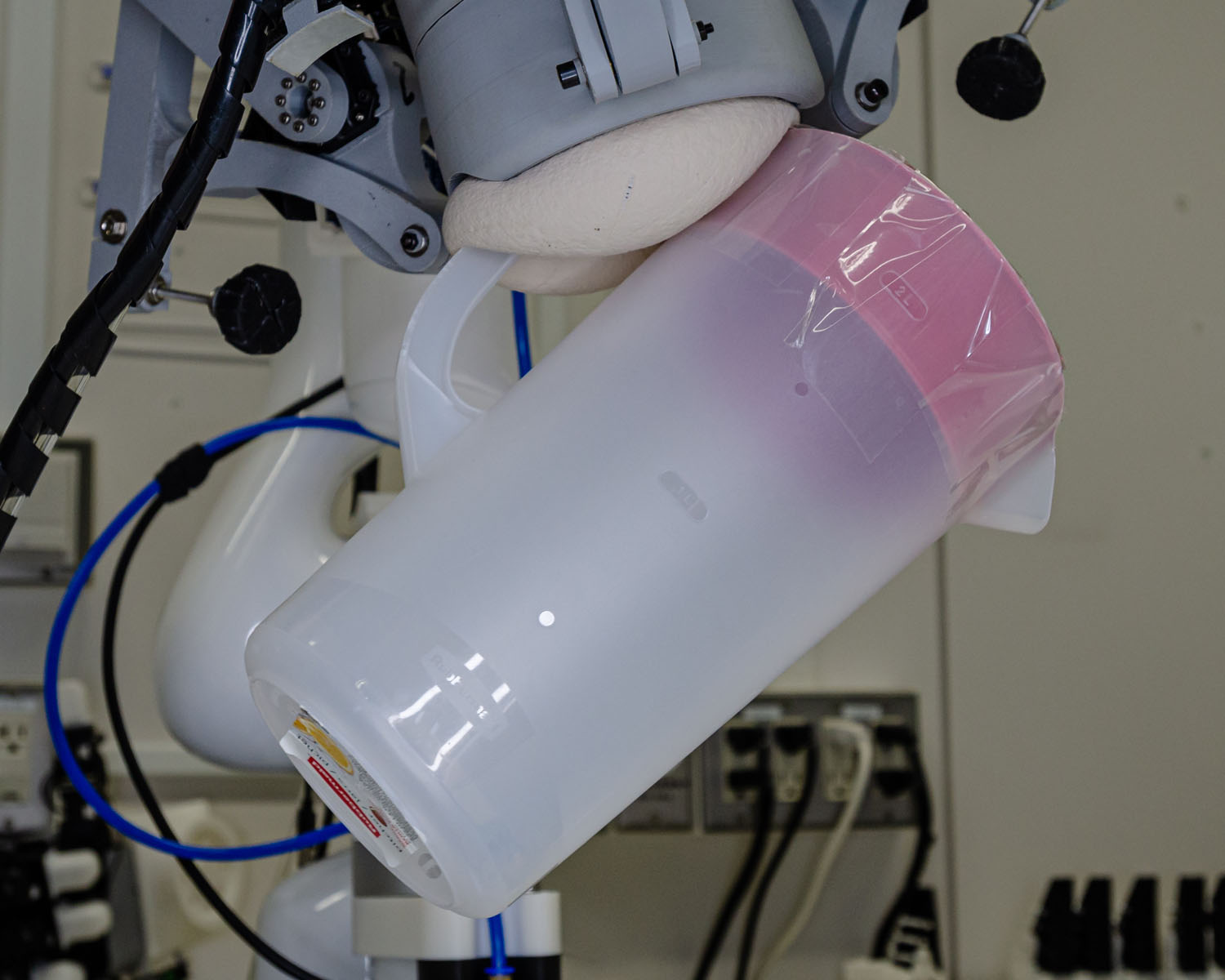}
        \includegraphics[width=0.19\linewidth, clip, trim=0pt 0pt 0pt 0pt]{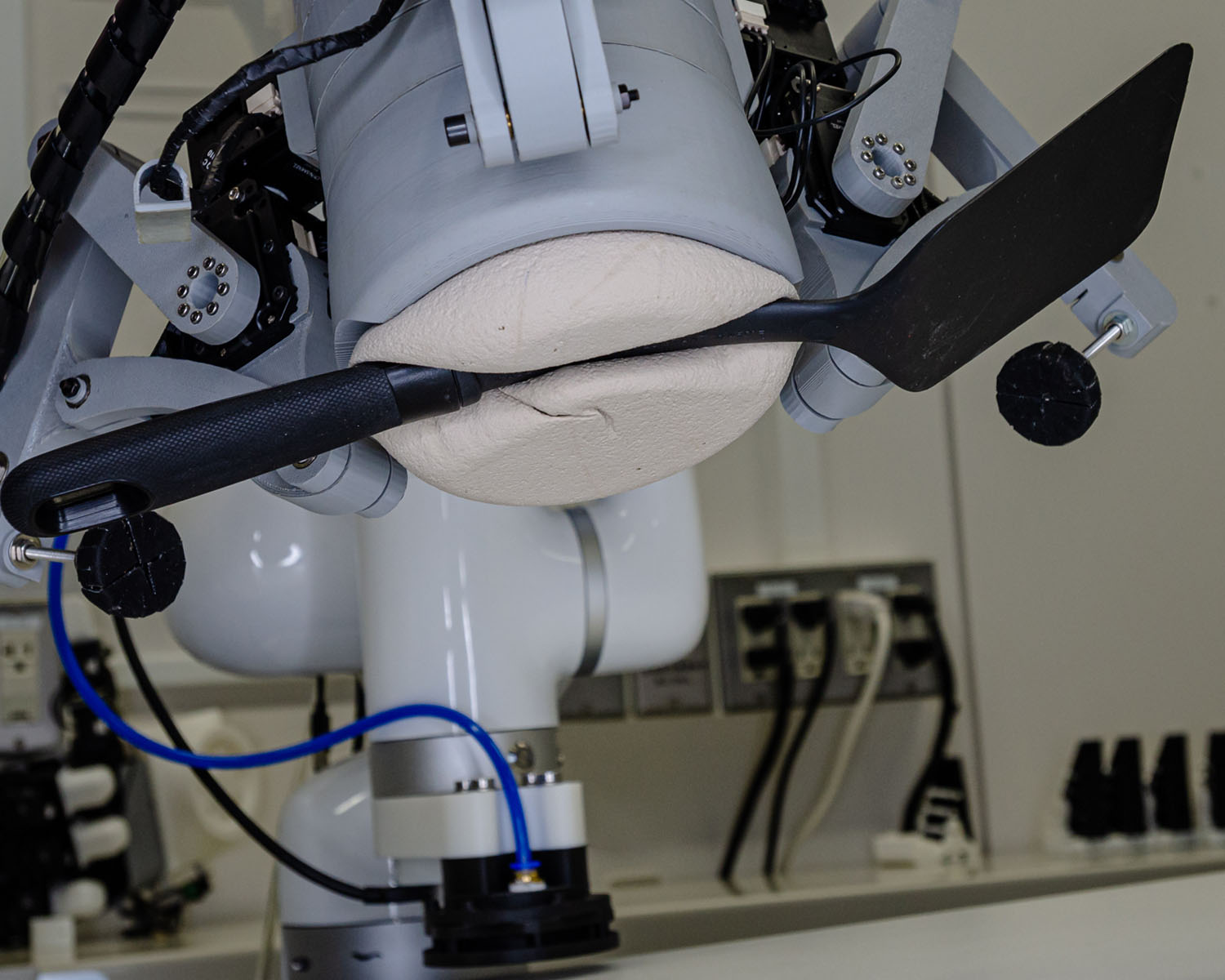}
        \includegraphics[width=0.19\linewidth, clip, trim=0pt 0pt 0pt 0pt]{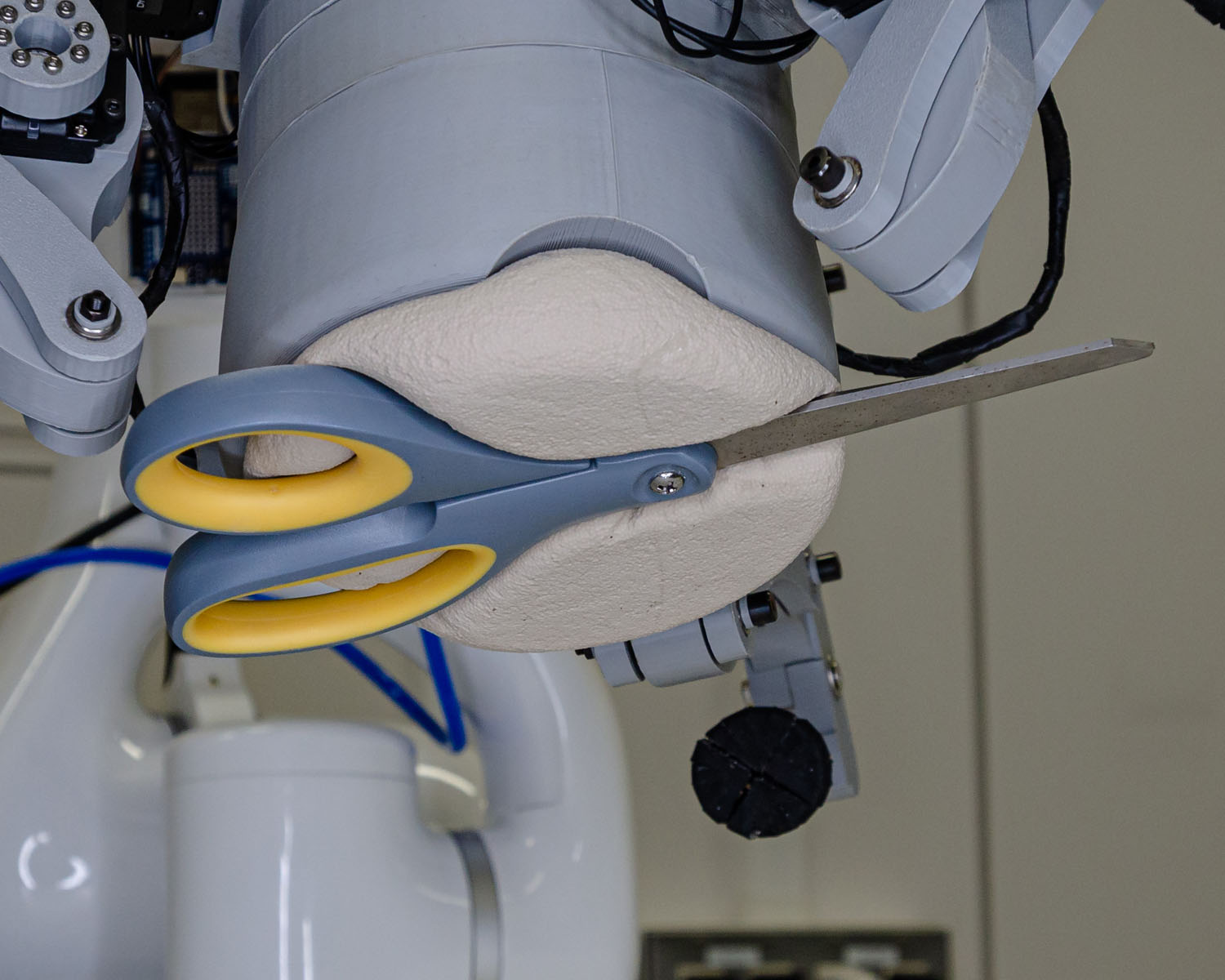}
        \includegraphics[width=0.19\linewidth, clip, trim=0pt 0pt 0pt 0pt]{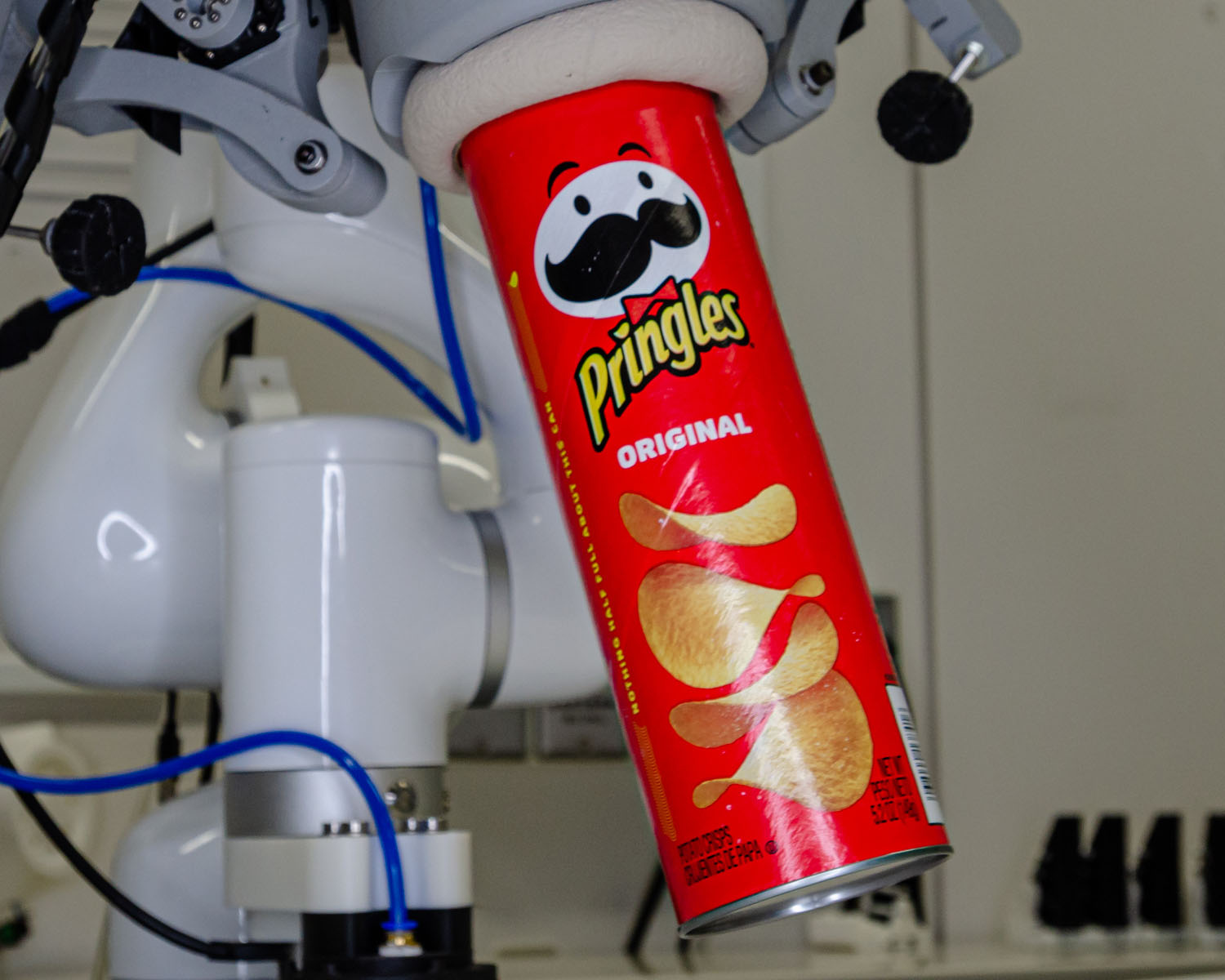}
        \\
        \vspace{1.2mm}
        \includegraphics[width=0.19\linewidth, clip, trim=0pt 0pt 0pt 0pt]{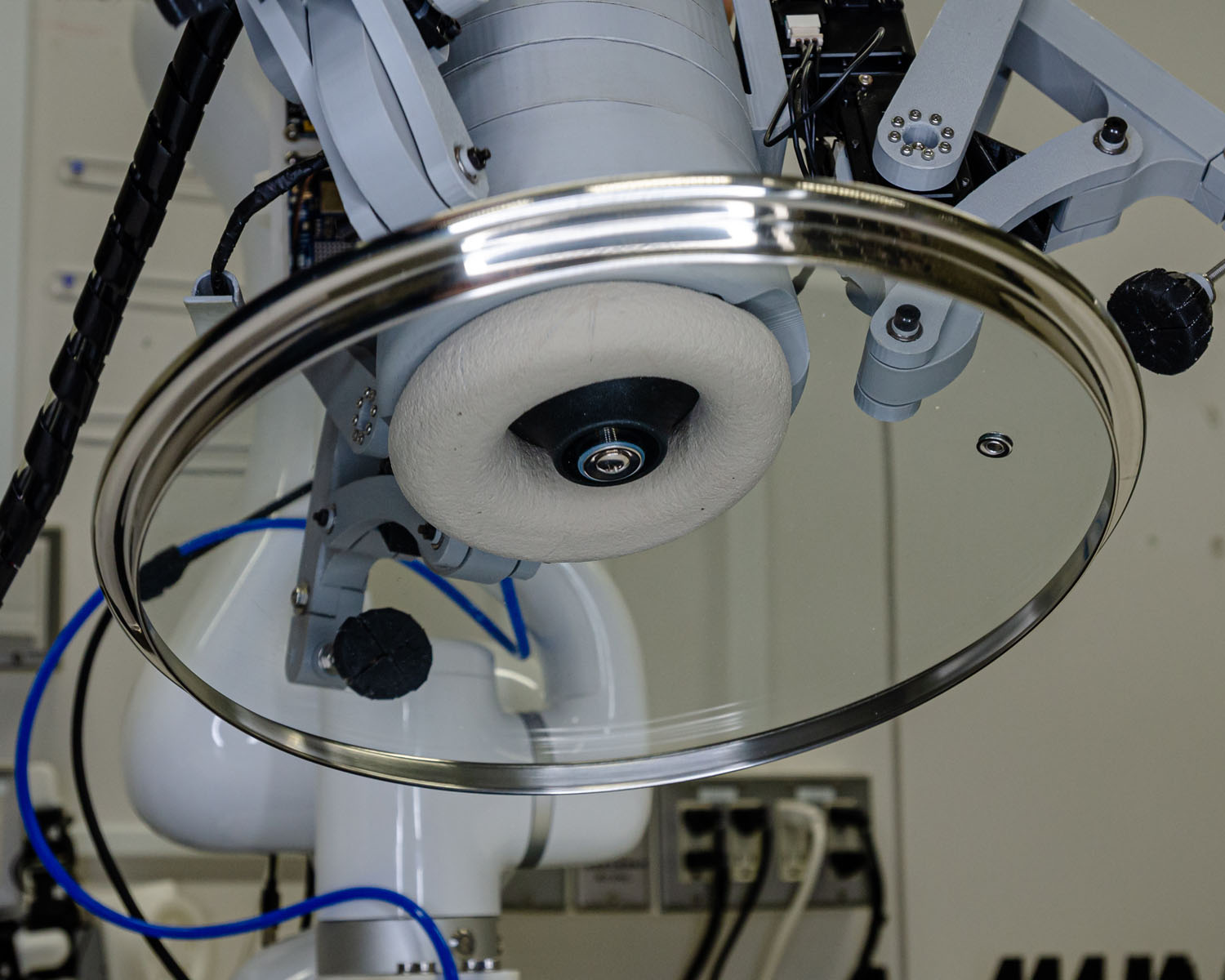}
        \includegraphics[width=0.19\linewidth, clip, trim=0pt 0pt 0pt 0pt]{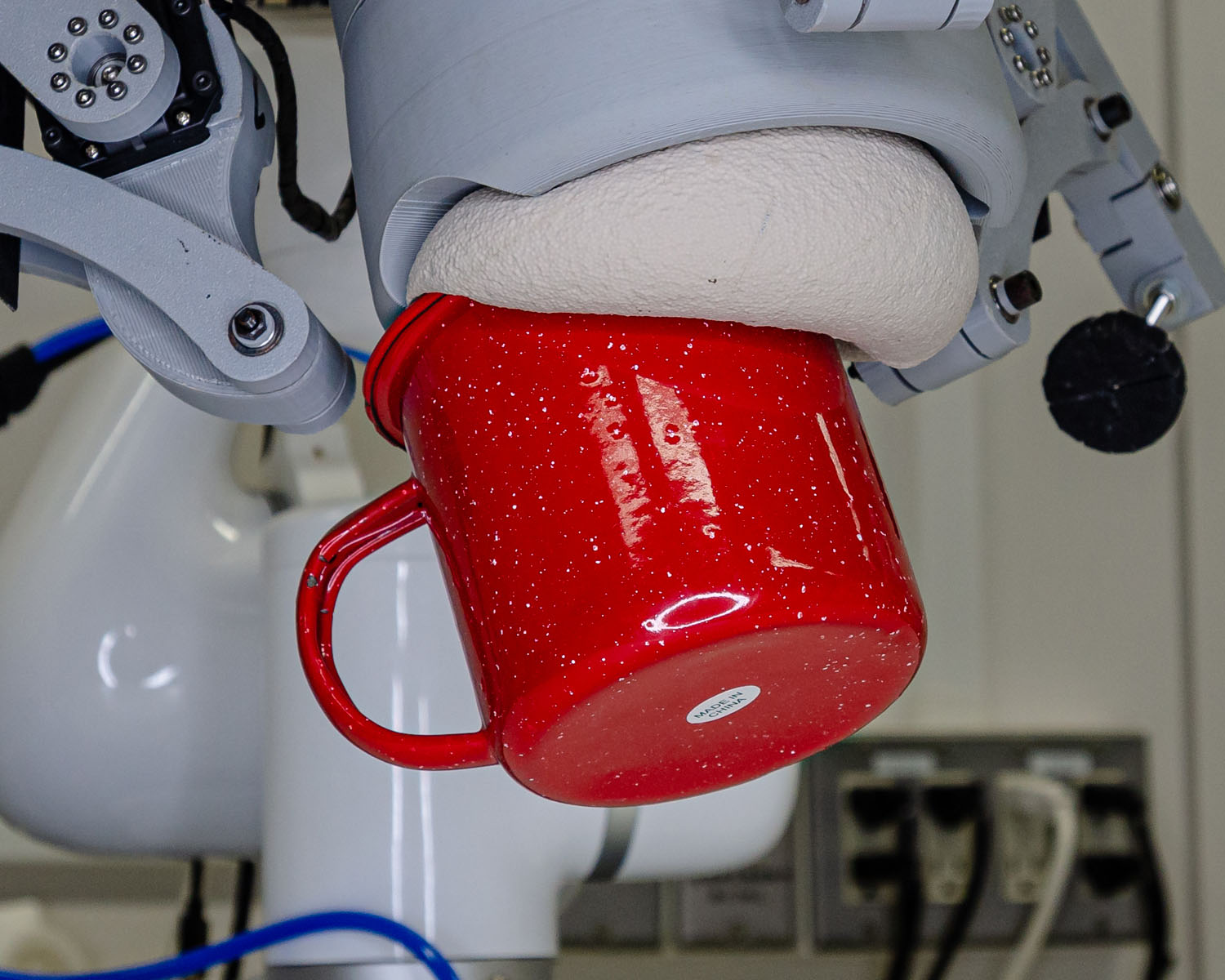}
        \includegraphics[width=0.19\linewidth, clip, trim=0pt 0pt 0pt 0pt]{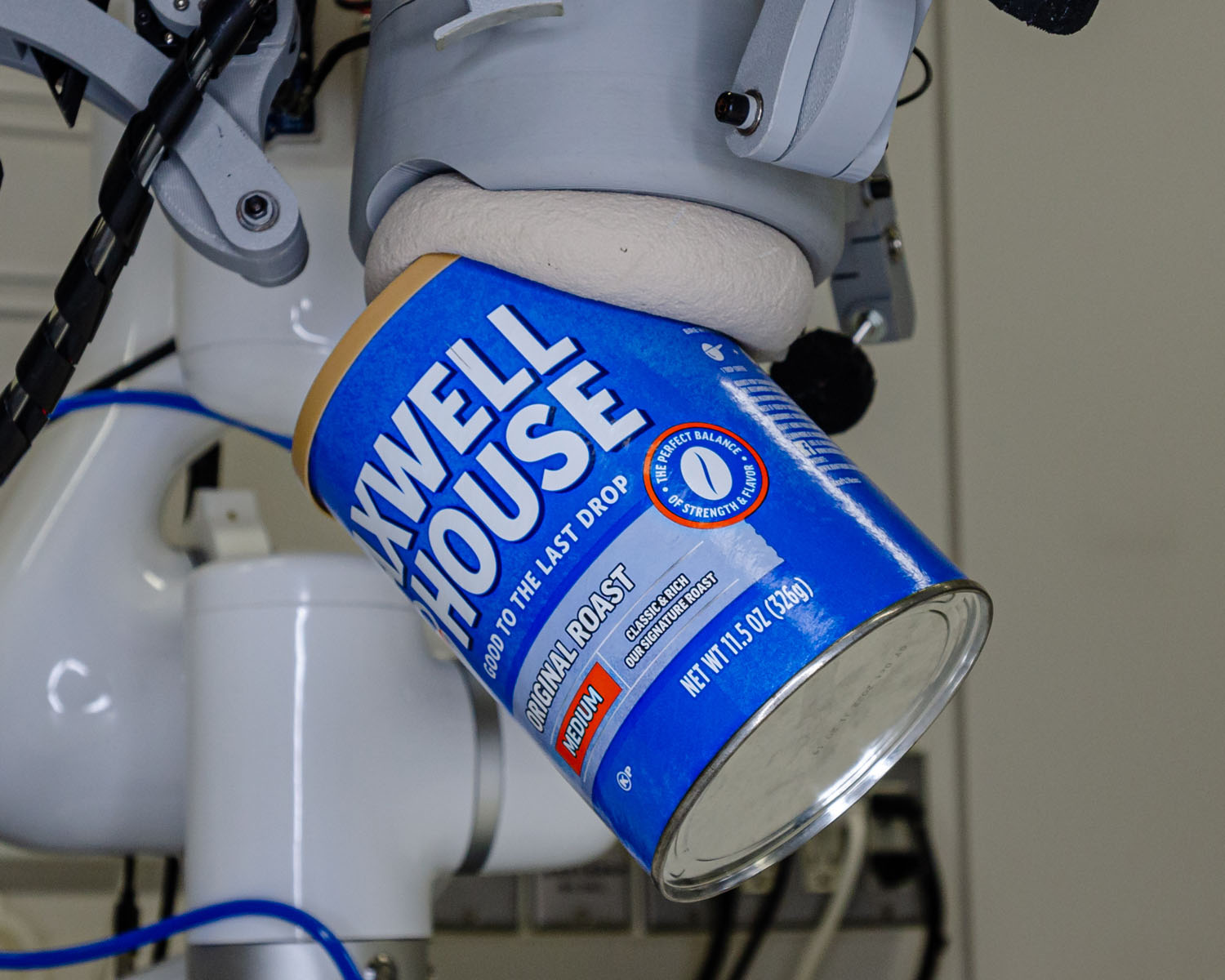}
        \includegraphics[width=0.19\linewidth, clip, trim=0pt 0pt 0pt 0pt]{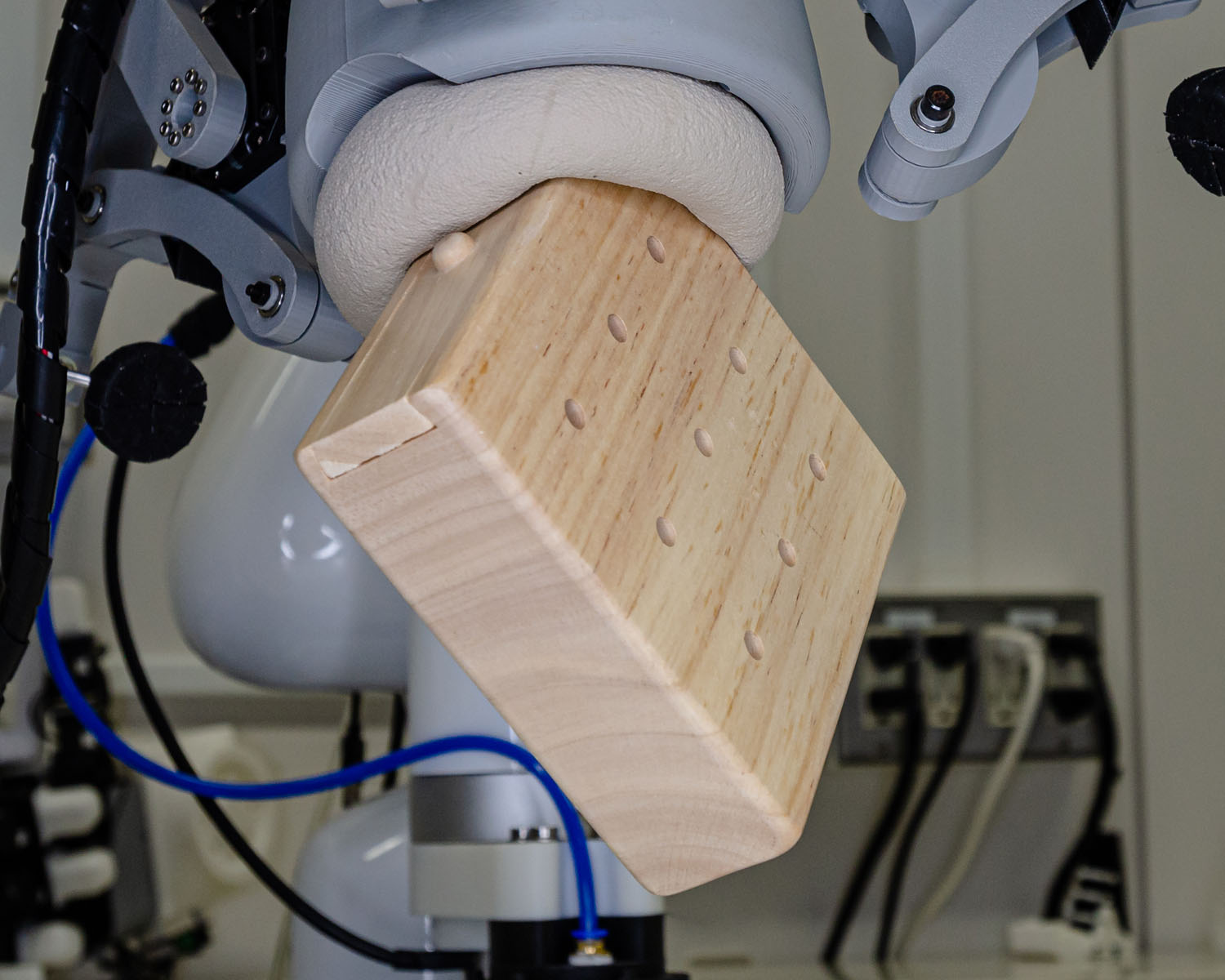}
        \includegraphics[width=0.19\linewidth, clip, trim=0pt 0pt 0pt 0pt]{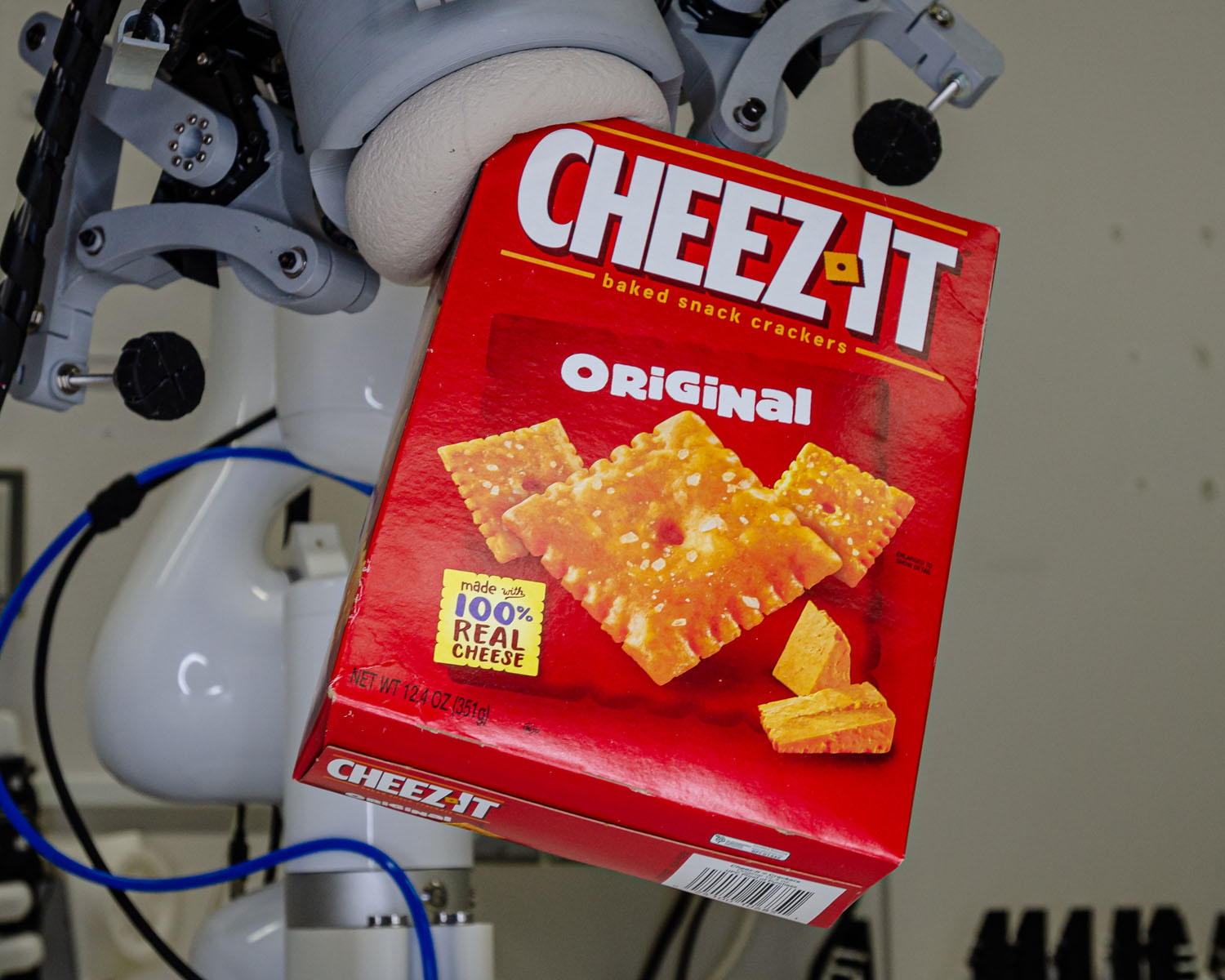}
        \\
        \vspace{1.2mm}
        \includegraphics[width=0.19\linewidth, clip, trim=0pt 0pt 0pt 0pt]{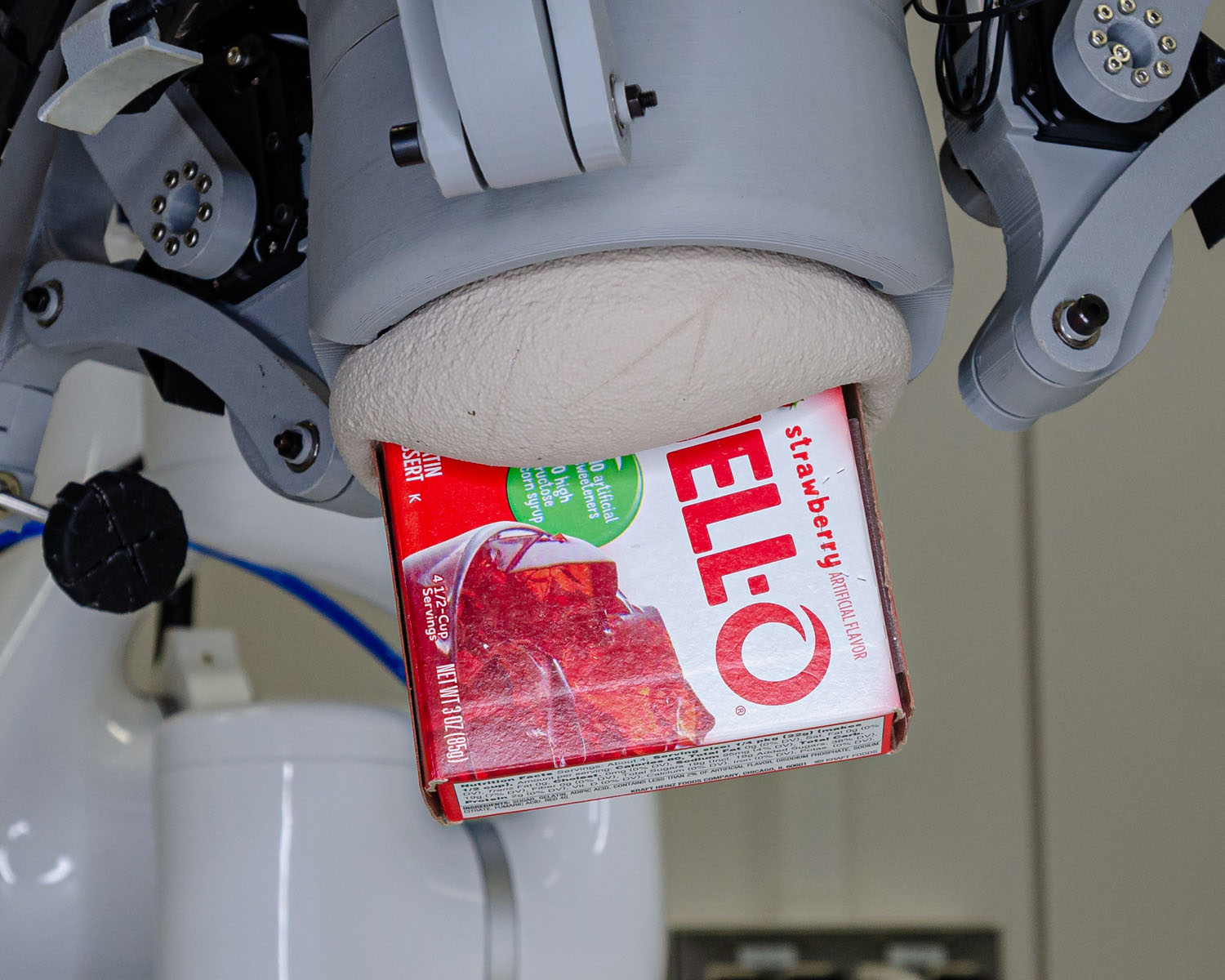}
        \includegraphics[width=0.19\linewidth, clip, trim=0pt 0pt 0pt 0pt]{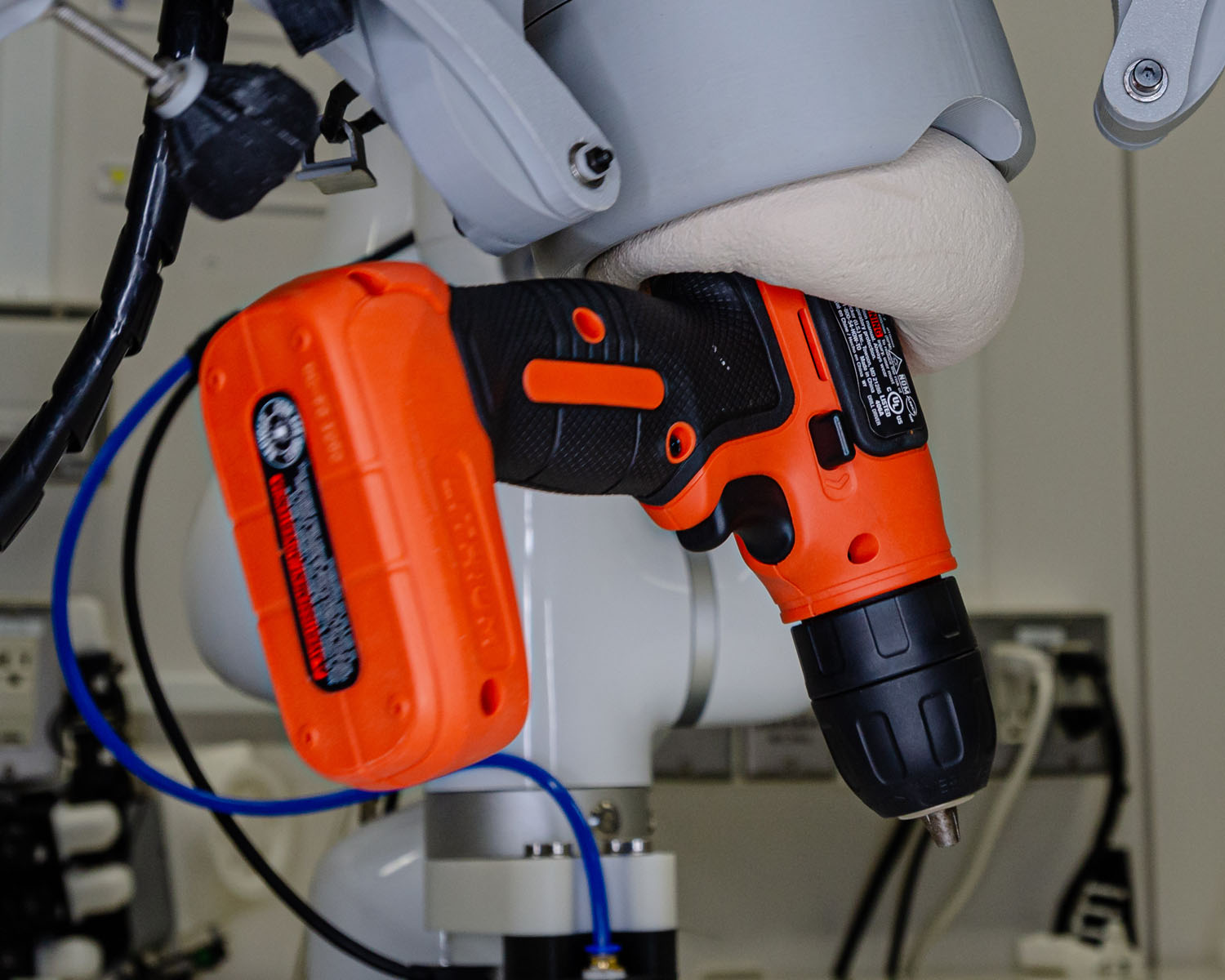}
        \includegraphics[width=0.19\linewidth, clip, trim=0pt 0pt 0pt 0pt]{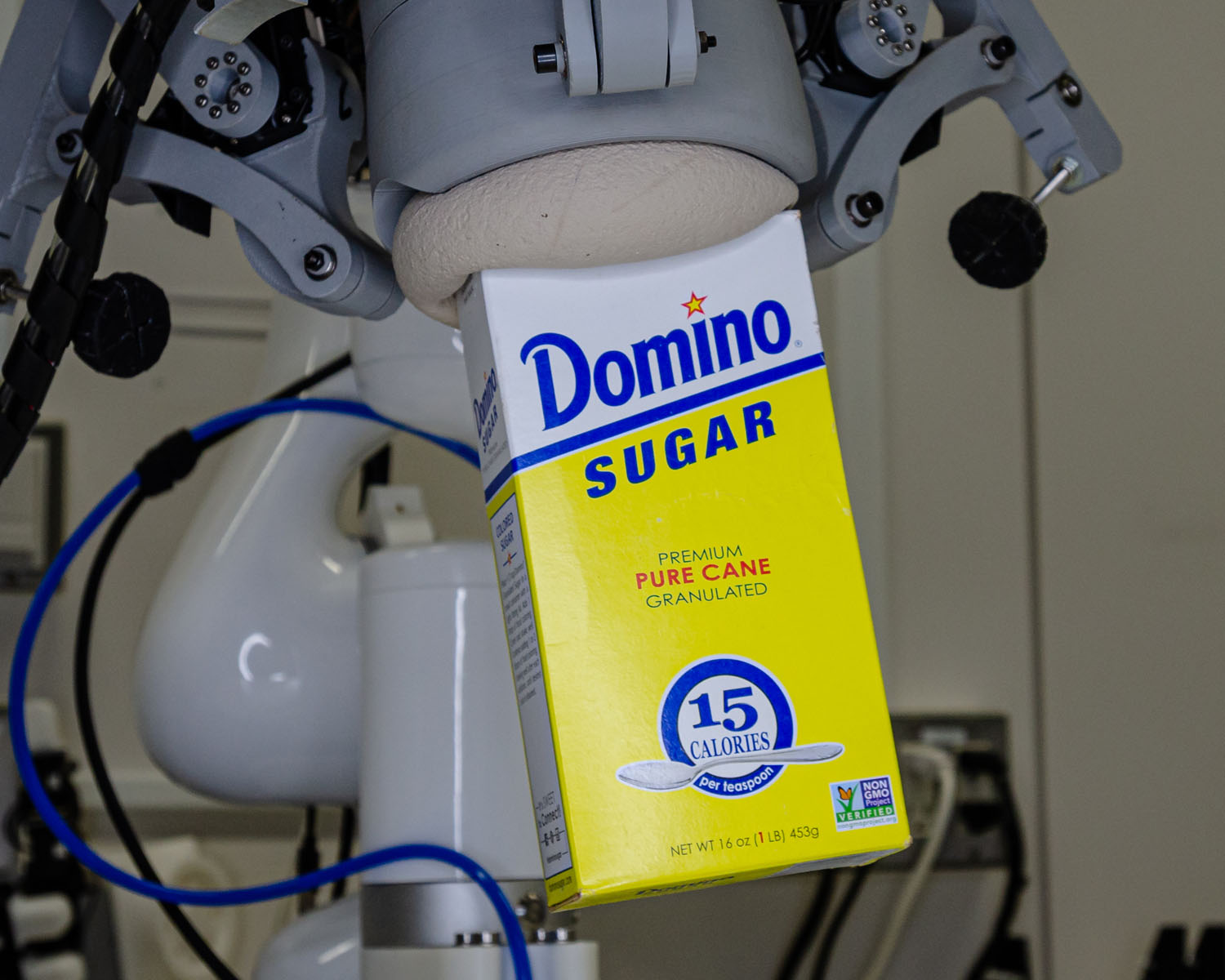}
        \includegraphics[width=0.19\linewidth, clip, trim=0pt 0pt 0pt 0pt]{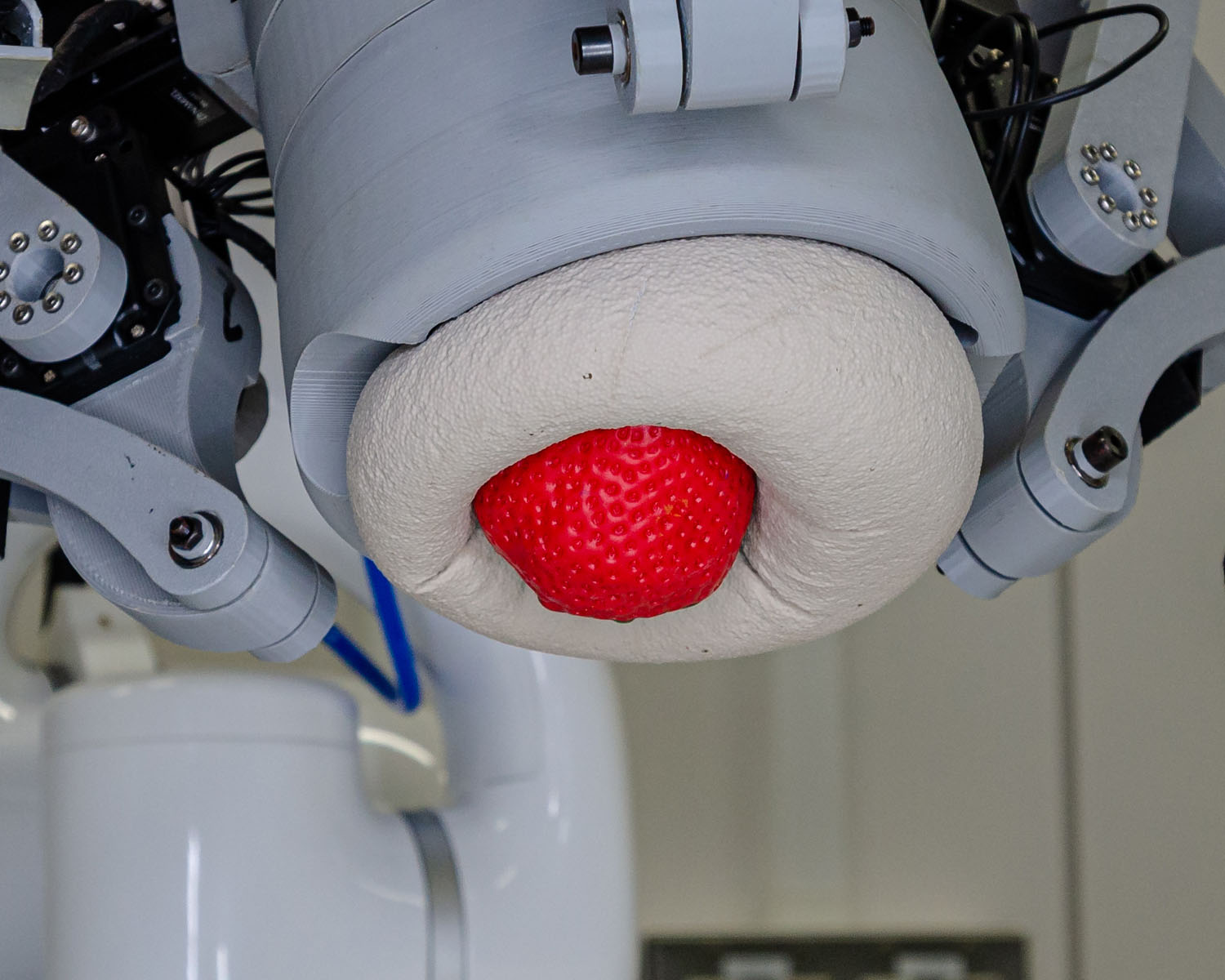}
        \includegraphics[width=0.19\linewidth, clip, trim=0pt 0pt 0pt 0pt]{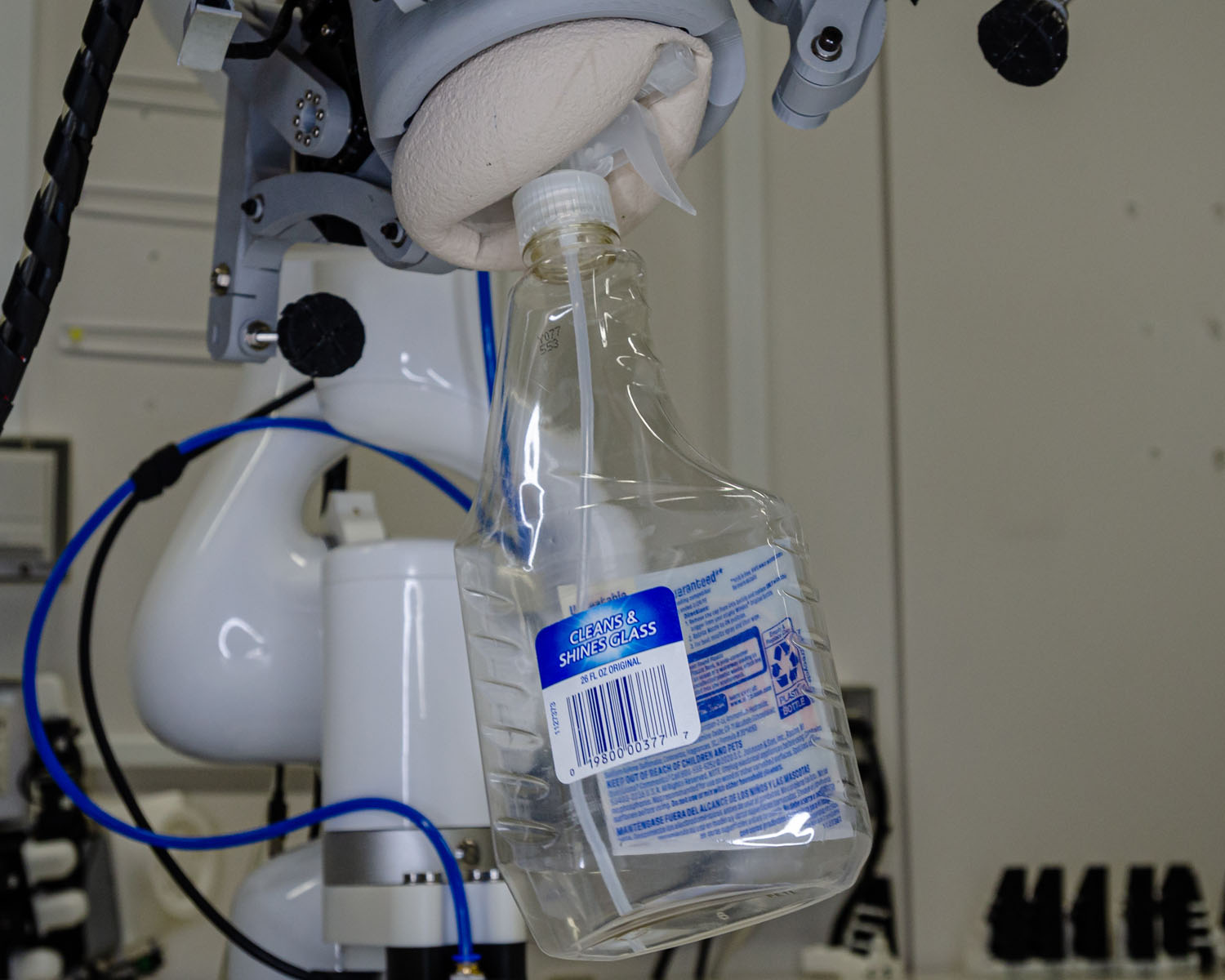}
        \\
        \vspace{1.2mm}
        \includegraphics[width=0.19\linewidth, clip, trim=0pt 0pt 0pt 0pt]{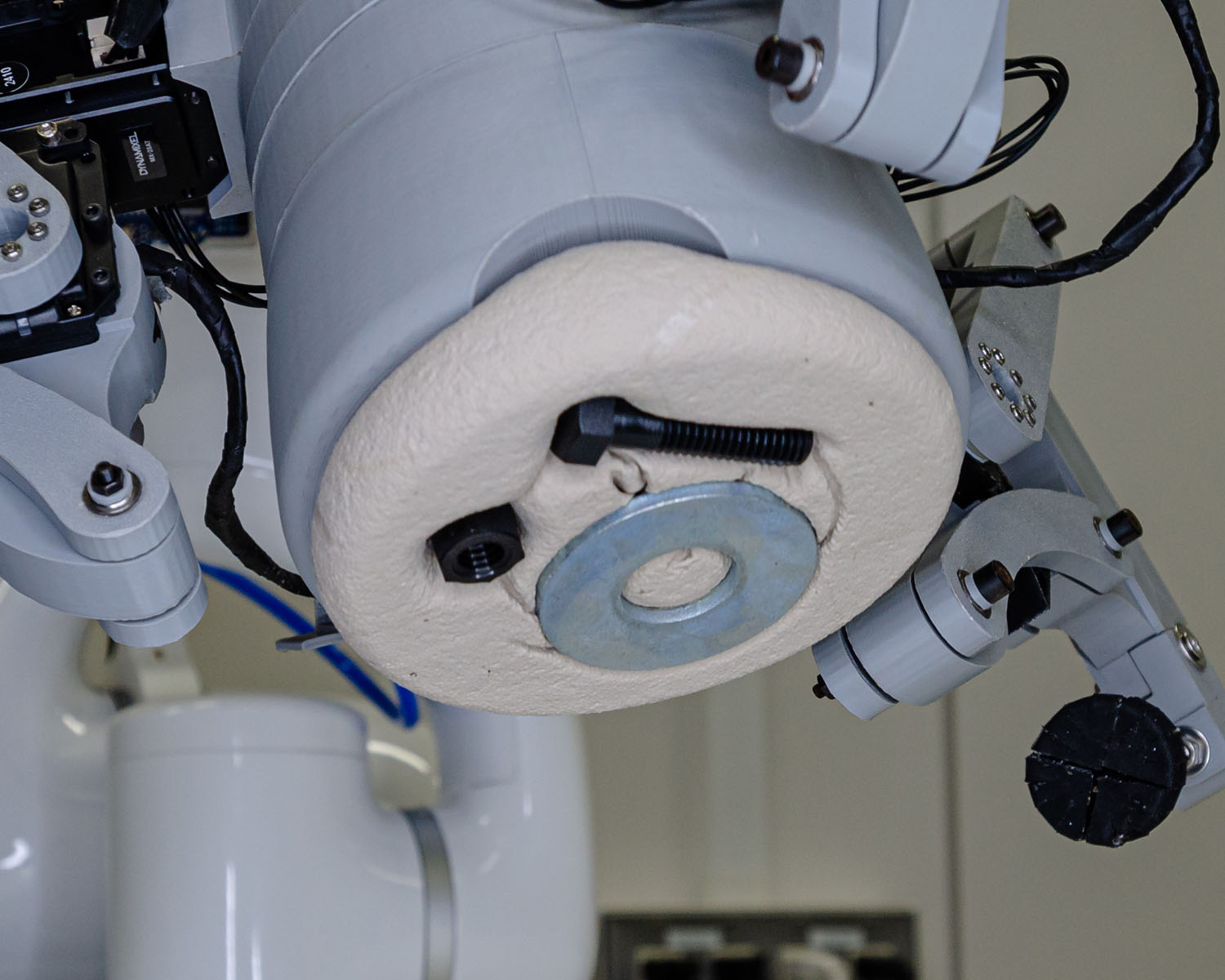}
        \includegraphics[width=0.19\linewidth, clip, trim=0pt 0pt 0pt 0pt]{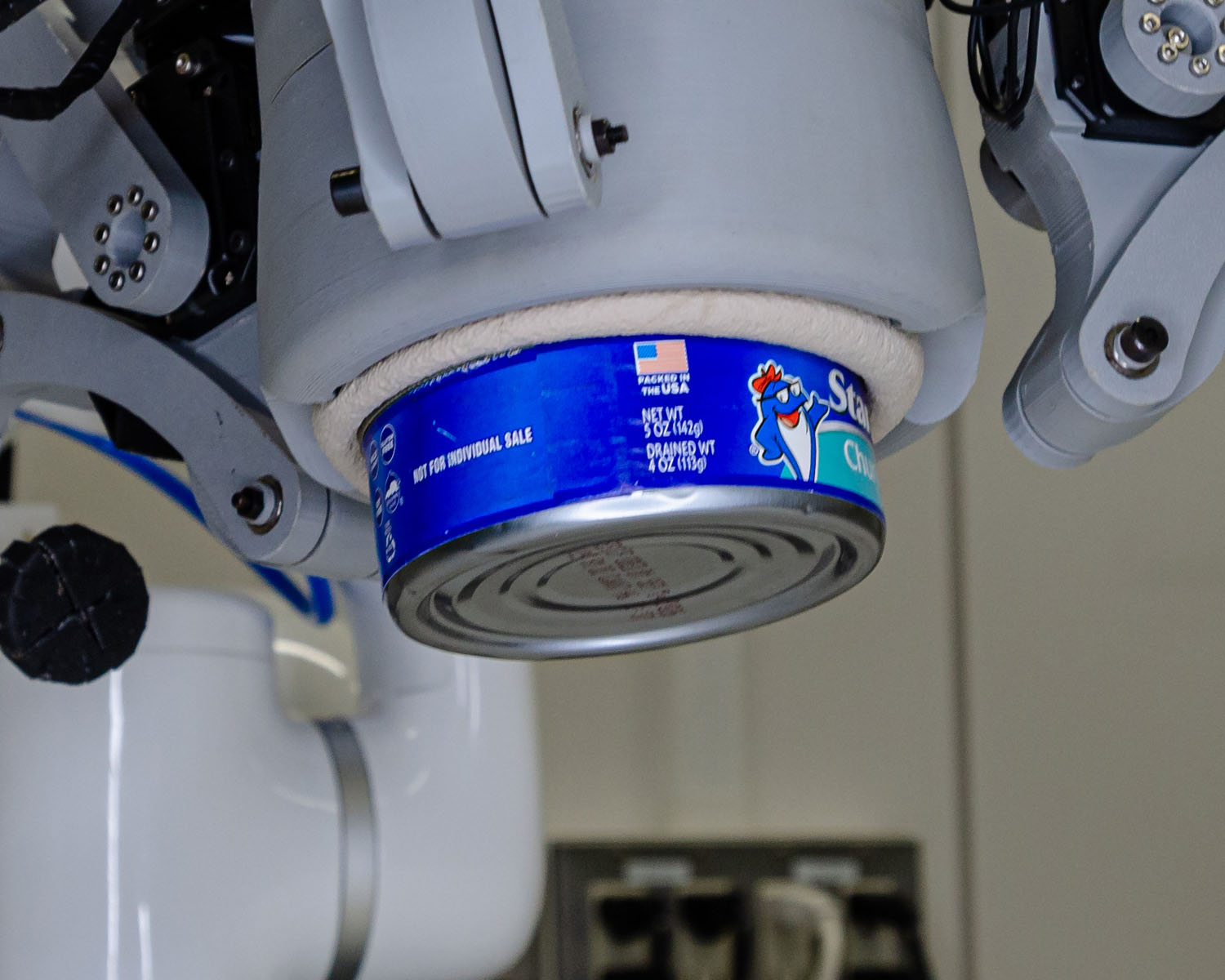}
        \includegraphics[width=0.19\linewidth, clip, trim=0pt 0pt 0pt 0pt]{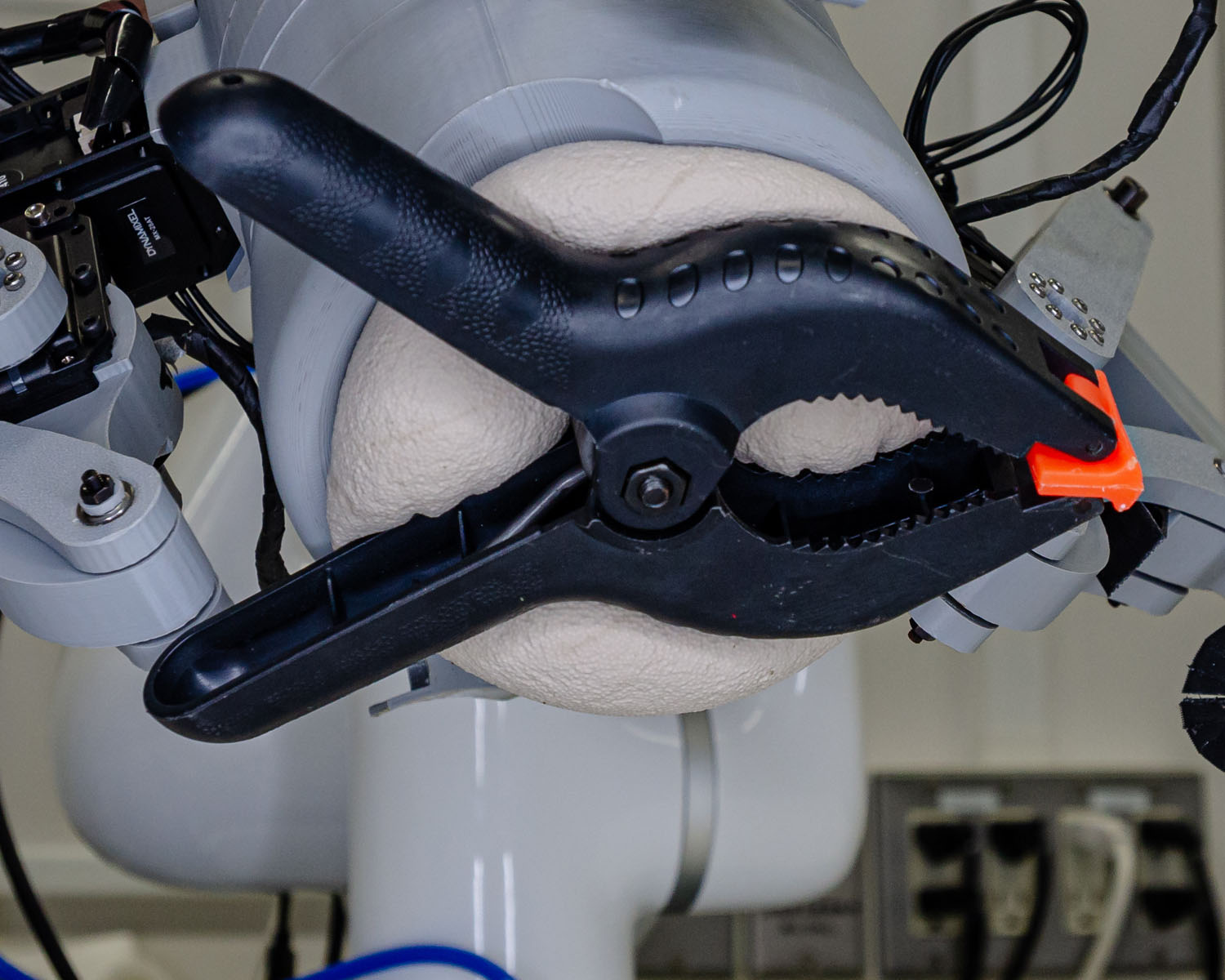}
        \includegraphics[width=0.19\linewidth, clip, trim=0pt 0pt 0pt 0pt]{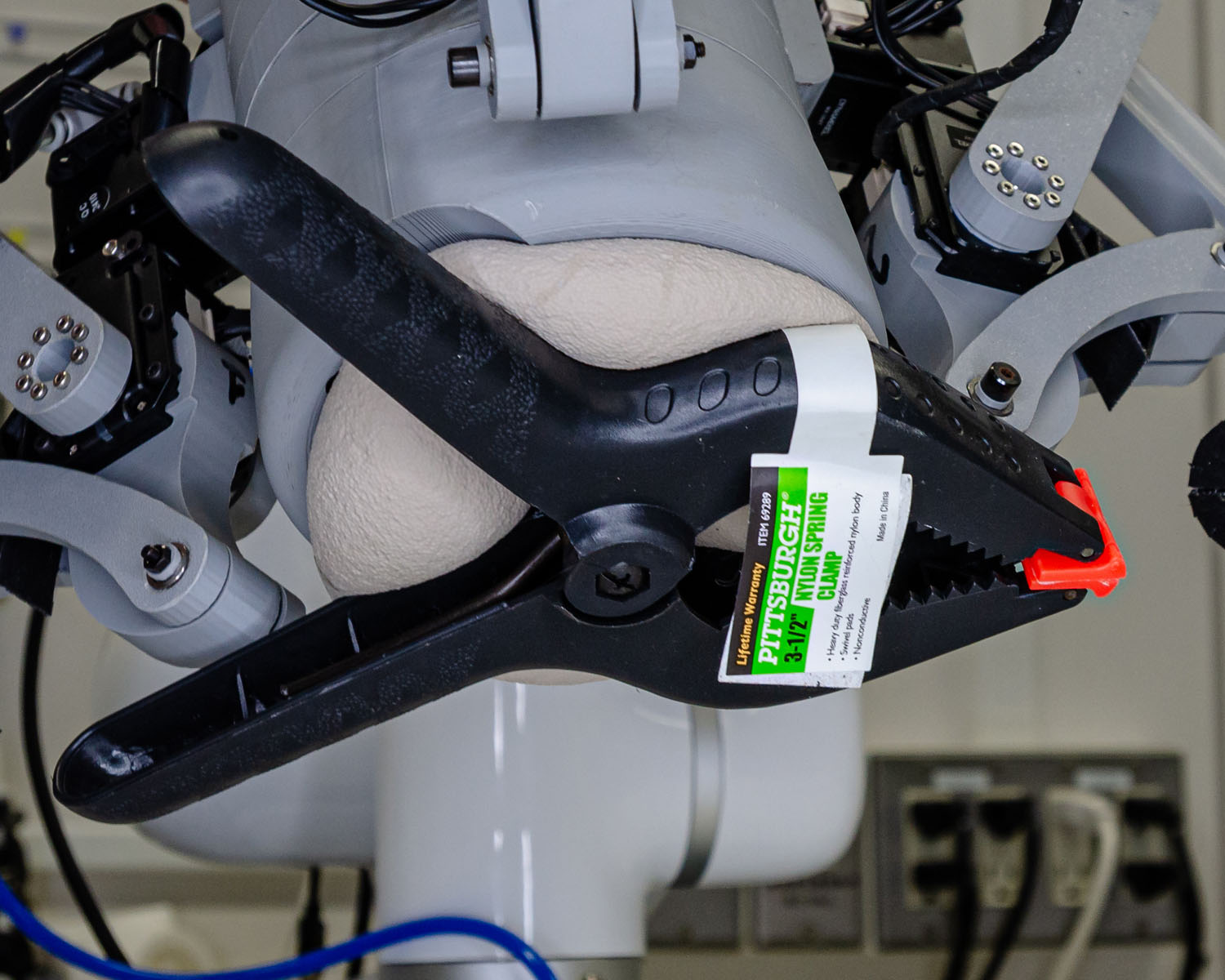}
        \includegraphics[width=0.19\linewidth, clip, trim=0pt 0pt 0pt 0pt]{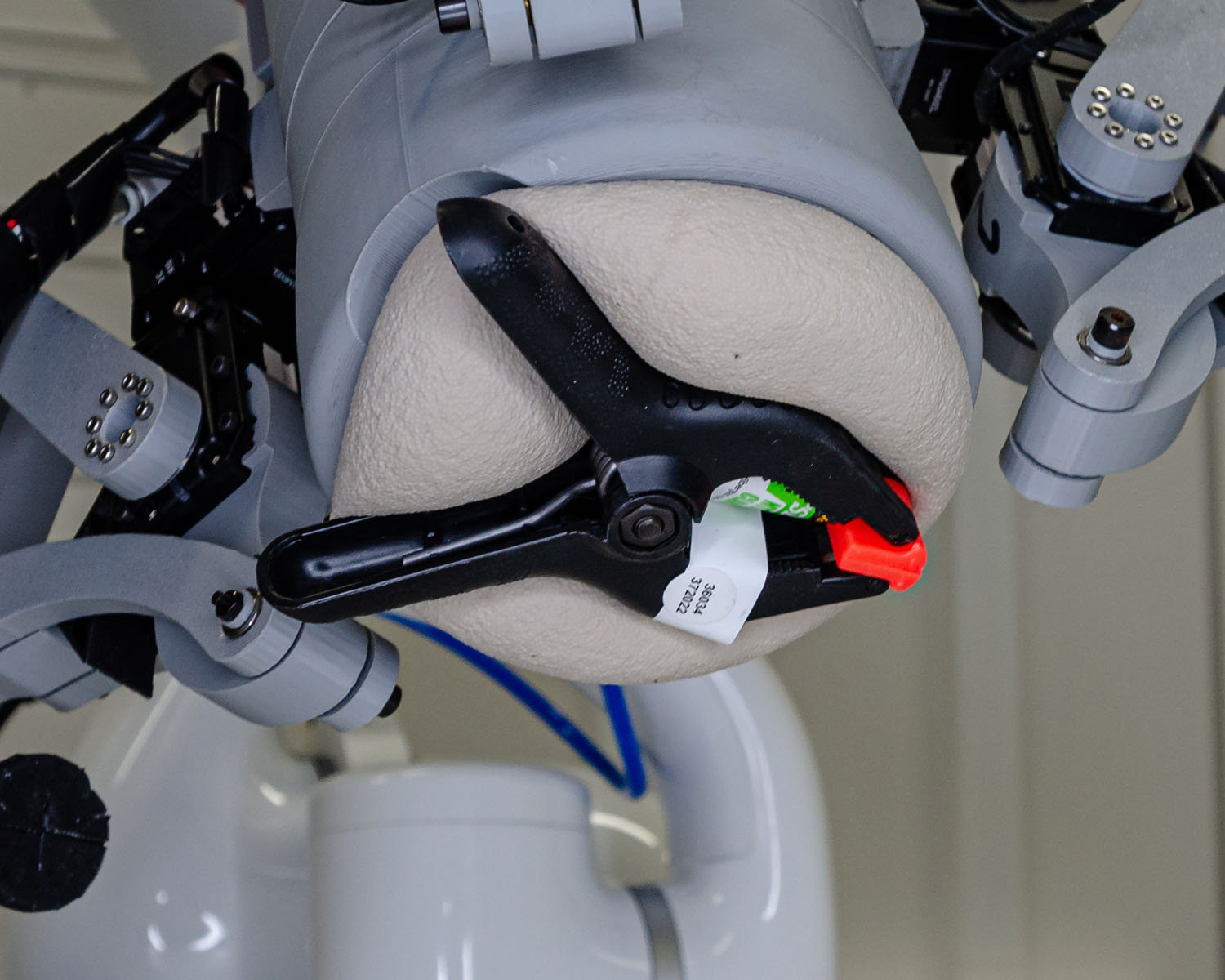}
        \\
        \vspace{1.2mm}
        \includegraphics[width=0.19\linewidth, clip, trim=0pt 0pt 0pt 0pt]{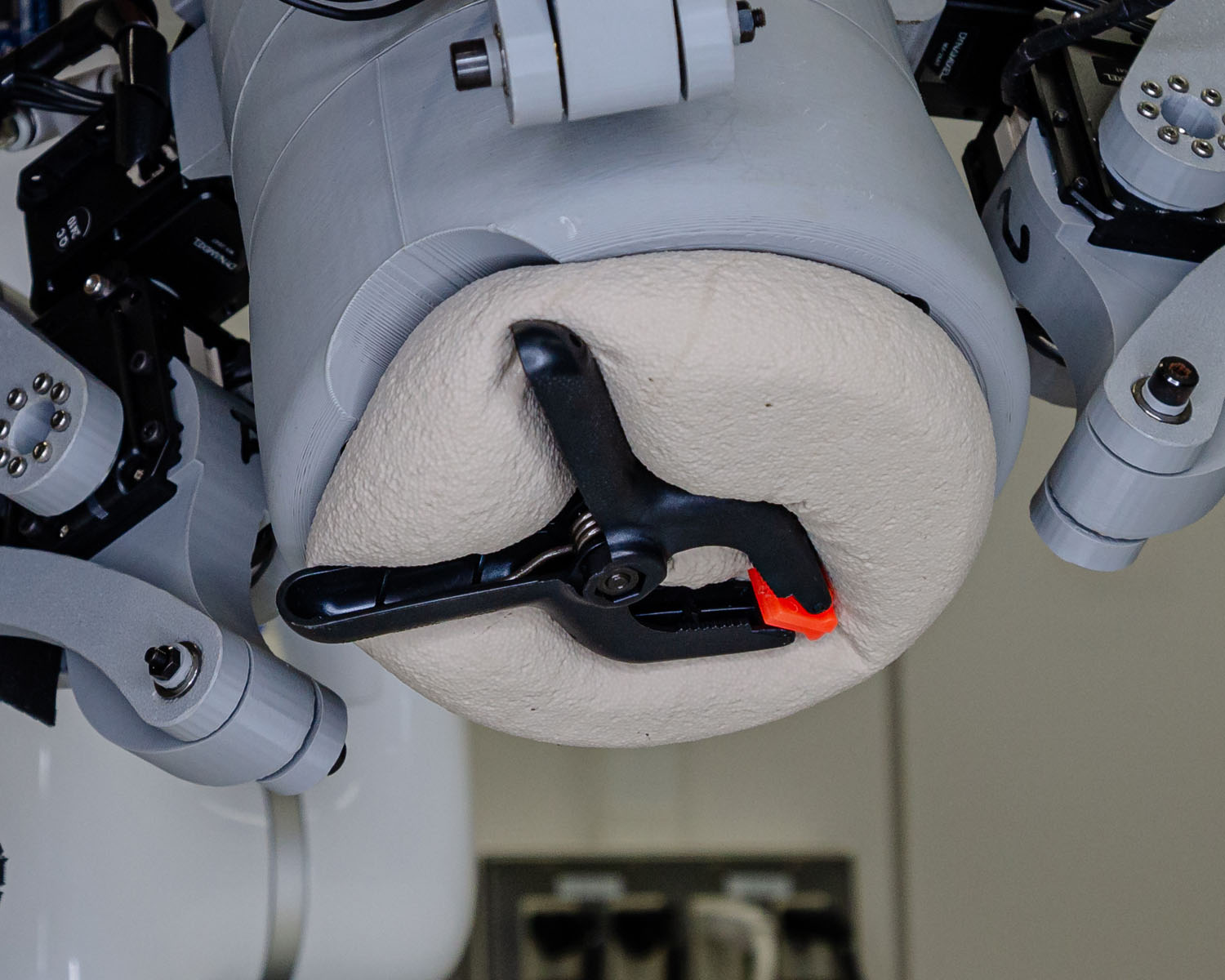}
        \includegraphics[width=0.19\linewidth, clip, trim=0pt 0pt 0pt 0pt]{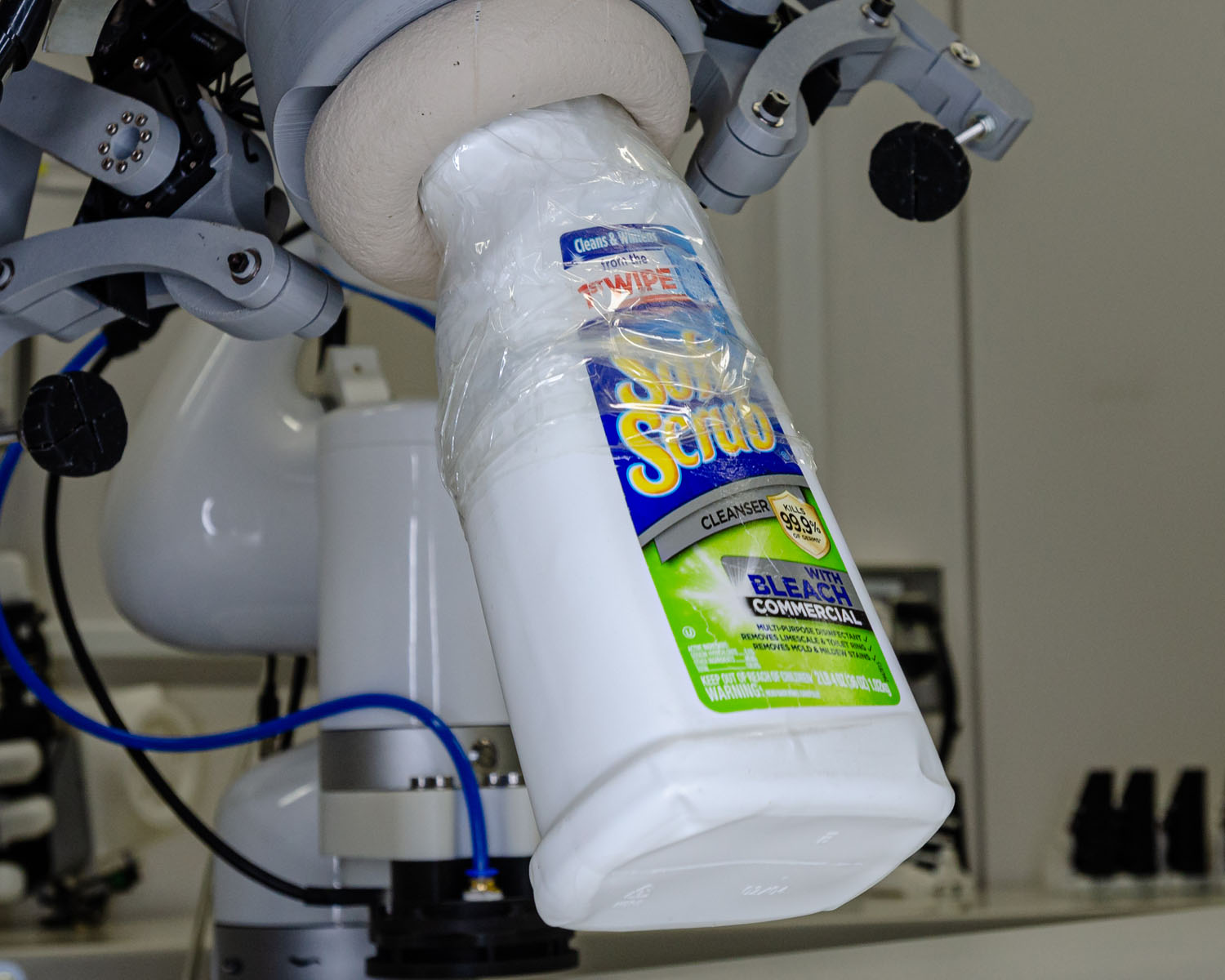}
        \includegraphics[width=0.19\linewidth, clip, trim=0pt 0pt 0pt 0pt]{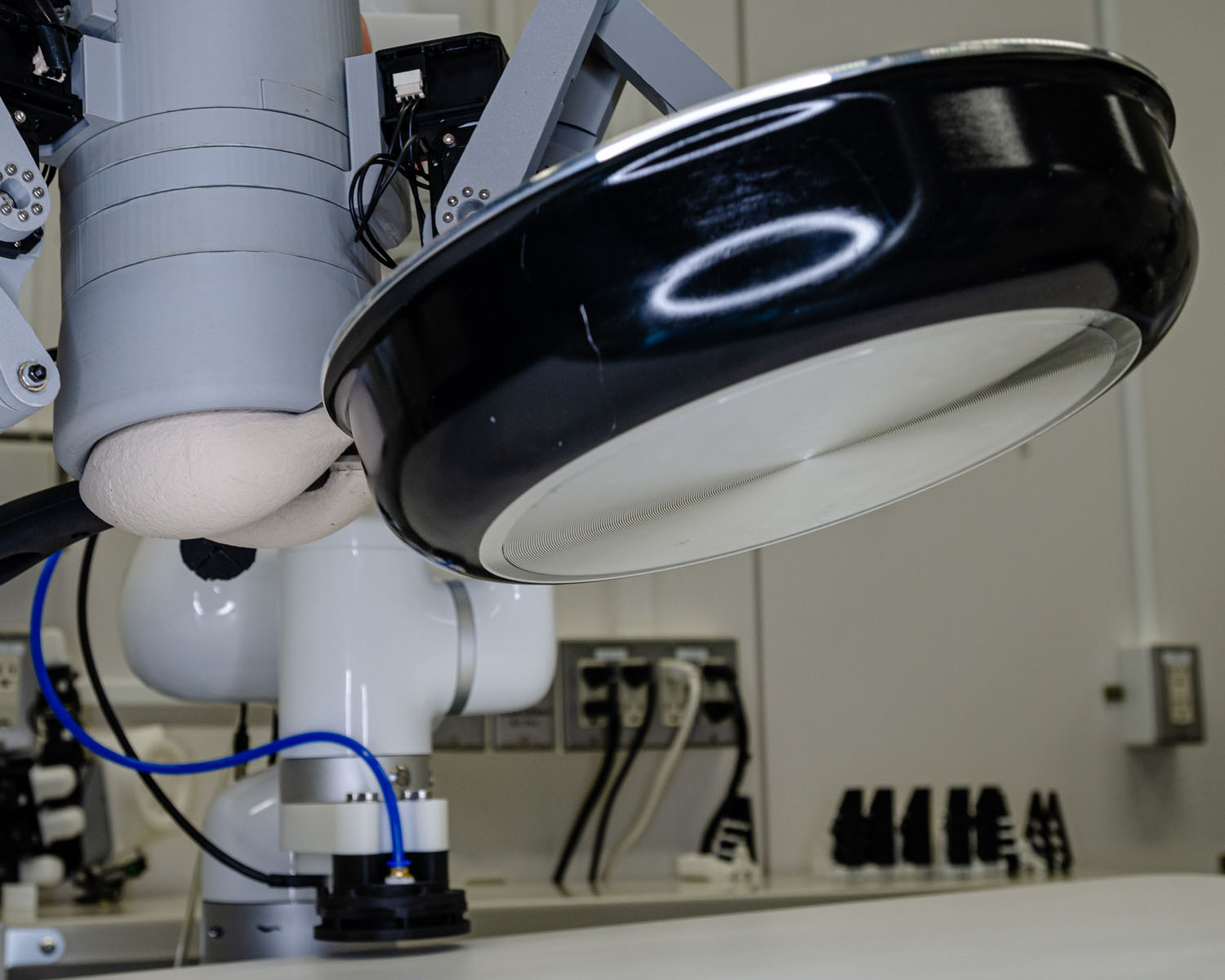}
        \includegraphics[width=0.19\linewidth, clip, trim=0pt 0pt 0pt 0pt]{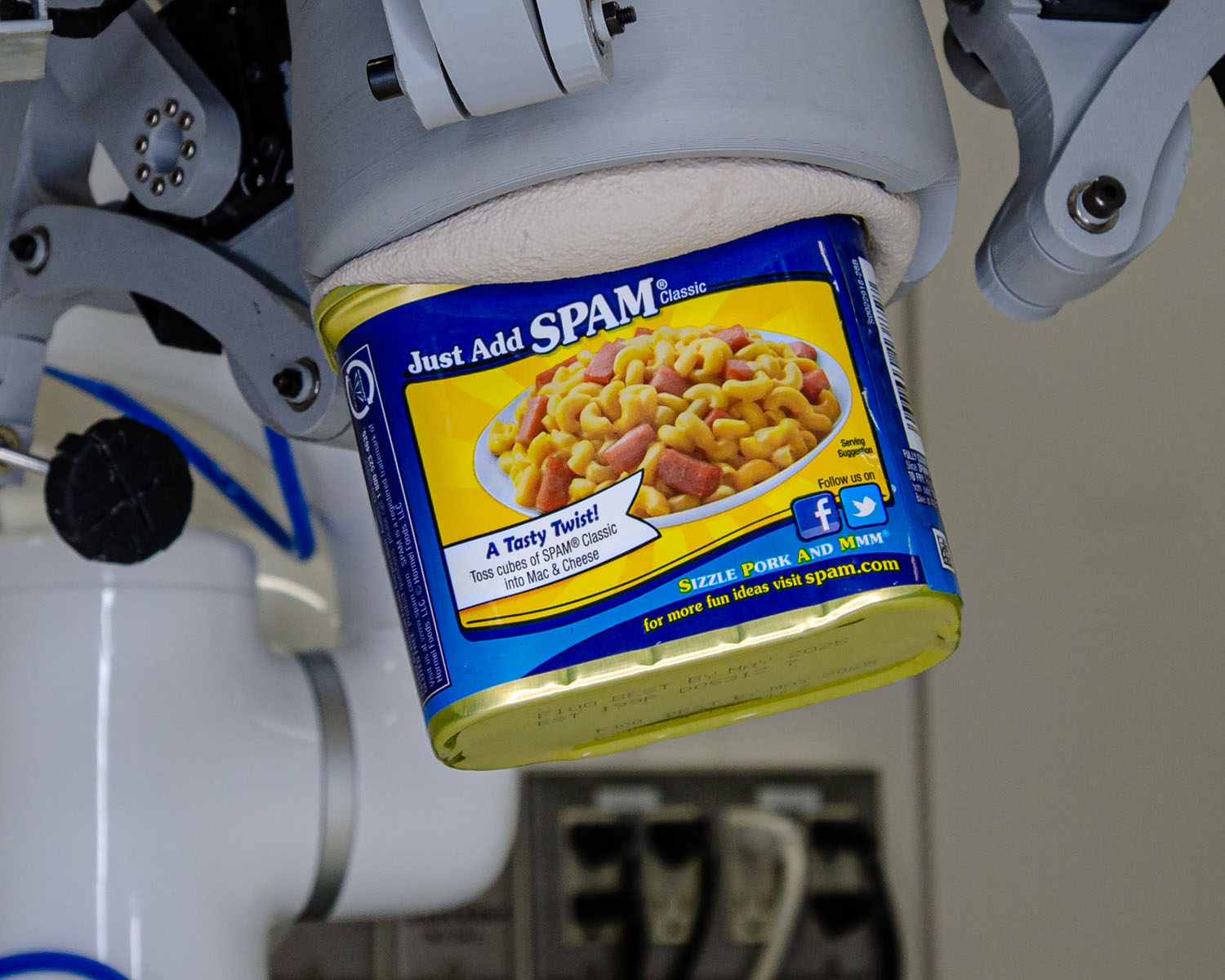}
        \includegraphics[width=0.19\linewidth, clip, trim=0pt 0pt 0pt 0pt]{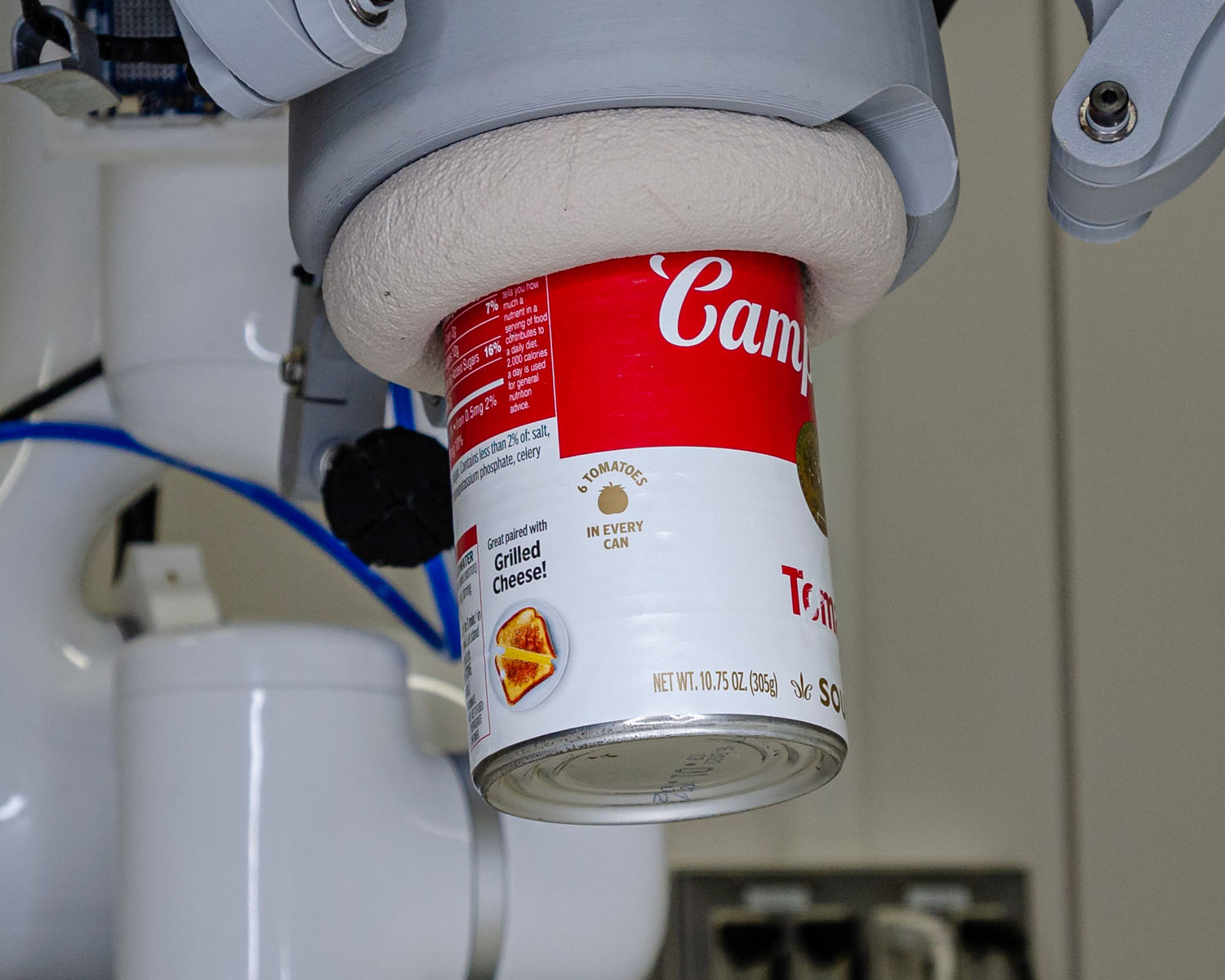}
        \\
        \vspace{1.2mm}
        \includegraphics[width=0.19\linewidth, clip, trim=0pt 0pt 0pt 0pt]{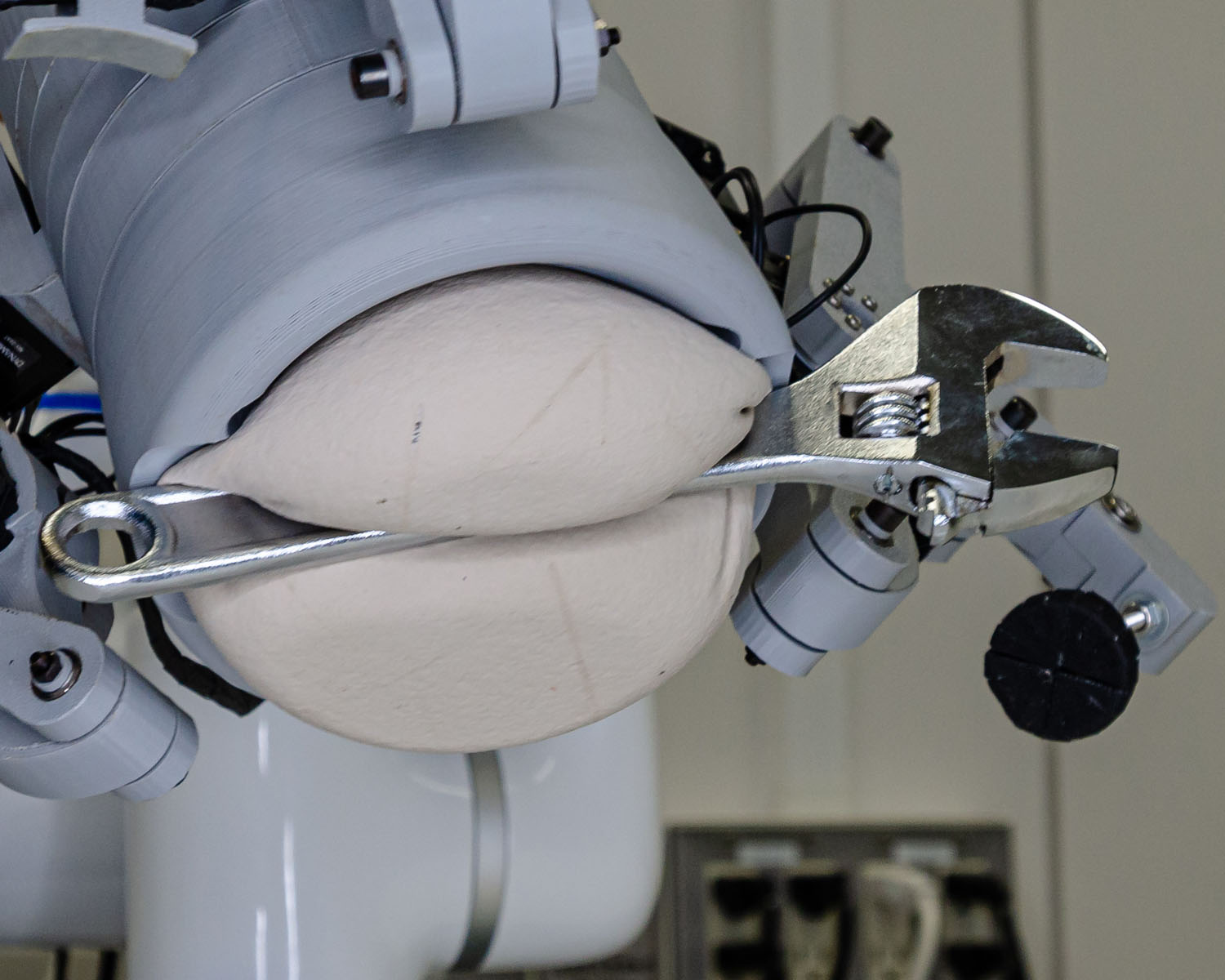}
        \includegraphics[width=0.19\linewidth, clip, trim=0pt 0pt 0pt 0pt]{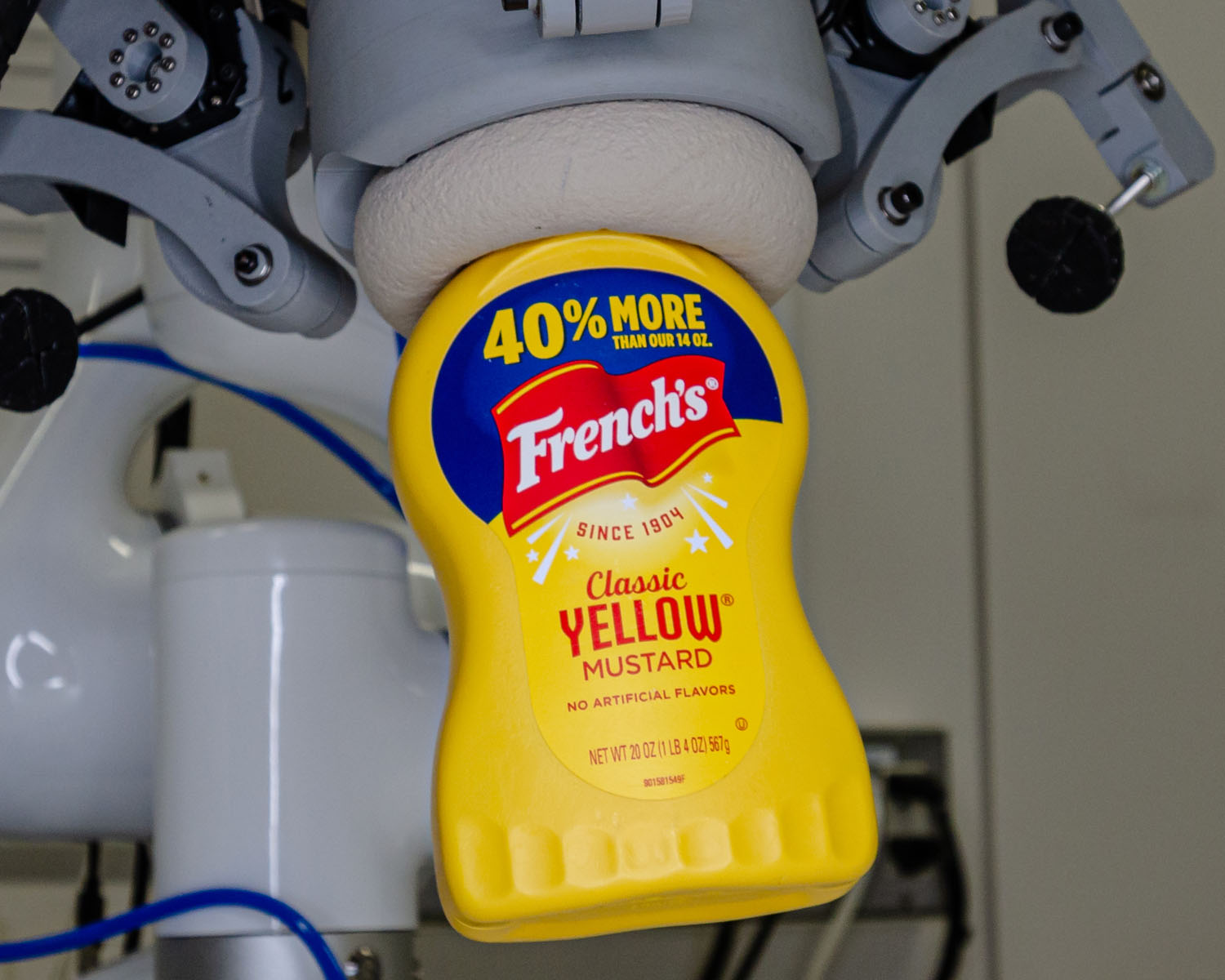}
        \includegraphics[width=0.19\linewidth, clip, trim=0pt 0pt 0pt 0pt]{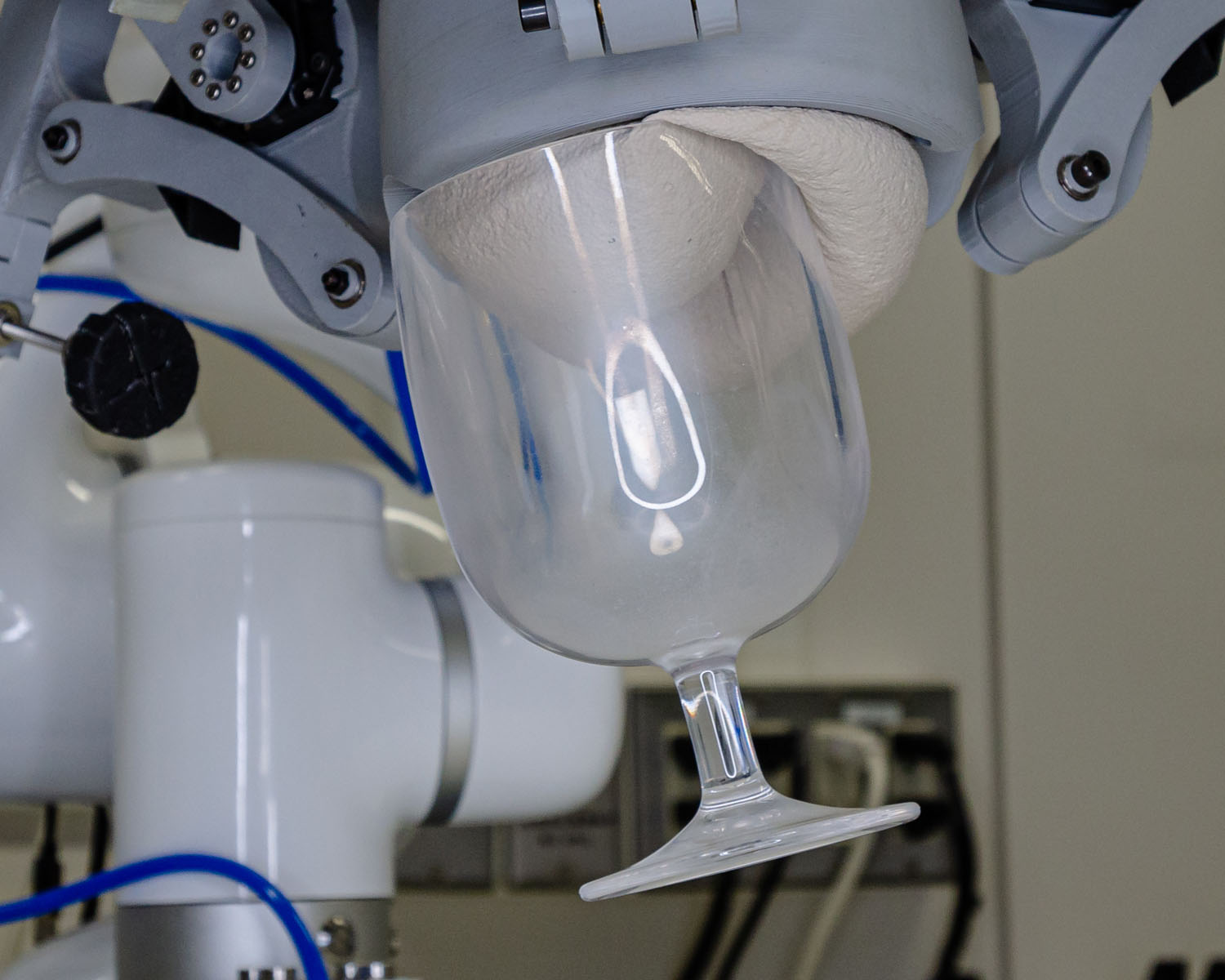}
        \includegraphics[width=0.19\linewidth, clip, trim=0pt 0pt 0pt 0pt]{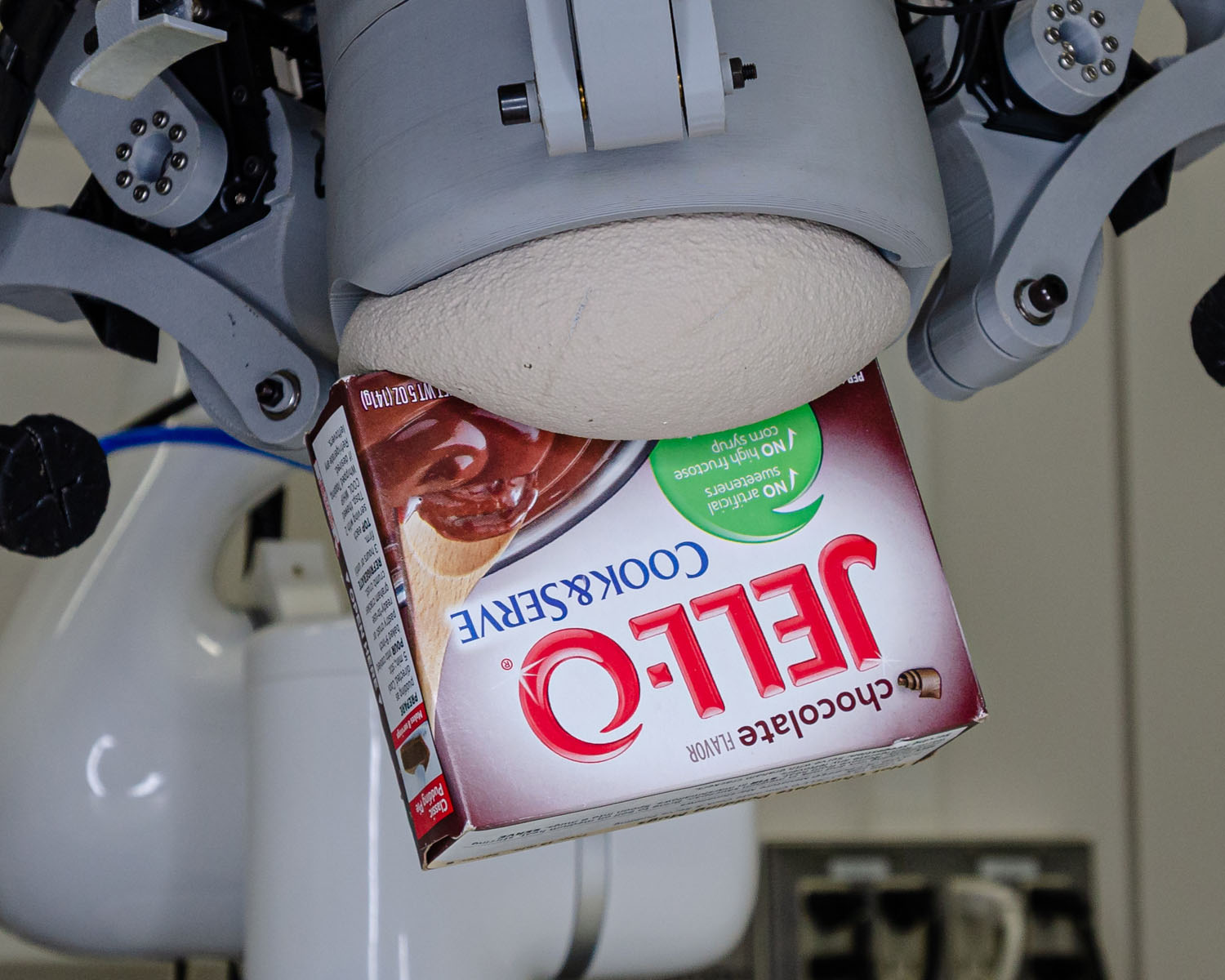}
        \includegraphics[width=0.19\linewidth, clip, trim=0pt 0pt 0pt 0pt]{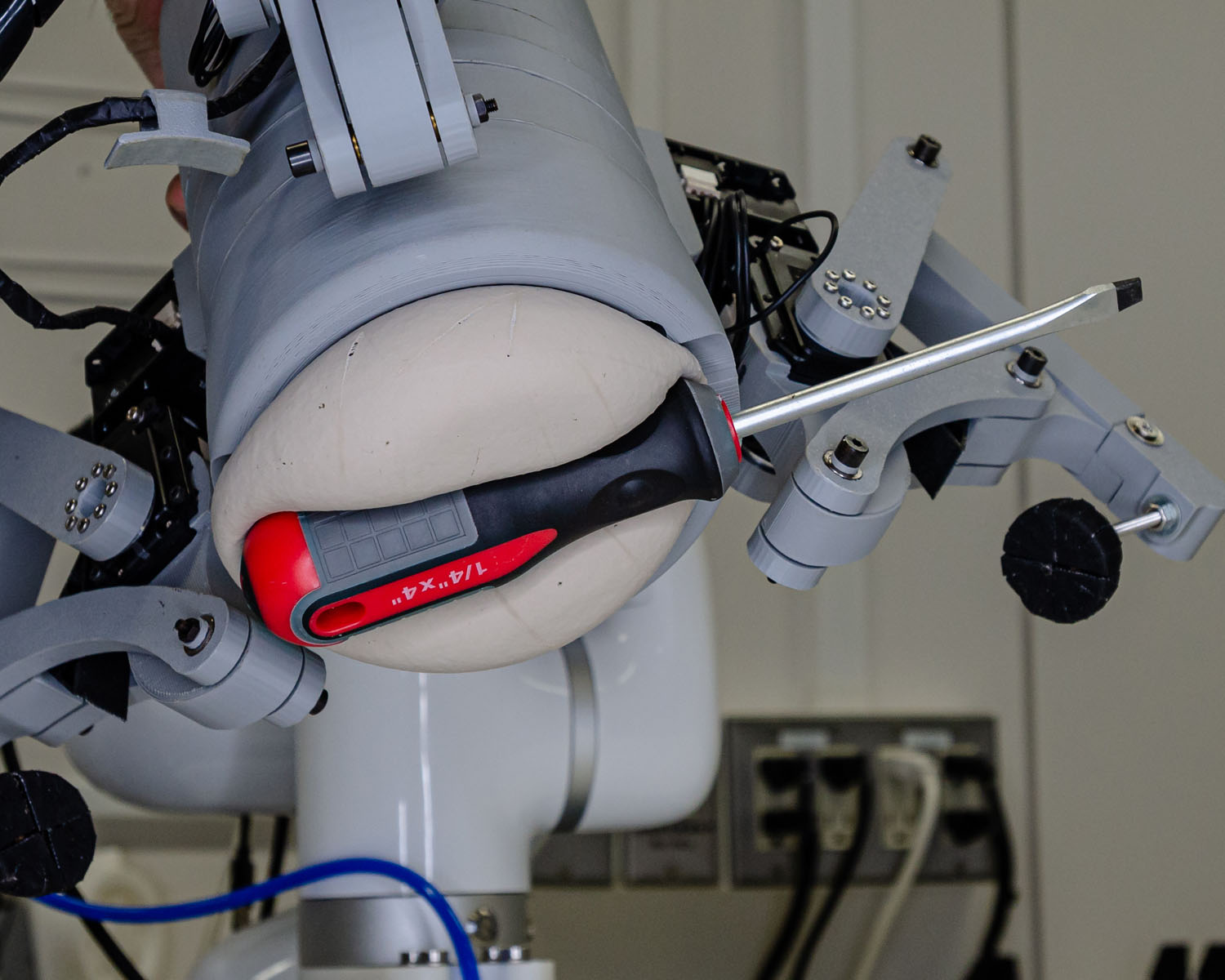}
        \\
 \caption{Loss of contact with screwdrivers under releasing: (a) holding \textnumero 3; (b) finger opening and falling \textnumero 3; (c) holding \textnumero 8; (d) finger opening and falling \textnumero 8.}
 \label{figYCB}
\end{figure}

\begin{table}[!t]
\centering
\vspace{2mm}
\caption{Grasping Results of ARTiS on YCB Object Subsets}
\label{tab:ycb_appendix_extended}
\scriptsize
\resizebox{\columnwidth}{!}{%
\begin{tabular}{l l c c c c}
\toprule
\textbf{YCB subset} &
\textbf{Object} &
\textbf{Successes/Attempts} &
\textbf{Failures} &
\textbf{Success rate} &
\textbf{95\% CI} \\
\midrule
\multirow{18}{*}{Food items}
& chips can & 30/30 & 0 & 100.0\% & 88.6-100.0\% \\
& coffee can & 26/30 & 4 & 86.7\% & 70.3-94.7\% \\
& cracker box & 23/30 & 7 & 76.7\% & 59.1-88.2\% \\
& box of sugar & 30/30 & 0 & 100.0\% & 88.6-100.0\% \\
& tomato soup can & 30/30 & 0 & 100.0\% & 88.6-100.0\% \\
& mustard container & 27/30 & 3 & 90.0\% & 74.4-96.5\% \\
& tuna fish can & 30/30 & 0 & 100.0\% & 88.6-100.0\% \\
& chocolate pudding box & 30/30 & 0 & 100.0\% & 88.6-100.0\% \\
& gelatin box & 30/30 & 0 & 100.0\% & 88.6-100.0\% \\
& potted meat can & 30/30 & 0 & 100.0\% & 88.6-100.0\% \\
& lemon & 30/30 & 0 & 100.0\% & 88.6-100.0\% \\
& apple & 30/30 & 0 & 100.0\% & 88.6-100.0\% \\
& pear & 30/30 & 0 & 100.0\% & 88.6-100.0\% \\
& orange & 30/30 & 0 & 100.0\% & 88.6-100.0\% \\
& banana & 30/30 & 0 & 100.0\% & 88.6-100.0\% \\
& peach & 30/30 & 0 & 100.0\% & 88.6-100.0\% \\
& strawberries & 30/30 & 0 & 100.0\% & 88.6-100.0\% \\
& plum & 30/30 & 0 & 100.0\% & 88.6-100.0\% \\
\midrule
\multirow{12}{*}{Kitchen items}
& pitcher & 25/30 & 5 & 83.3\% & 66.4-92.7\% \\
& bleach cleanser & 22/30 & 8 & 73.3\% & 55.5-85.8\% \\
& glass cleaner & 28/30 & 2 & 93.3\% & 78.7-98.1\% \\
& plastic wine glass & 30/30 & 0 & 100.0\% & 88.6-100.0\% \\
& enamel-coated metal bowl & 30/30 & 0 & 100.0\% & 88.6-100.0\% \\
& metal mug & 30/30 & 0 & 100.0\% & 88.6-100.0\% \\
& cooking skillet & 19/30 & 11 & 63.3\% & 45.5-78.1\% \\
& glass lid & 30/30 & 0 & 100.0\% & 88.6-100.0\% \\
& metal plate & 17/30 & 13 & 56.7\% & 39.2-72.6\% \\
& spoon & 30/30 & 0 & 100.0\% & 88.6-100.0\% \\
& fork & 30/30 & 0 & 100.0\% & 88.6-100.0\% \\
& spatula & 30/30 & 0 & 100.0\% & 88.6-100.0\% \\
\midrule
\multirow{13}{*}{Tool items}
& power drill & 30/30 & 0 & 100.0\% & 88.6-100.0\% \\
& wood block & 25/30 & 5 & 83.3\% & 66.4-92.7\% \\
& scissors & 27/30 & 3 & 90.0\% & 74.4-96.5\% \\
& keys & 30/30 & 0 & 100.0\% & 88.6-100.0\% \\
& marker & 30/30 & 0 & 100.0\% & 88.6-100.0\% \\
& adjustable wrench & 30/30 & 0 & 100.0\% & 88.6-100.0\% \\
& Phillips-head screwdriver & 30/30 & 0 & 100.0\% & 88.6-100.0\% \\
& flat-head screwdriver & 30/30 & 0 & 100.0\% & 88.6-100.0\% \\
& wood screws & 30/30 & 0 & 100.0\% & 88.6-100.0\% \\
& plastic bolts & 30/30 & 0 & 100.0\% & 88.6-100.0\% \\
& nuts & 30/30 & 0 & 100.0\% & 88.6-100.0\% \\
& hammer & 30/30 & 0 & 100.0\% & 88.6-100.0\% \\
& spring clamps & 30/30 & 0 & 100.0\% & 88.6-100.0\% \\
\midrule
\textbf{Food it. total} & - & \textbf{526/540} & \textbf{14} & \textbf{97.4\%} & \textbf{95.7-98.4\%} \\
\textbf{Kitchen it. total} & - & \textbf{321/360} & \textbf{39} & \textbf{89.2\%} & \textbf{85.5-91.9\%} \\
\textbf{Tool it. total} & - & \textbf{382/390} & \textbf{8} & \textbf{97.9\%} & \textbf{96.0-98.9\%} \\
\midrule
\textbf{Overall} & \textbf{43 objects} & \textbf{1229/1290} & \textbf{61} & \textbf{95.3\%} & \textbf{93.9-96.3\%} \\
\bottomrule
\end{tabular}}
\begin{flushleft}
\scriptsize
The confidence intervals were calculated using the Wilson score interval. Sharp or puncturing objects (knives and needles) from the selected YCB subsets were excluded to prevent damage to the jamming palm.
\end{flushleft}
\end{table}


\section{Generalization Evaluation Using YCB Objects}\label{A3}

\noindent To further evaluate the generalization capability of ARTiS beyond the studied disassembly tools (Fig.~\ref{fig_A2}), an additional grasping experiment was conducted using objects from the YCB benchmark set (Fig.~\ref{figYCB}). The grasping of objects from the YCB benchmark set was carried out from the table, and since the jamming palm is quite large, almost all objects were captured at their central geometric point. There are objects that are much larger than the palm of ARTiS, so the choice of the grasping pose was made according to a different logic. These objects can be divided into two categories: objects with protruding elements or handles, and boxes or cans. When talking about objects with protruding elements or handles, include a pan, pitcher, glass cleaner, etc. The results show that such objects usually do not have a one hundred percent success rate because it is very difficult to determine the ideal position and orientation of their interaction with the handle. When discussing objects such as boxes or cans, including a chip can, coffee can, wood block, etc. The logic for grasping these objects, the grasping logic, was to grip the object by the edge, as there is a possibility of creating an encompassing effect. However, if you select the middle of the object, the palm will not be able to wrap around the object due to its geometry.

Table~\ref{tab:ycb_appendix_extended} provides the detailed per-object results for the additional YCB generalization experiment (results for each tested object, the number of successful and failed trials, the corresponding success rate, and the 95\% confidence interval). The confidence intervals were calculated using the Wilson score interval, which is suitable for binomial success/failure data and provides a more stable estimate than the normal approximation, especially for objects with success rates close to 0\% or 100\%. 

The extended results show that ARTiS achieved robust grasping performance over a broad range of object geometries and physical properties. Out of 43 tested YCB objects, 33 objects were grasped with 100\% success over 30 trials, and 36 objects achieved a success rate of at least 90\%. The highest subset-level success rates were obtained for tool items and food items, with 97.95\% and 97.41\%, respectively. The kitchen subset showed a lower success rate of 89.17\%, mainly due to several challenging objects with thin, flat, asymmetric, or difficult-to-support geometries, such as the metal plate and cooking skillet. 

These detailed results support the generalization analysis presented in the main text by showing that the high overall success rate was not produced by a small number of easy objects only. Instead, ARTiS maintained reliable grasping performance across many object types, including boxes, cans, fruit-like objects, cups, bowls, utensils, hand tools, and small tool-related components. The per-object statistics in Table~\ref{tab:ycb_appendix_extended} also identify the remaining failure cases, which are mainly associated with objects that provide a limited contact area or create an unfavorable load distribution on the jamming palm.


\appendix[Torque-Based Performance Evaluation]\label{An}

This appendix provides a detailed analysis of the interaction of a screwdriver with an active jamming palm during fastening. Nine commercial screwdrivers (Fig.~\ref{subfig10_1}) were evaluated using a torque–time processing pipeline. Each data set consisted of 30 repeated tests, with torque measured in a $10\,$s window at fixed palm rotation speed (3 deg/s). For each torque data set $\tau(t)$, we compute performance metrics that capture strength, stability, controllability, and safety factors (Table~\ref{tab_A2}). Before that, we smoothed the torque data using a sliding window:
\[ \tilde{\tau}(t)=\frac{1}{w}\sum_{i=0}^{w-1} \tau(t-i), 
\qquad w=5. \]

\begin{table*}[!t]
\centering
\caption{Torque-based screwdrivers performance metrics - mean $\pm$ (standard deviation)}
\label{tab_A2}
\footnotesize
\begin{tabular}{c|cc|cc|cc|cc|cc|cc}
\toprule
Tool &
\multicolumn{2}{c|}{$\tau_{\max}$ (N·m)} &
\multicolumn{2}{c|}{$T_{\mathrm{slip}}$ (s)} &
\multicolumn{2}{c|}{$V$ (N·m)} &
\multicolumn{2}{c|}{$E_\theta$} &
\multicolumn{2}{c|}{$S$} &
\multicolumn{2}{c}{$R$}
\\
\midrule
1 & 2.297 & (0.100) & 9.625 & (0.390) & 0.01397 & (0.00343) & 0.03495 & (0.00436) & 6.119 & (0.728) & 1.0061 & (0.00153)\\
2 & 1.536 & (0.113) & 9.391 & (0.687) & 0.01301 & (0.00368) & 0.02970 & (0.00398) & 5.760 & (0.919) & 1.0085 & (0.00241)\\
3 & 1.149 & (0.080) & 5.098 & (1.113) & 0.01092 & (0.00275) & 0.01512 & (0.00393) & 5.178 & (0.224) & 1.0097 & (0.00282)\\
4 & 2.644 & (0.094) & 7.515 & (1.335) & 0.01675 & (0.00569) & 0.02630 & (0.00763) & 10.202 & (5.725) & 1.0063 & (0.00202)\\
5 & 1.348 & (0.071) & 6.988 & (0.825) & 0.01289 & (0.00434) & 0.02158 & (0.00303) & 5.567 & (0.877) & 1.0097 & (0.00359)\\
6 & 0.902 & (0.061) & 8.256 & (1.069) & 0.00908 & (0.00247) & 0.02289 & (0.00307) & 7.505 & (6.991) & 1.0101 & (0.00249)\\
7 & 0.404 & (0.034) & 9.535 & (0.618) & 0.00911 & (0.00258) & 0.02842 & (0.00237) & 7.144 & (3.435) & 1.0229 & (0.00581)\\
8 & 0.413 & (0.087) & 9.064 & (1.577) & 0.00914 & (0.00209) & 0.02635 & (0.00441) & 4.869 & (0.359) & 1.0247 & (0.01124)\\
9 & 4.798 & (0.126) & 9.917 & (0.116) & 0.02218 & (0.00663) & 0.04157 & (0.00817) & 8.282 & (2.853) & 1.0046 & (0.00135)\\
\bottomrule
\end{tabular}
\end{table*}

All further analysis is based on $\tilde{\tau}(t)$. To obtain the performance matrix, we calculate:

\paragraph{Maximum torque}
The maximum torque applied before the screwdriver slips:
\[ \tau_{\max} = \max_t \tilde{\tau}(t). \]
Screwdriver 9 achieves the highest maximum torque ($4.80\,$N·m), indicating the strongest mechanical coupling. Screwdrivers 7 and 8 produce the lowest torque ($\approx 0.41\,$N·m), which is due to weak geometric and frictional interaction.

\paragraph{Slip time}
Time required to reach maximum torque ($\tau_{\max}$):
\[ T_{\mathrm{slip}} = t(\tau_{\max}) - t_0. \]
The slip time correlates with the shorter time, which corresponds to rapid torque loss. Screwdriver 3 reaches the maximum torque the fastest ($\approx 5\,$s), while screwdrivers 7 and 8 require $>9\,$s, indicating a very gradual accumulation of torque.

\paragraph{Pre-slip torque and violation magnitude}
The pre-slip torque is estimated from a window $50$ samples before the peak:
\[ \bar{\tau}_{\mathrm{pre}} = \frac{1}{N} \sum_{i=1}^{N} \tilde{\tau}(t_{\max}-i). \]
Violation magnitude (slip severity):
\[ V = \tau_{\max} - \bar{\tau}_{\mathrm{pre}}. \]
The magnitude of slip violation reflects how sharply the interface fails. Screwdriver 9 shows the largest violation ($0.022$), consistent with a high maximum torque as a result of damage to the palm. Screwdrivers 6–8 show minimal violations, matching their softer torque and weaker grip.

\paragraph{Angular error}
Deviation from a slowly varying trend ($\tau_{\mathrm{trend}}(t) = LPF(\tilde{\tau}(t))$):
\[ E_\theta = \int_0^{t_{\max}} |\tilde{\tau}(t) - \tau_{\mathrm{trend}}(t)| \; dt.\]

The angular error measures how “noisy” the tightening is. Screwdrivers 3–6 show the lowest error, which means smooth and predictable torque growth. Screwdrivers 7–9 show the highest error, consistent with insufficient or strong grip.

\paragraph{Torque ramp slope}
Captures how aggressively torque increases:
\[ S = \max_t \left| \frac{d\tilde{\tau}(t)}{dt} \right|. \]
Screwdriver 4 gives the highest slope ($S=10.20$), meaning that it grips extremely well but has a rapid and less controllable torque. Screwdrivers 3 and 5 show moderate slopes ($S=5$–$6$), balancing engagement and user control. Screwdrivers 7–8 show intermediate slopes that reflect inconsistent contact quality rather than strong interaction.

\paragraph{Max-to-mean ratio}
Measures whether torque increases sharply or gradually:
\[ R = \frac{\tau_{\max}}{\bar{\tau}_{\mathrm{pre}}}. \]

All screwdrivers show $R\approx1$, indicating a smooth, non-impulsive torque buildup. Slightly higher ratios in screwdrivers 7–8 ($R>1.02$) indicate small sharp upward fluctuations that match their noisier torque data.

The extracted metrics characterize two competing mechanical aspects of tool–palm interaction:

\begin{itemize}
    \item \textbf{Frictional engagement quality}: strong contact, rapid torque transfer, high slope.
    \item \textbf{Controllability}: ability to apply torque smoothly and predictably without sudden jumps.
\end{itemize}

A screwdriver with \emph{excellent interaction} may feel “grippy” but may also produce spikes that are harder to modulate. In contrast, a screwdriver with \emph{excellent controllability} may produce limited torque early on. Screwdriver 9 provides the strongest torque transfer, but the biggest slip reduces controllability. Screwdrivers 7 and 8 give weak interaction, long increasing times, and limited torque, but their behavior is predictable and safe. Tool 4 shows the highest slope torque transfer, representing excellent friction but small controllability.

\end{document}